\documentclass[10pt,twocolumn,letterpaper]{article}

\usepackage[T2A,T1]{fontenc}
\usepackage[utf8]{inputenc}
\usepackage[russian,english]{babel}

\usepackage[pagenumbers]{cvpr} 
\usepackage{listings}
\usepackage{multirow}
\usepackage[dvipsnames]{xcolor}

\usepackage{epsfig}
\usepackage{graphicx}
\usepackage{amsmath}
\usepackage{amssymb}
\usepackage{listings}
\usepackage{enumerate}
\usepackage{pbox}
\usepackage{url}
\usepackage{mathtools}
\usepackage{amsmath}
\usepackage{amssymb}

\usepackage{blindtext}
\usepackage{algorithm}
\usepackage{algpseudocode}

\usepackage{epsfig}
\usepackage{graphicx}
\usepackage{float}
\usepackage{subcaption}
\usepackage{caption}
\usepackage{makecell}

\usepackage{epsfig}
\usepackage{graphicx}
\usepackage{subcaption}
\usepackage{enumerate}

\usepackage{booktabs}
\usepackage{pifont}
\usepackage{CJKutf8}

\usepackage{multirow}

\usepackage[misc]{ifsym}

\newcommand{\comment}[1]{}

\definecolor{cvprblue}{rgb}{0.21,0.49,0.74}
\usepackage[pagebackref,breaklinks,colorlinks,citecolor=cvprblue]{hyperref}

\newcommand{\dalle}{DALL$\cdot$E3\xspace}
\newcommand{\midjourney}{Midjourney-v6\xspace}
\newcommand{\ideogram}{Ideogram 1.0\xspace}

\newcommand\paperurl[1]{{\footnotesize{\color{blue}{\url{#1}}}}}

\definecolor{codegreen}{rgb}{0,0.6,0}
\definecolor{codegray}{rgb}{0.5,0.5,0.5}
\definecolor{codepurple}{rgb}{0.58,0,0.82}
\definecolor{backcolour}{rgb}{0.95,0.95,0.92}

\lstdefinestyle{mystyle}{
    backgroundcolor=\color{backcolour},   
    commentstyle=\color{codegreen},
    keywordstyle=\color{magenta},
    numberstyle=\tiny\color{codegray},
    stringstyle=\color{codepurple},
    basicstyle=\sffamily\scriptsize,
    breakatwhitespace=false,
    breaklines=true,
    captionpos=b,     
    keepspaces=true,
    numbers=left,
    numbersep=5pt,
    showspaces=false,
    showstringspaces=false,
    showtabs=false,
    tabsize=2
}

\usepackage{mathtools}

\newlength\savewidth\newcommand\shline{\noalign{\global\savewidth\arrayrulewidth
  \global\arrayrulewidth 1pt}\hline\noalign{\global\arrayrulewidth\savewidth}}
\newcommand{\tablestyle}[2]{\setlength{\tabcolsep}{#1}\renewcommand{\arraystretch}{#2}\centering\footnotesize}

\definecolor{cvprblue}{rgb}{0.21,0.49,0.74}

\newcommand{\redtext}[1]{\noindent \textcolor{red}{{#1}}}

\definecolor{light-light-gray}{gray}{0.90} 

\usepackage{wrapfig} 
\usepackage{amssymb}
\usepackage{pifont}

\RequirePackage[breakable,skins]{tcolorbox}
\RequirePackage{xcolor}

\newcommand\codeurl[1]{{{\color{blue}{\url{#1}}}}}

\makeatletter
\def\@fnsymbol#1{\ensuremath{\ifcase#1\or \ddagger\or \dagger\or
\mathsection\or \mathparagraph\or \|\or **\or \ddagger\ddagger
\or \dagger\dagger \else\@ctrerr\fi}}
\makeatother

\title{Glyph-ByT5-v2: A Strong Aesthetic Baseline for \\Accurate Multilingual Visual Text Rendering}

\author{
{\normalsize \quad Zeyu Liu$^{\dagger}$
 \quad\; Weicong Liang$^{\dagger}$ \quad Yiming Zhao$^{\dagger}$ \quad Bohan Chen$^{\dagger}$}\\
{ \normalsize Lin Liang \quad\quad\; Lijuan Wang \quad\quad\;\; Ji Li \quad\quad\;\; Yuhui Yuan$^{\sharp}$}\\[0mm]
{\footnotesize  $^\dagger$interns at microsoft \qquad  $^\sharp$project lead}\\[1mm]
\normalsize{Microsoft}\\
{\footnotesize\codeurl{{https://glyph-byt5-v2.github.io}}}\vspace{-4mm}}

\begin{document}
\twocolumn[{%
\renewcommand\twocolumn[1][]{#1}%
\maketitle
\begin{center}
\begin{minipage}[t]{1\linewidth}
\vspace{-5mm}
\centering
\begin{minipage}{0.19\textwidth}
{\includegraphics[width=\textwidth]{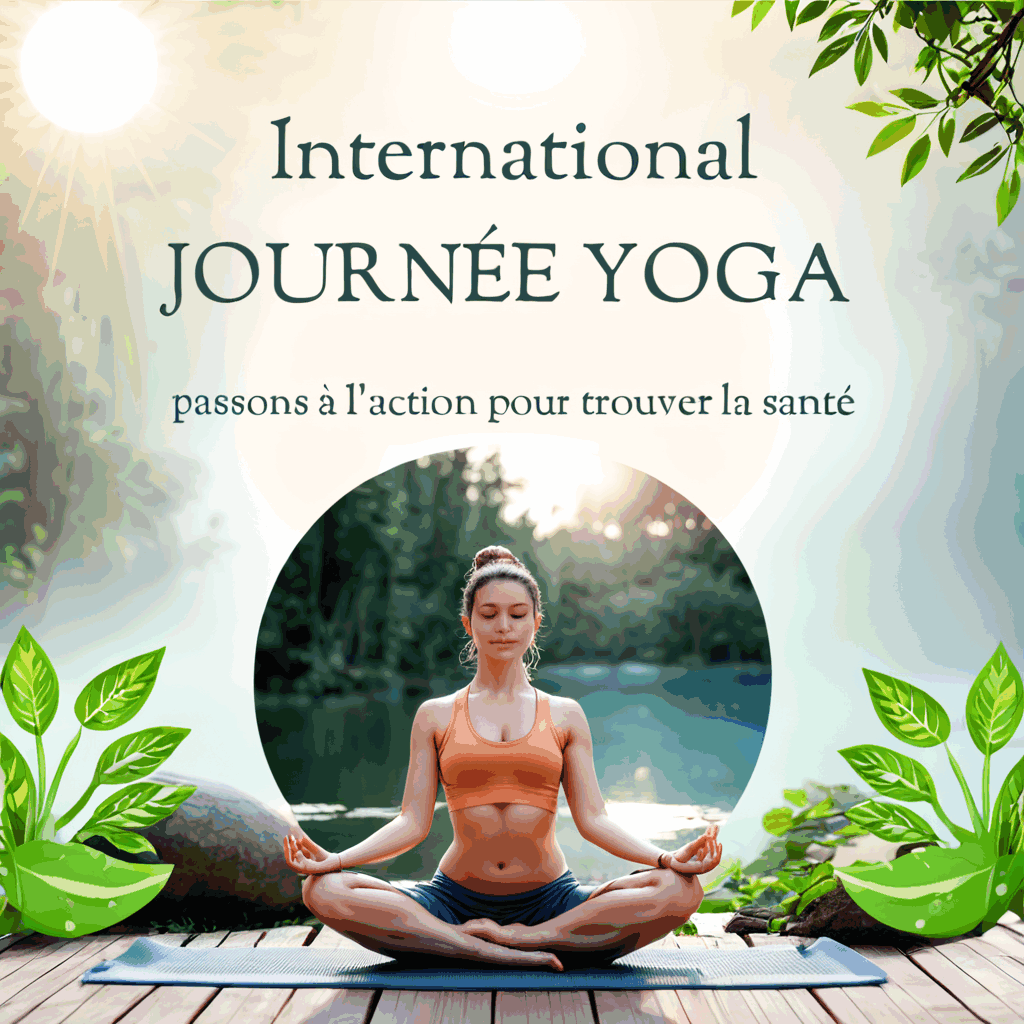}}
\vspace{-4mm}
\end{minipage}
\begin{minipage}{0.19\textwidth}
{\includegraphics[width=\textwidth]{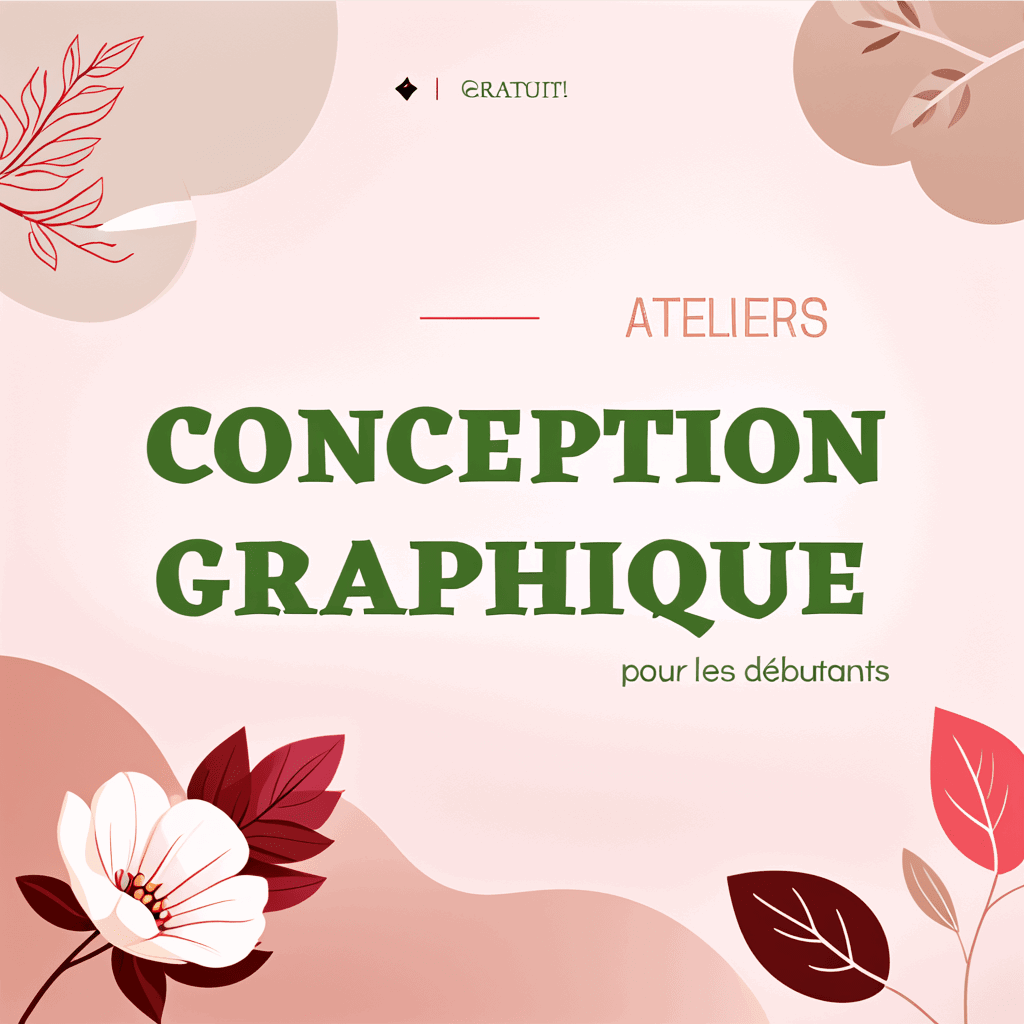}}
\vspace{-4mm}
\end{minipage}
\begin{minipage}{0.19\textwidth}
{\includegraphics[width=\textwidth]{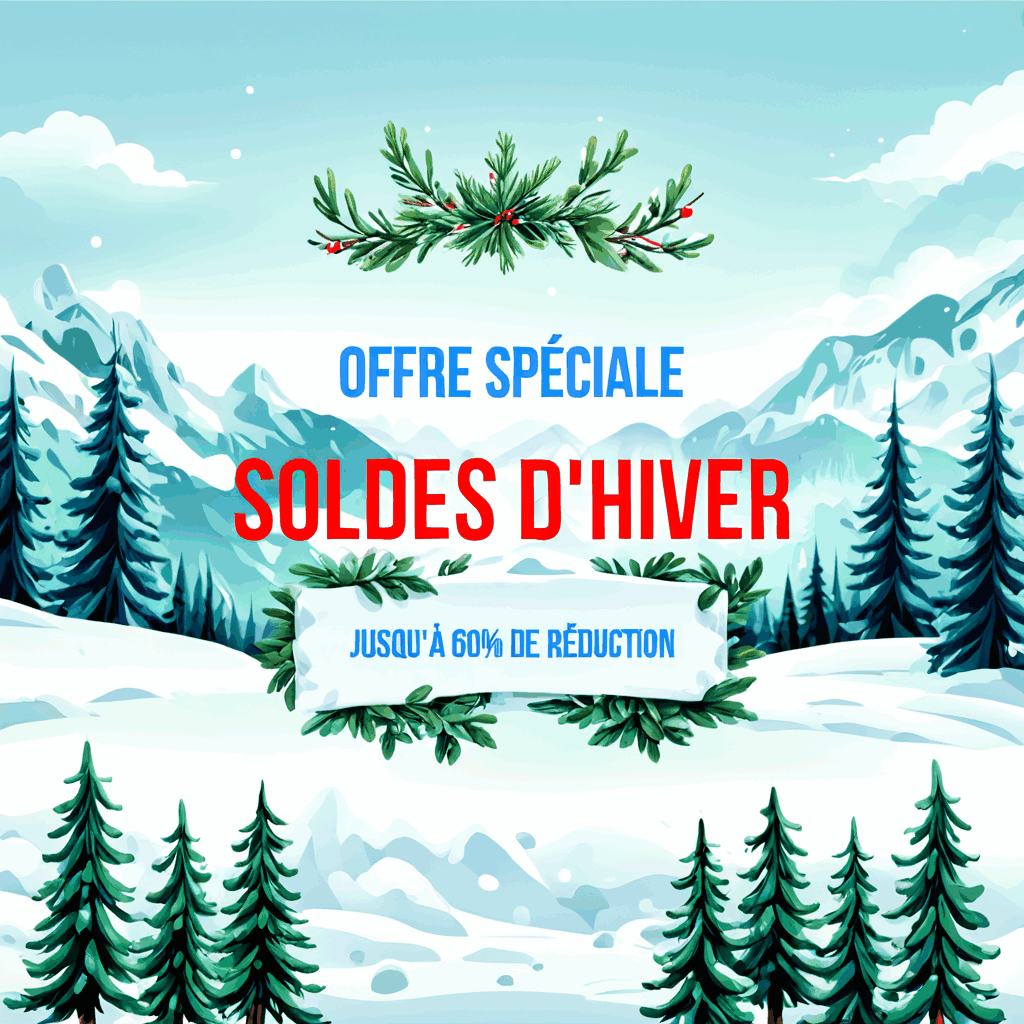}}
\vspace{-4mm}
\end{minipage}
\begin{minipage}{0.19\textwidth}
{\includegraphics[width=\textwidth]{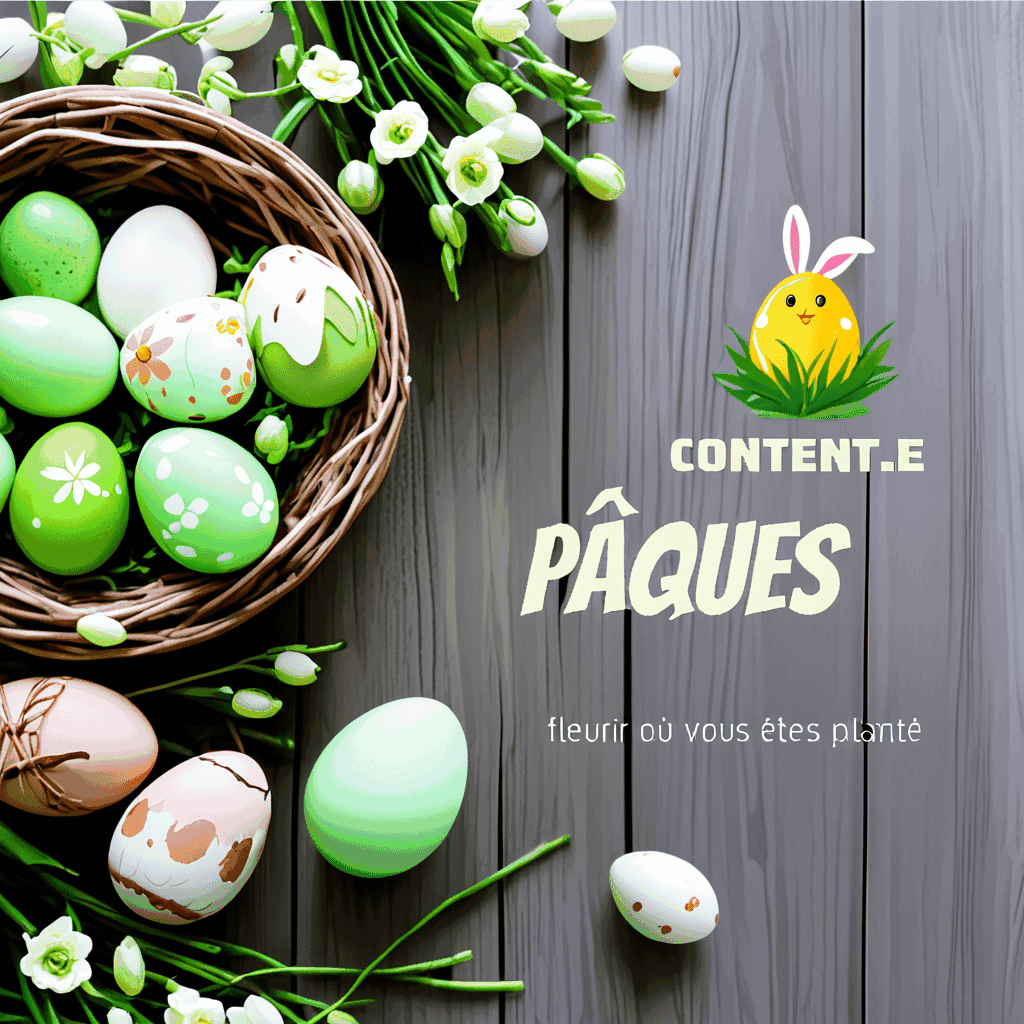}}
\vspace{-4mm}
\end{minipage}
\begin{minipage}{0.19\textwidth}
{\includegraphics[width=\textwidth]{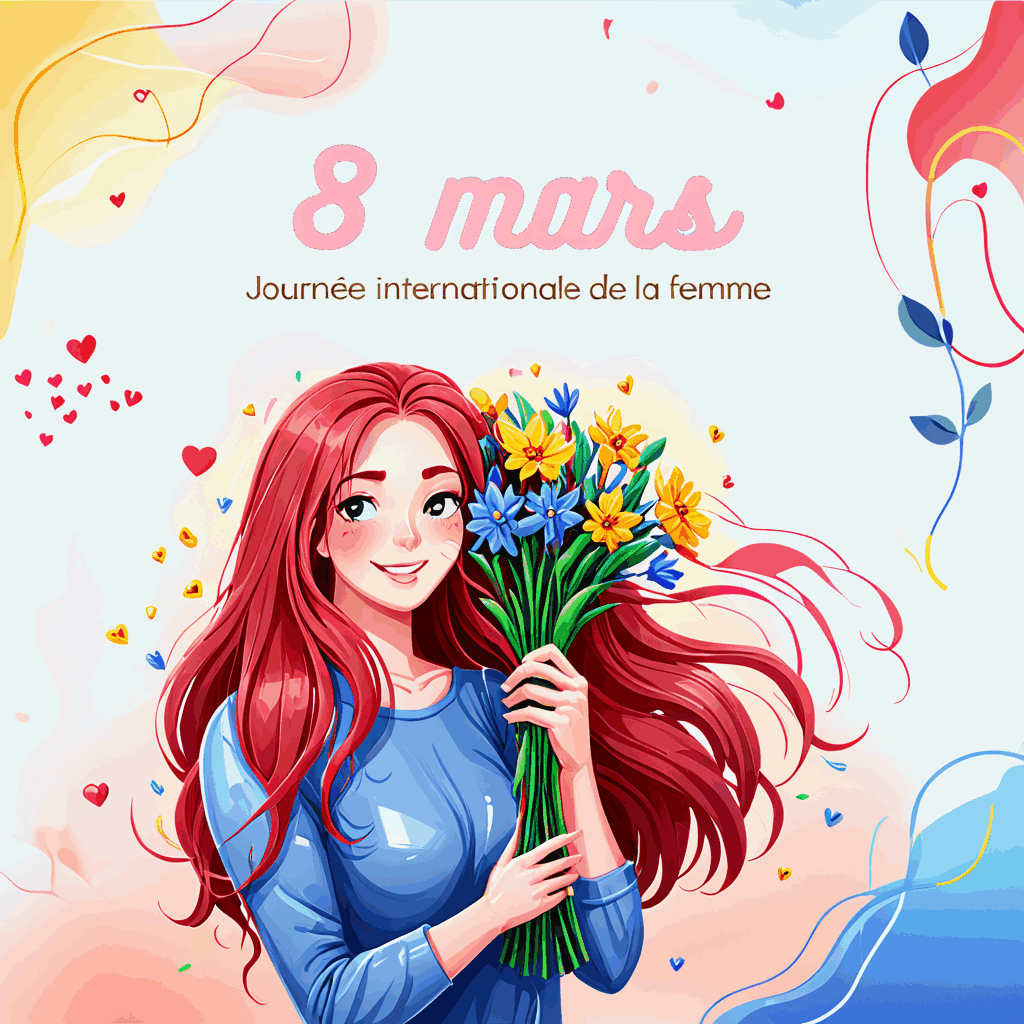}}
\vspace{-4mm}
\end{minipage}\\
\begin{minipage}{0.19\textwidth}
{\includegraphics[width=\textwidth]{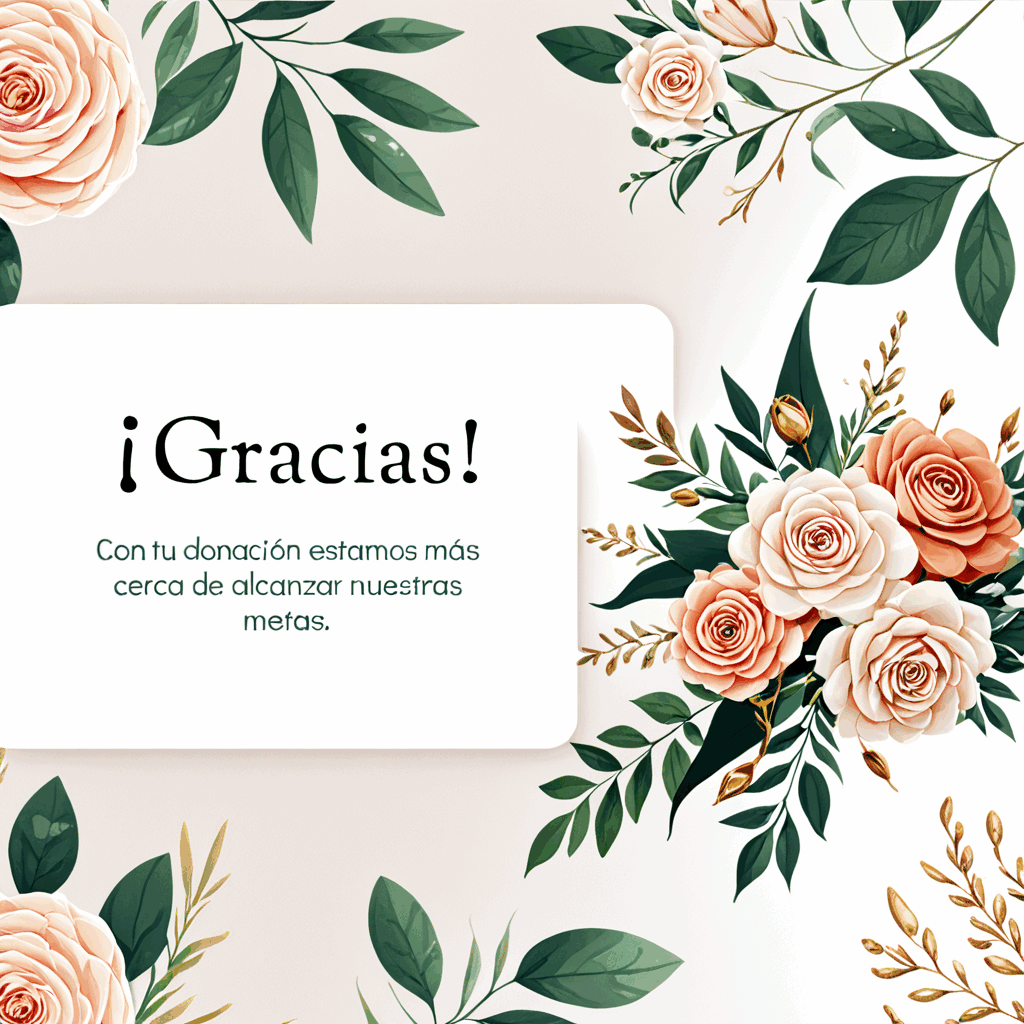}}
\vspace{-4mm}
\end{minipage}
\begin{minipage}{0.19\textwidth}
{\includegraphics[width=\textwidth]{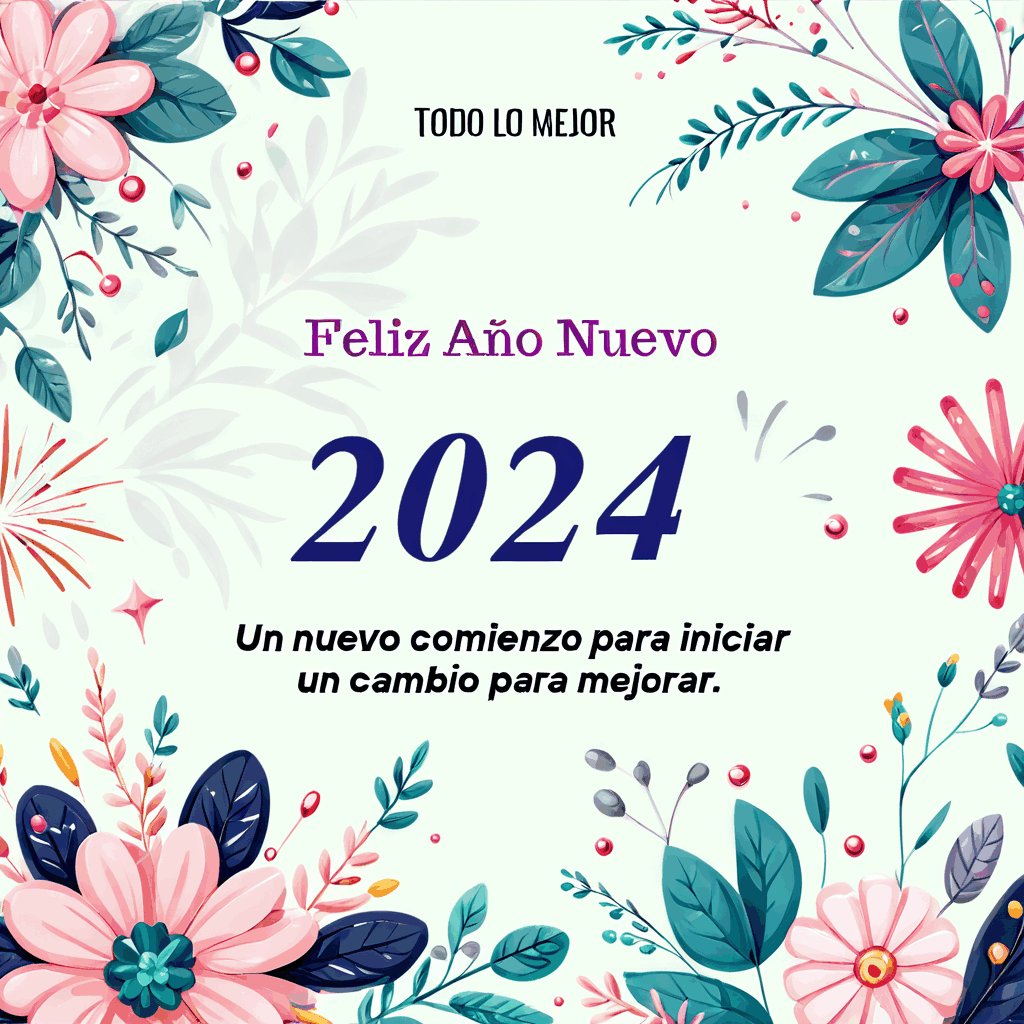}}
\vspace{-4mm}
\end{minipage}
\begin{minipage}{0.19\textwidth}
{\includegraphics[width=\textwidth]{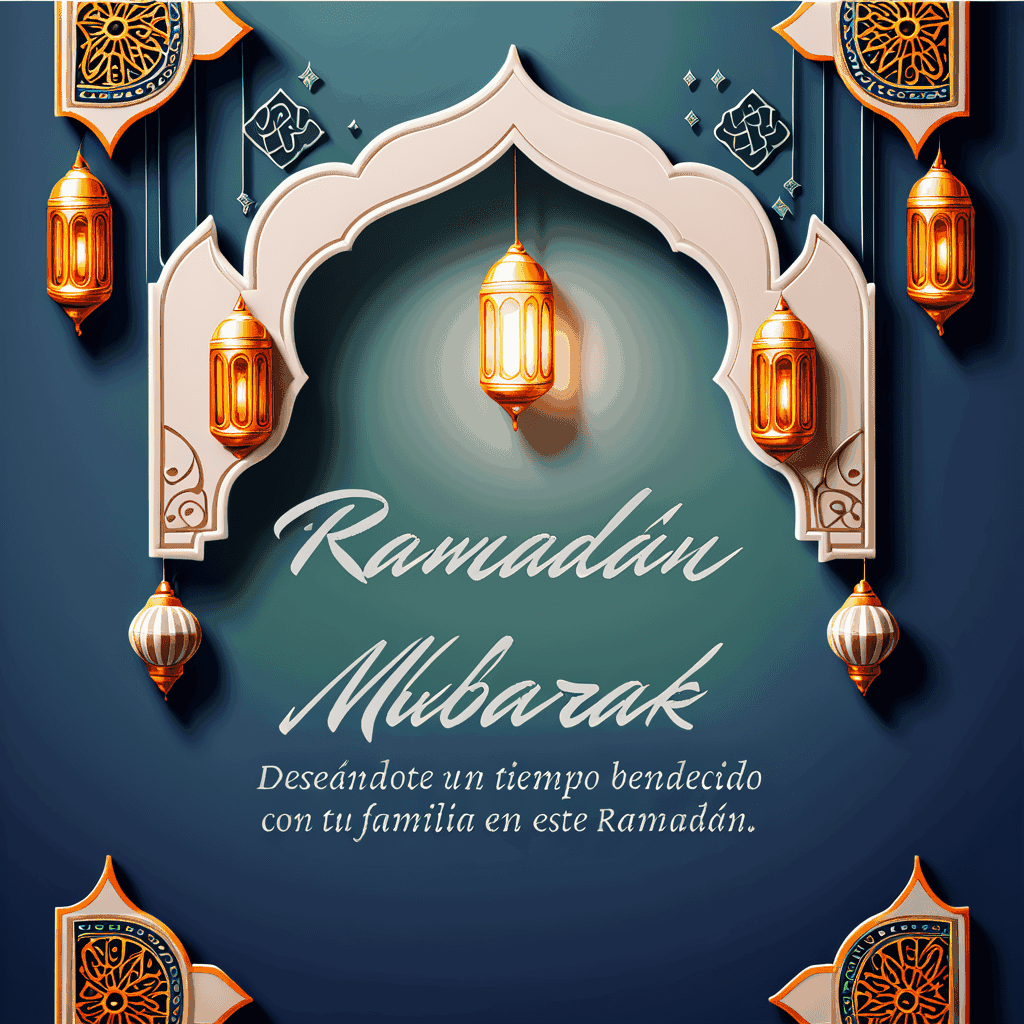}}
\vspace{-4mm}
\end{minipage}
\begin{minipage}{0.19\textwidth}
{\includegraphics[width=\textwidth]{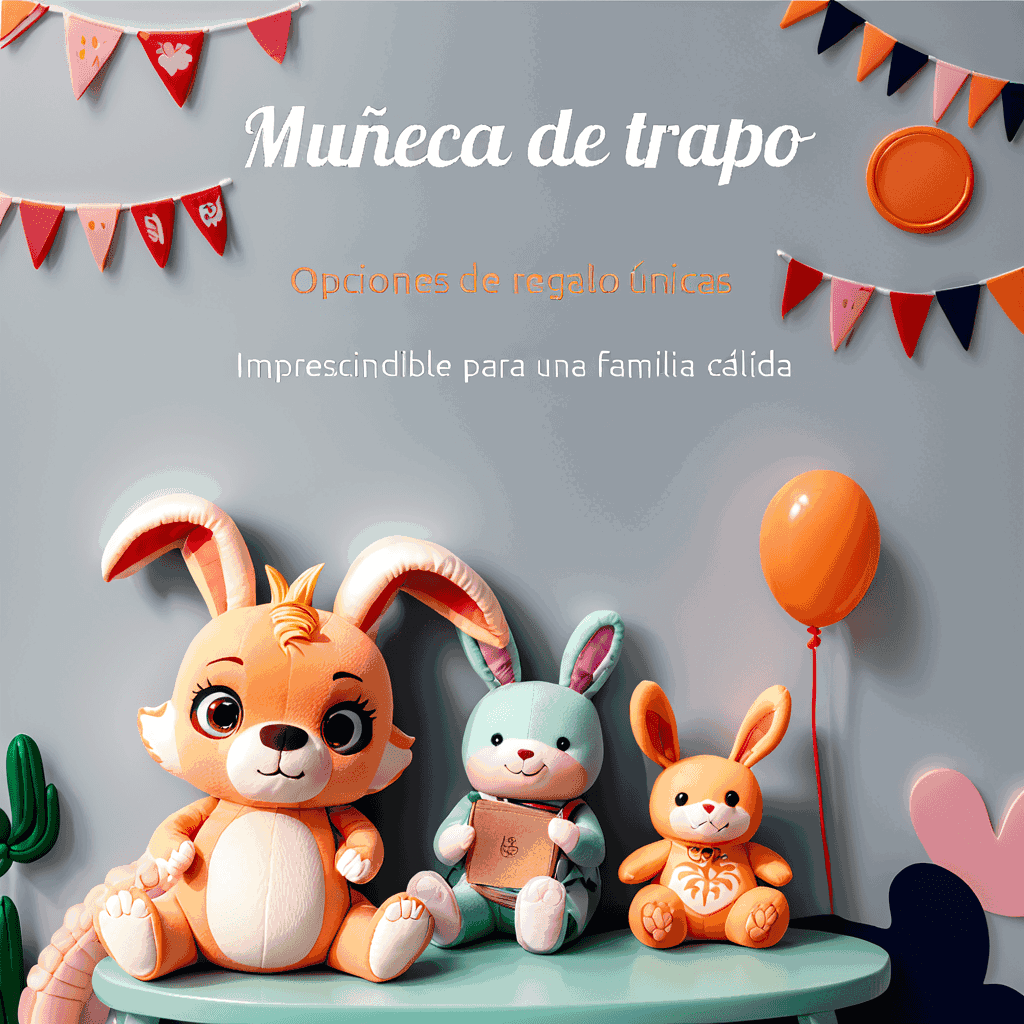}}
\vspace{-4mm}
\end{minipage}
\begin{minipage}{0.19\textwidth}
{\includegraphics[width=\textwidth]{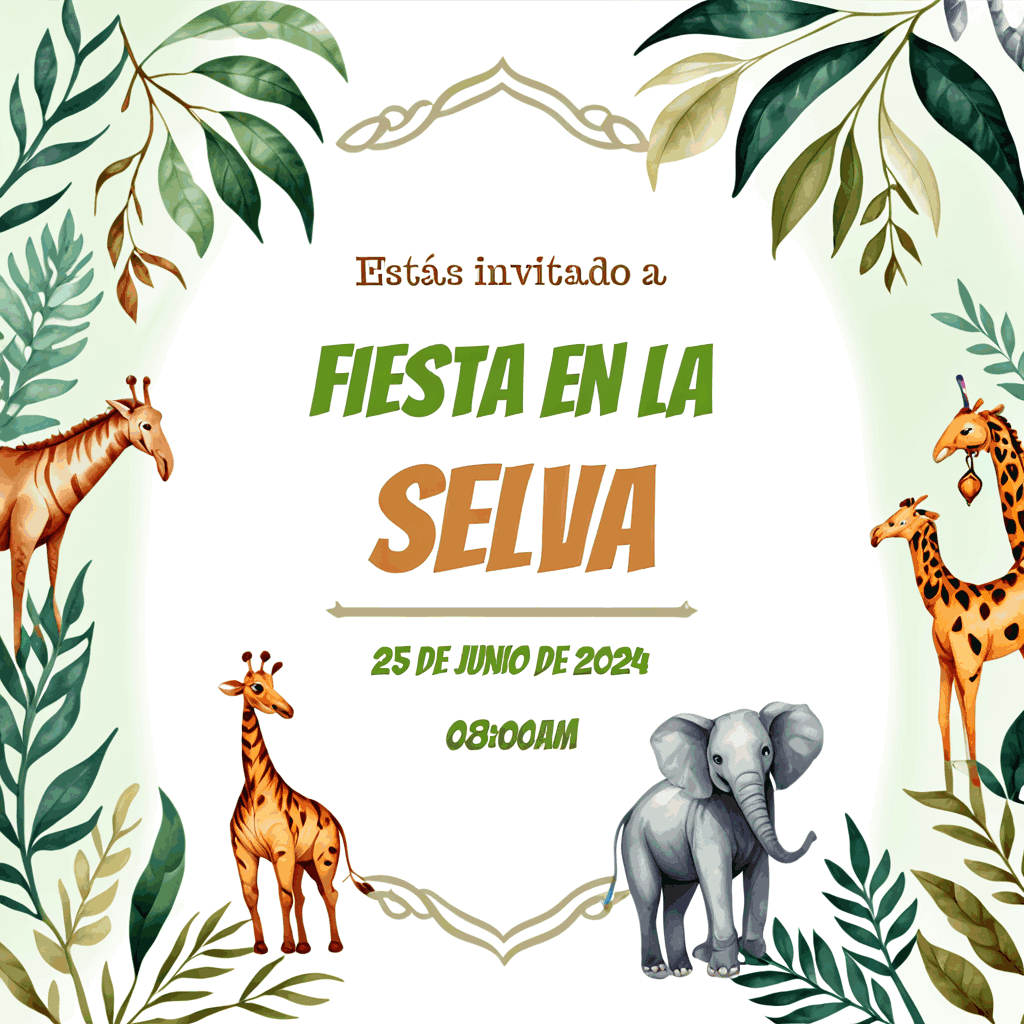}}
\vspace{-4mm}
\end{minipage}\\
\begin{minipage}{0.19\textwidth}
{\includegraphics[width=\textwidth]{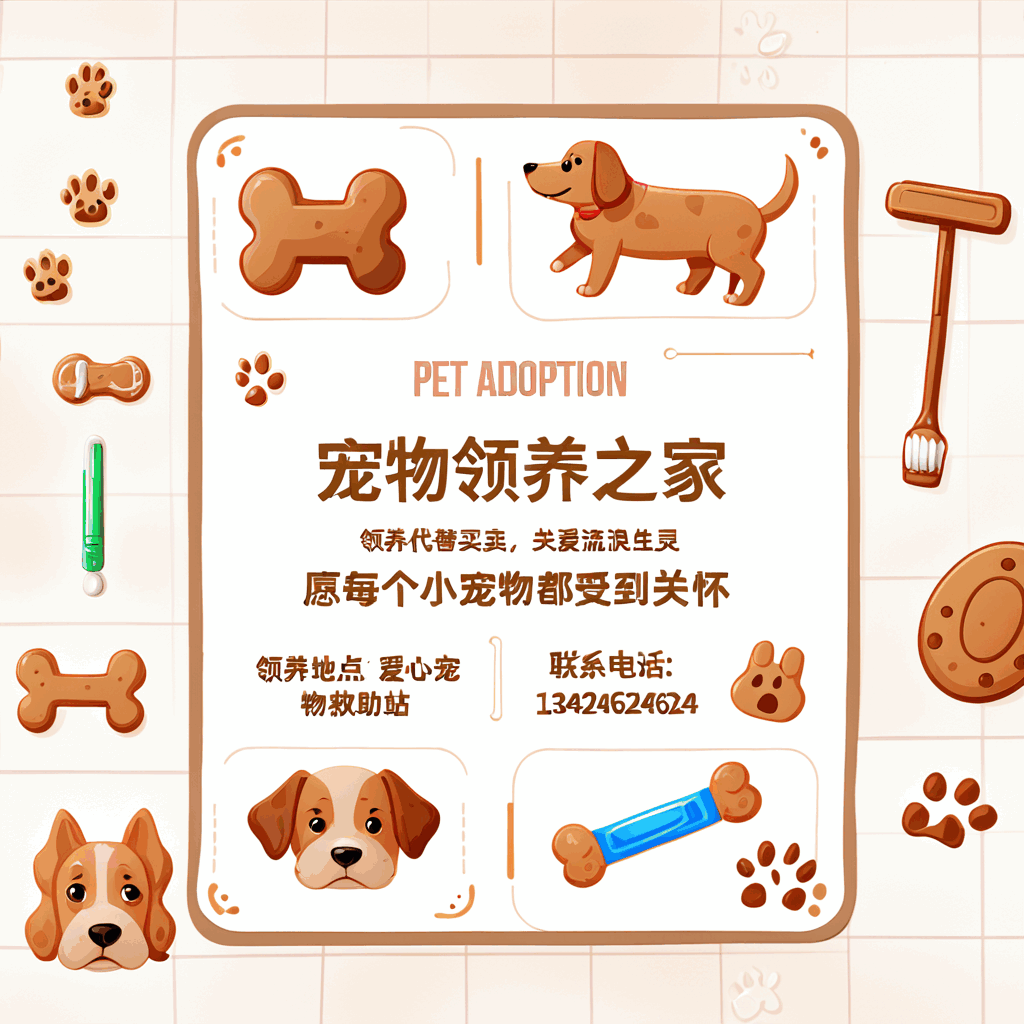}}
\vspace{-4mm}
\end{minipage}
\begin{minipage}{0.19\textwidth}
{\includegraphics[width=\textwidth]{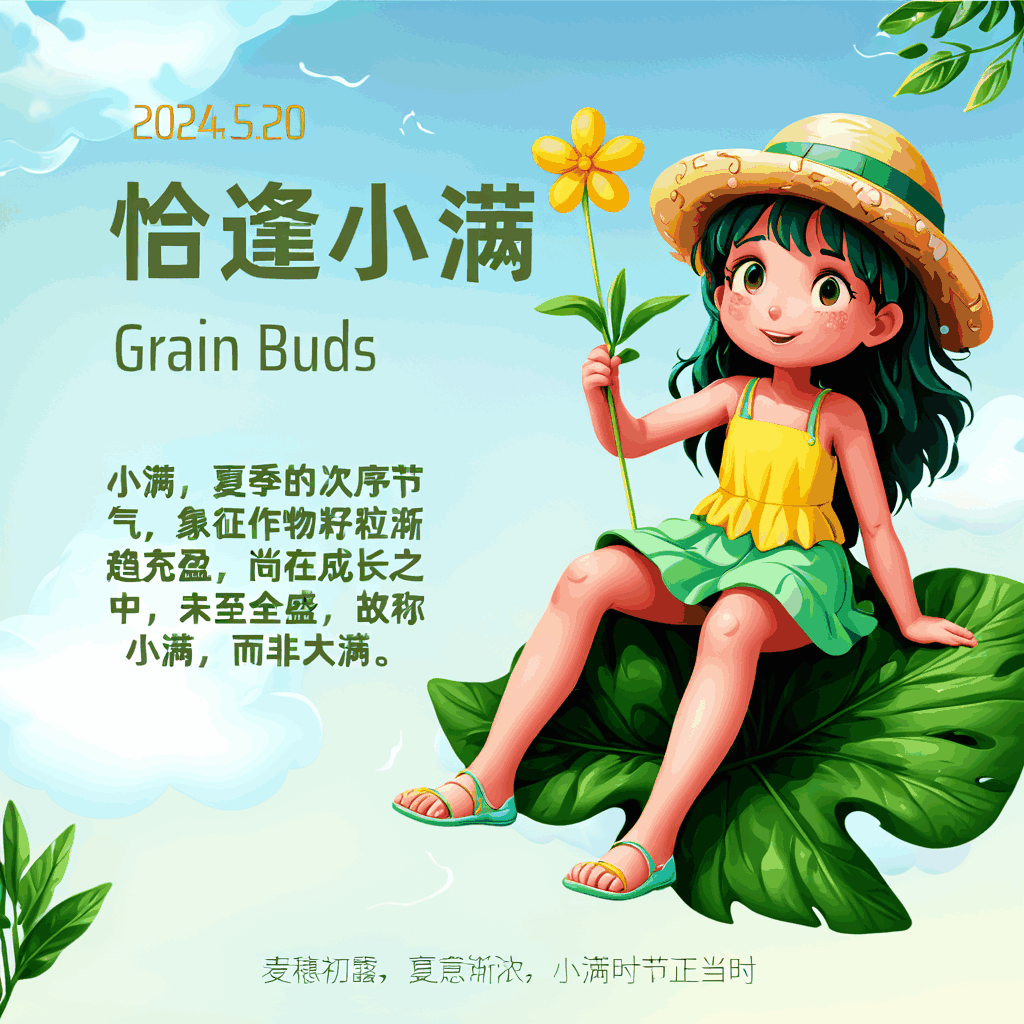}}
\vspace{-4mm}
\end{minipage}
\begin{minipage}{0.19\textwidth}
{\includegraphics[width=\textwidth]{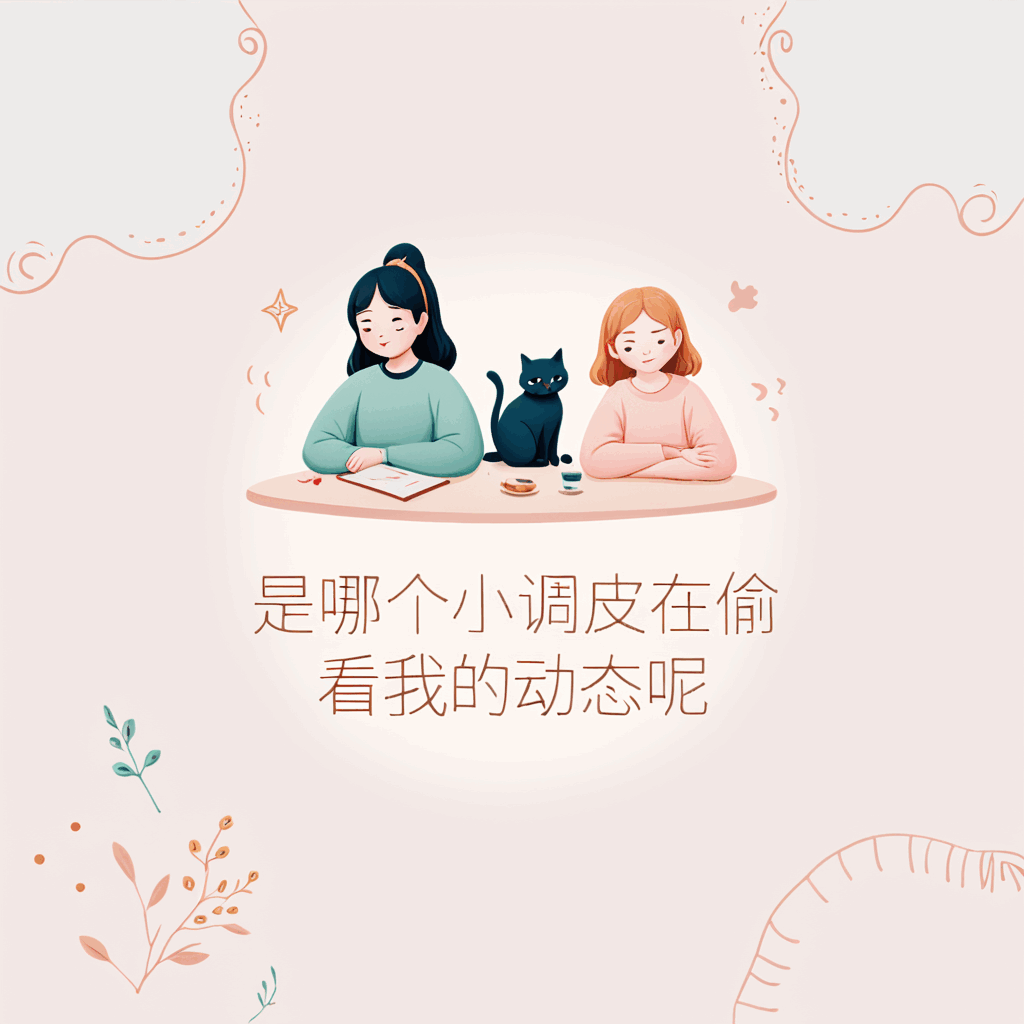}}
\vspace{-4mm}
\end{minipage}
\begin{minipage}{0.19\textwidth}
{\includegraphics[width=\textwidth]{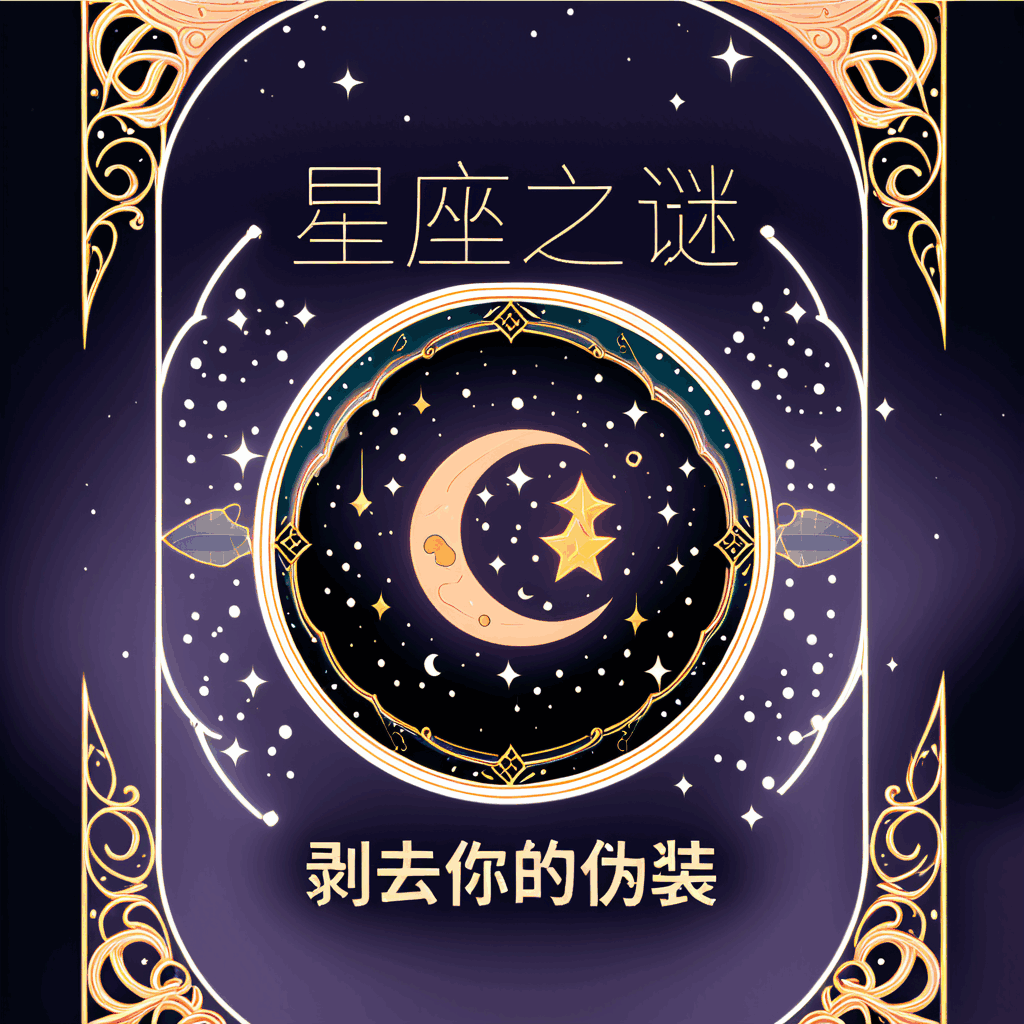}}
\vspace{-4mm}
\end{minipage}
\begin{minipage}{0.19\textwidth}
{\includegraphics[width=\textwidth]{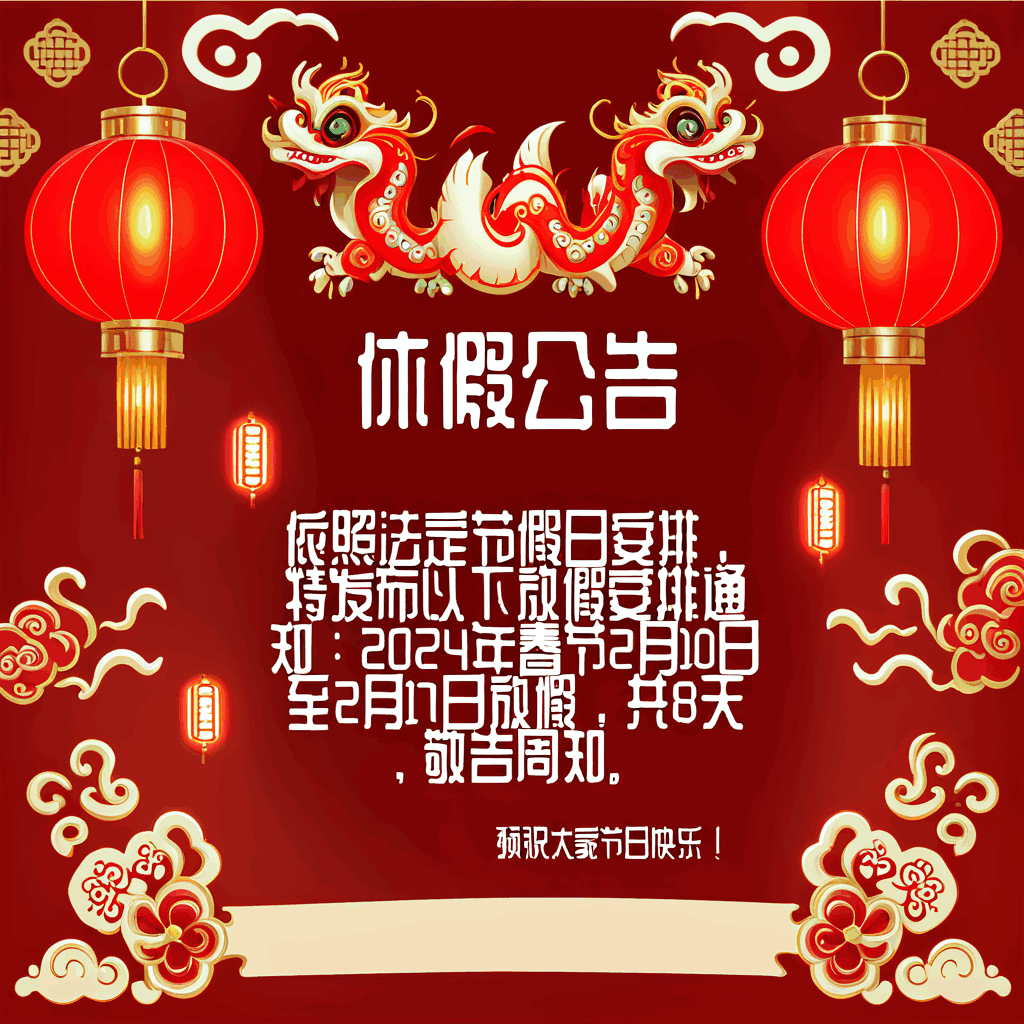}}
\vspace{-4mm}
\end{minipage}\\
\begin{minipage}{0.19\textwidth}
{\includegraphics[width=\textwidth]{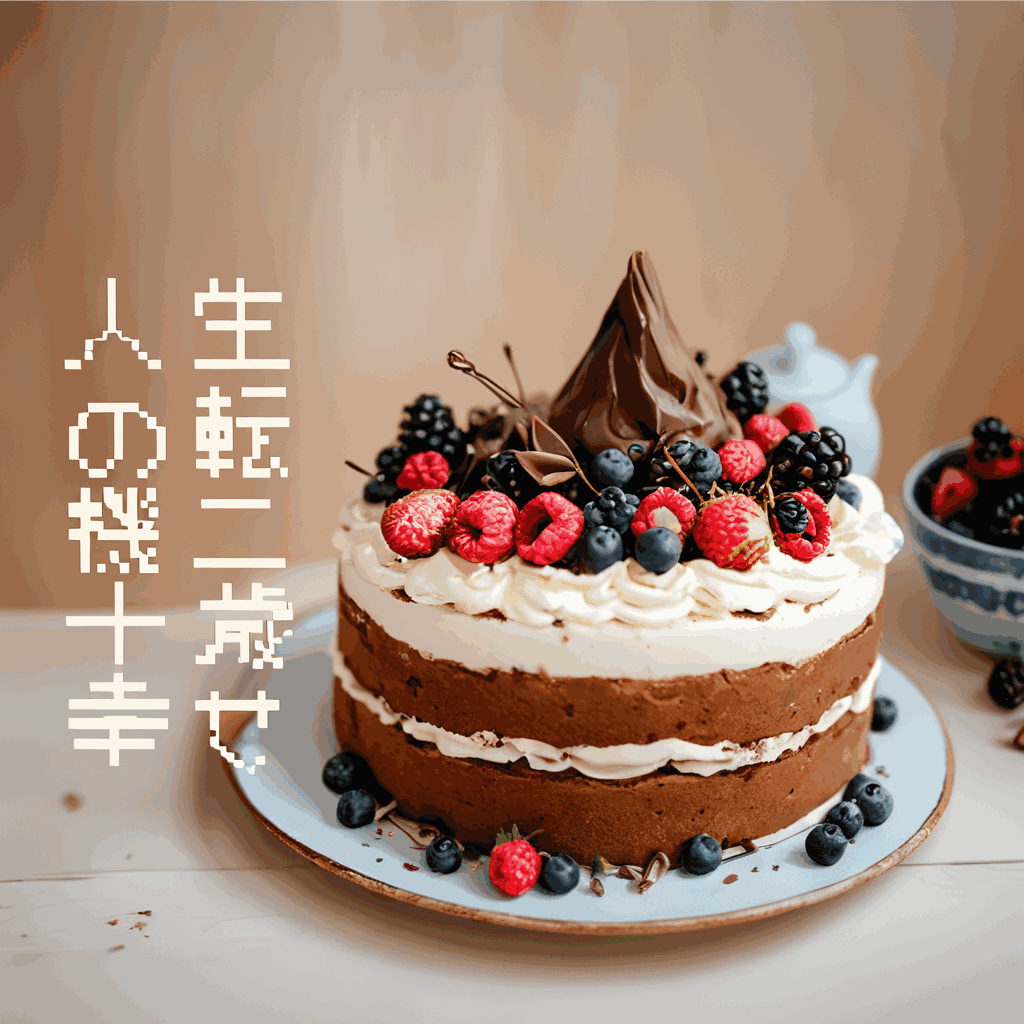}}
\vspace{-4mm}
\end{minipage}
\begin{minipage}{0.19\textwidth}
{\includegraphics[width=\textwidth]{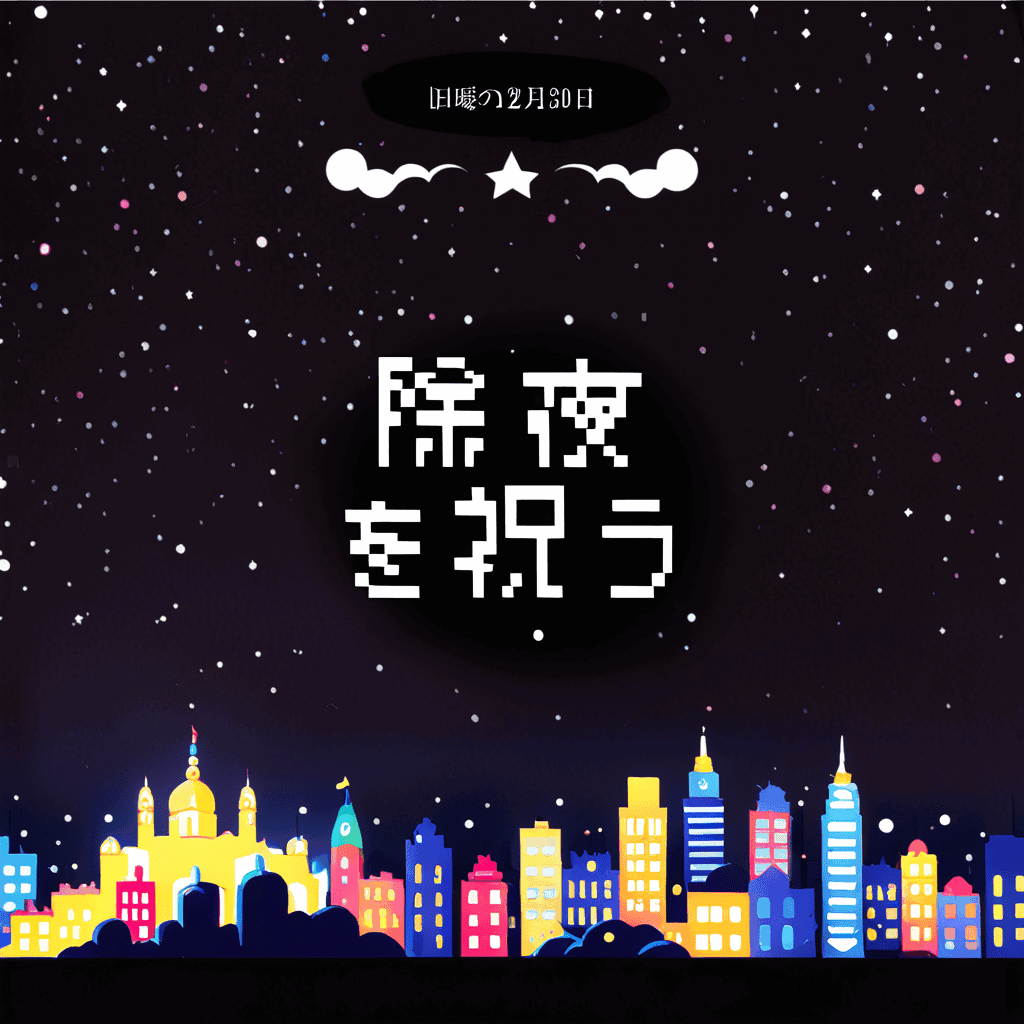}}
\vspace{-4mm}
\end{minipage}
\begin{minipage}{0.19\textwidth}
{\includegraphics[width=\textwidth]{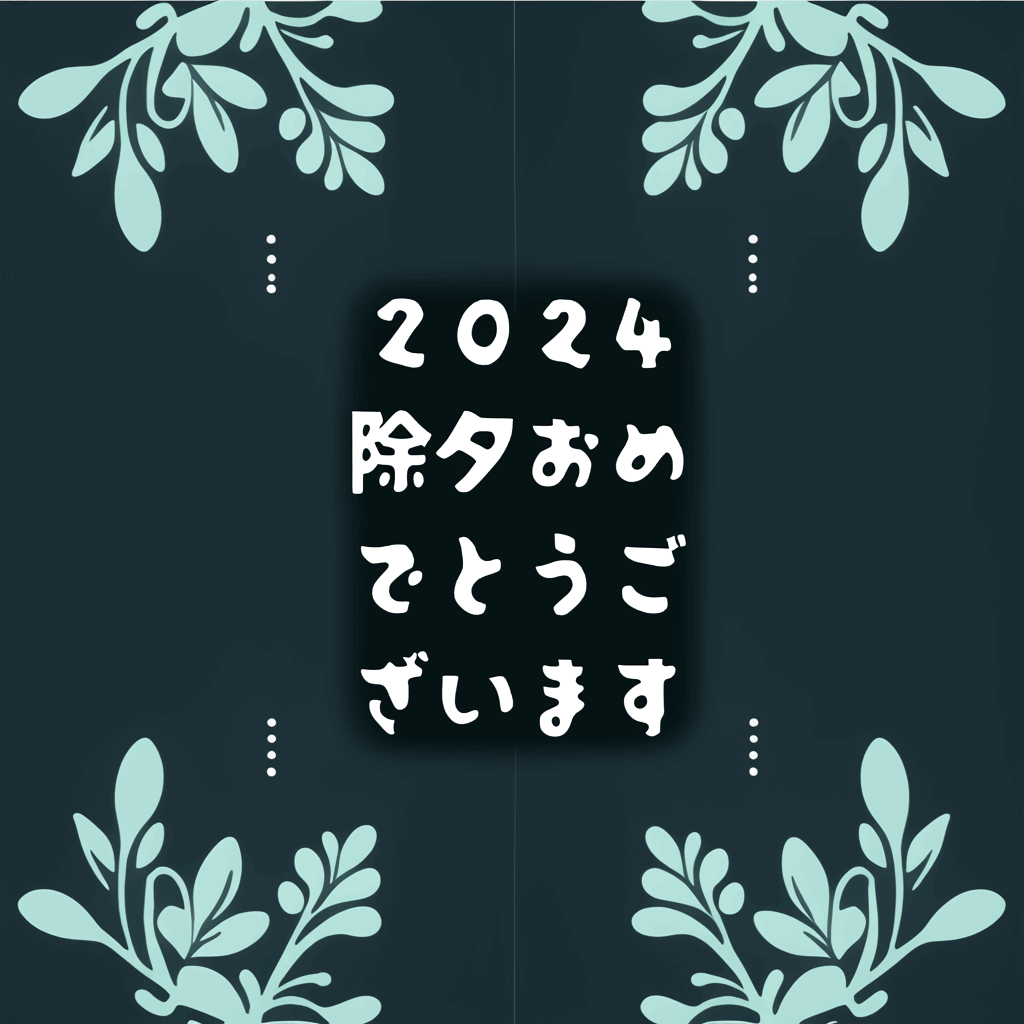}}
\vspace{-4mm}
\end{minipage}
\begin{minipage}{0.19\textwidth}
{\includegraphics[width=\textwidth]{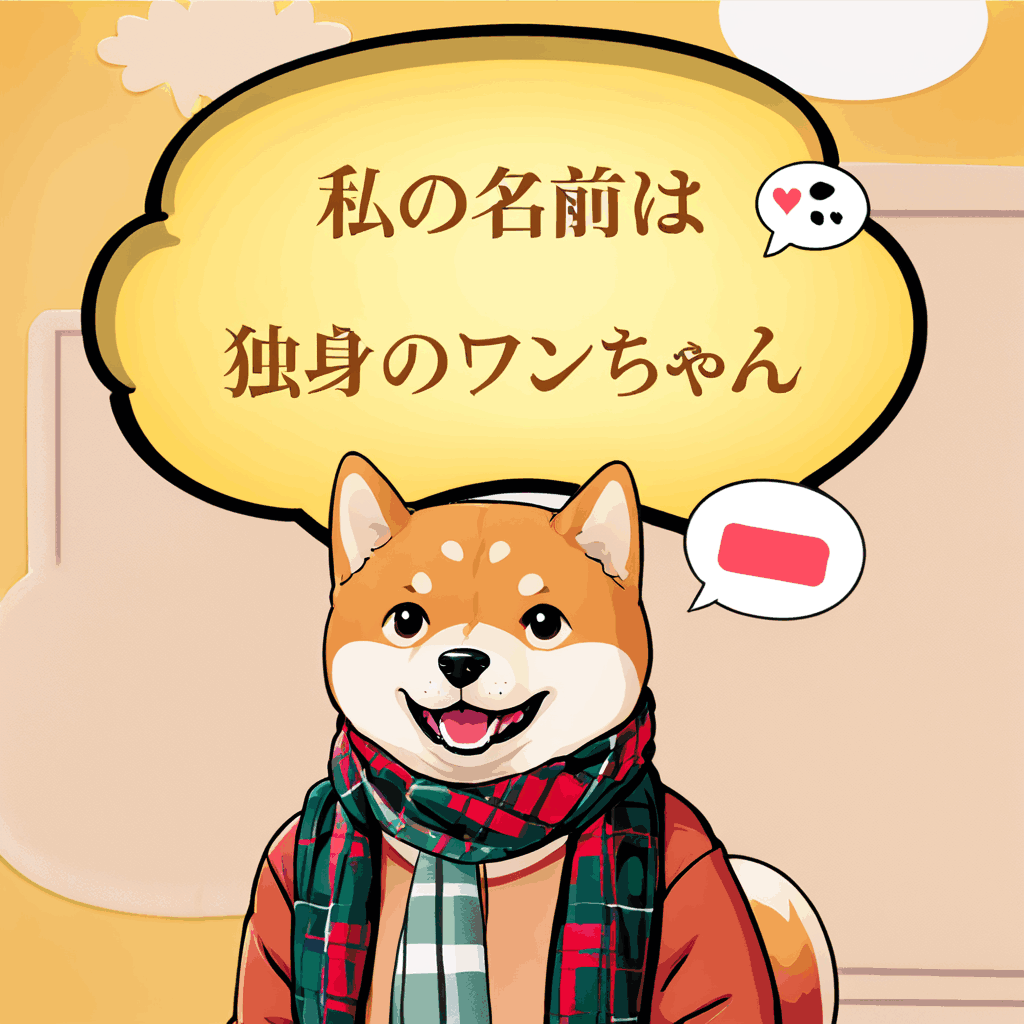}}
\vspace{-4mm}
\end{minipage}
\begin{minipage}{0.19\textwidth}
{\includegraphics[width=\textwidth]{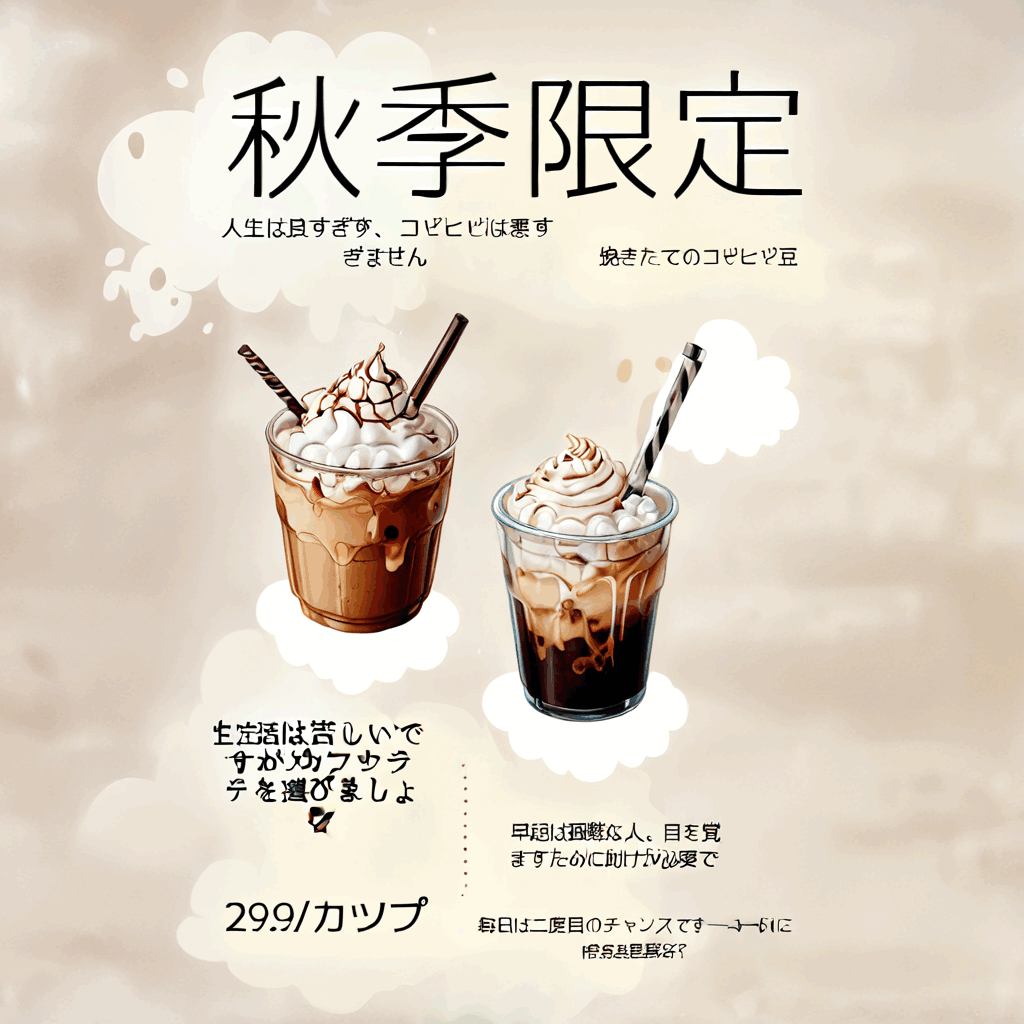}}
\vspace{-4mm}
\end{minipage}\\
\begin{minipage}{0.19\textwidth}
{\includegraphics[width=\textwidth]{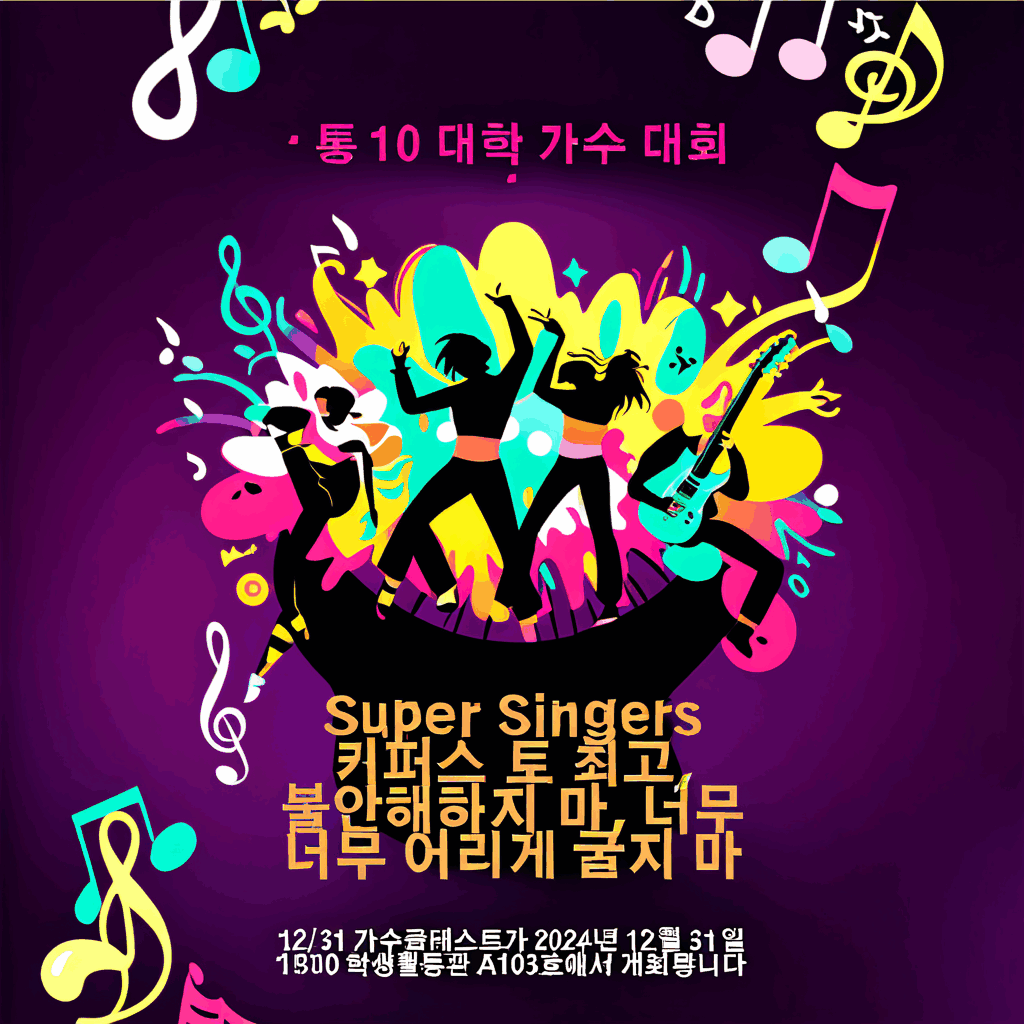}}
\vspace{-4mm}
\end{minipage}
\begin{minipage}{0.19\textwidth}
{\includegraphics[width=\textwidth]{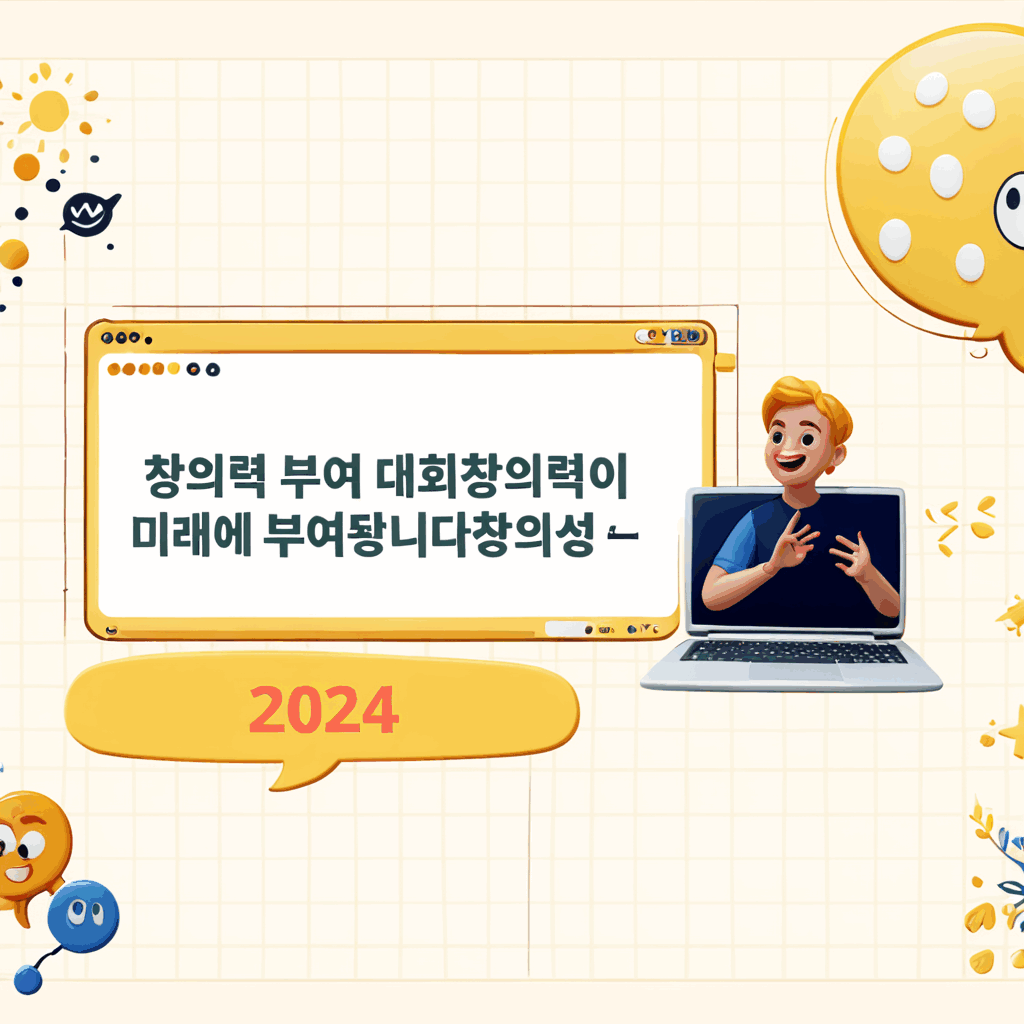}}
\vspace{-4mm}
\end{minipage}
\begin{minipage}{0.19\textwidth}
{\includegraphics[width=\textwidth]{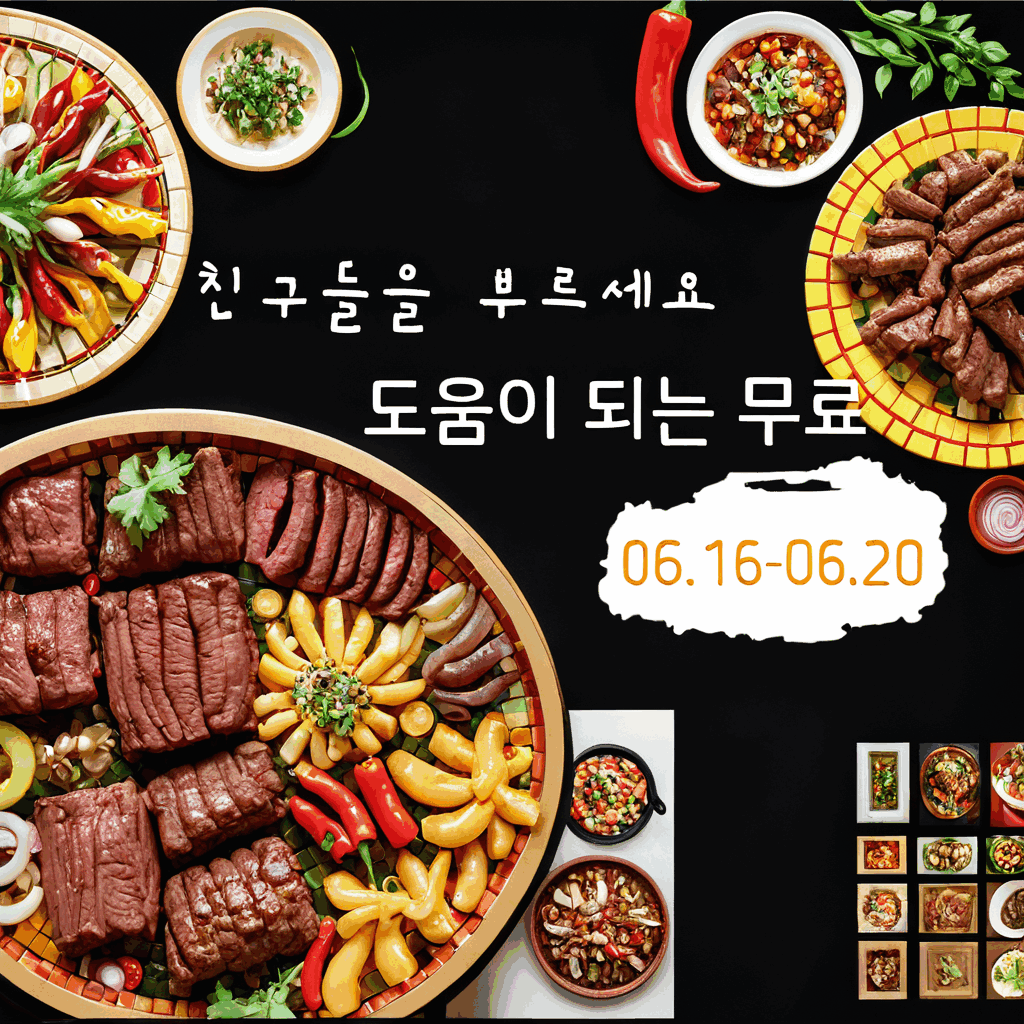}}
\vspace{-4mm}
\end{minipage}
\begin{minipage}{0.19\textwidth}
{\includegraphics[width=\textwidth]{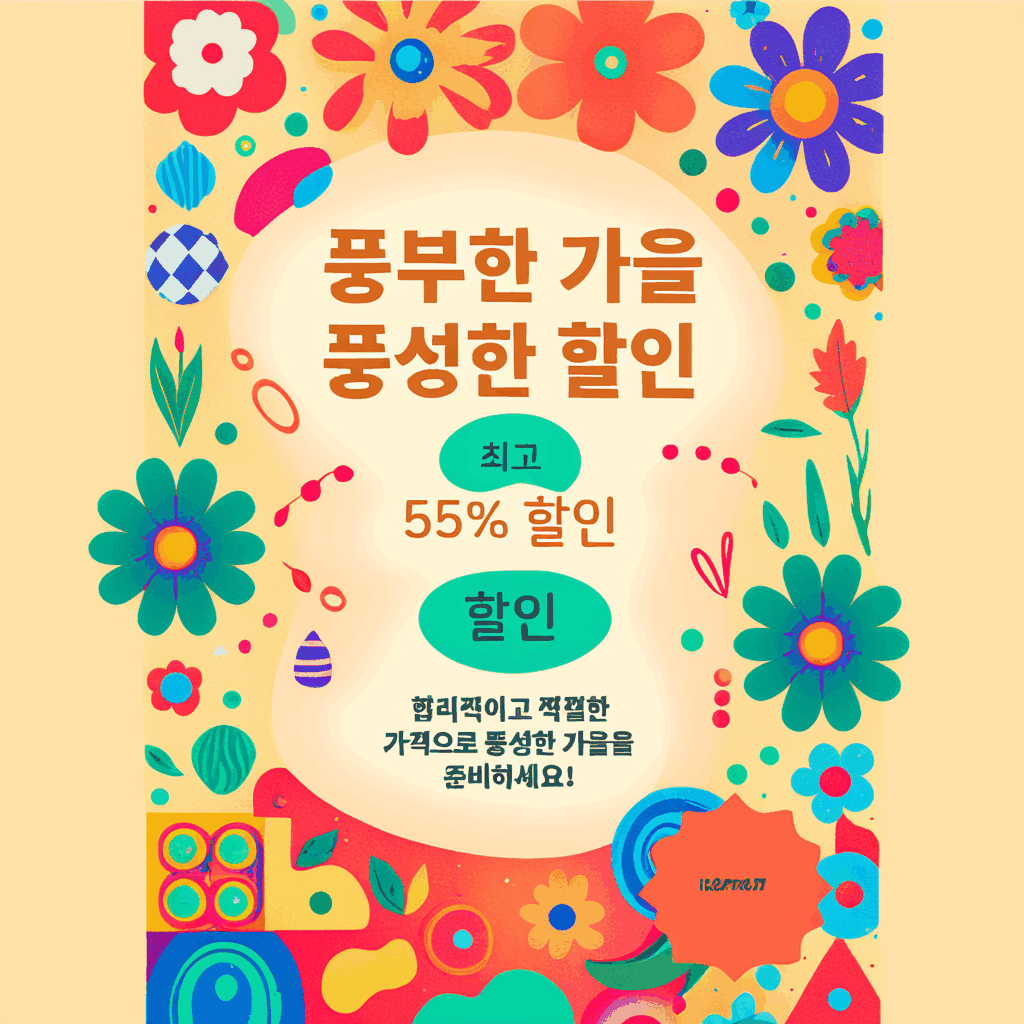}}
\vspace{-4mm}
\end{minipage}
\begin{minipage}{0.19\textwidth}
{\includegraphics[width=\textwidth]{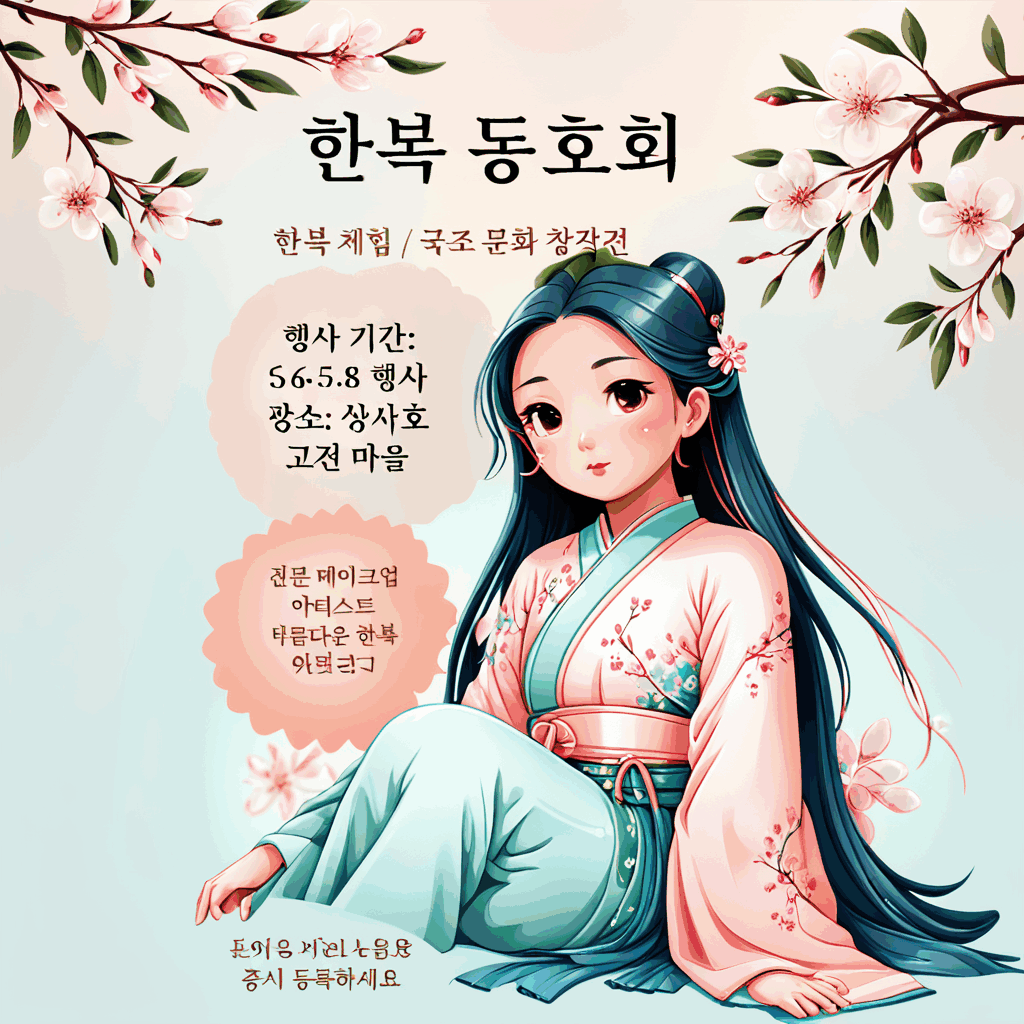}}
\vspace{-4mm}
\end{minipage}
\vspace{-3mm}
\captionof{figure}{\small{Illustrating the multilingual visual text rendering results with our approach. We show the French, Spanish, Chinese, Japanese, and Korean visual text results in the 1st, 2nd, 3rd, 4th, and 5th rows, respectively.}}
\label{fig:teaser}
\end{minipage}
\end{center}
}]

\begin{abstract}
Recently, Glyph-ByT5 has achieved highly accurate visual text rendering performance in graphic design images. However, it still focuses solely on English and performs relatively poorly in terms of visual appeal.
In this work, we address these two fundamental limitations by presenting Glyph-ByT5-v2 and Glyph-SDXL-v2, which not only support accurate visual text rendering for 10 different languages but also achieve much better aesthetic quality.
To achieve this, we make the following contributions: (i) creating a high-quality multilingual glyph-text and graphic design dataset consisting of more than 1 million glyph-text pairs and 10 million graphic design image-text pairs covering nine other languages, (ii) building a multilingual visual paragraph benchmark consisting of 1,000 prompts, with 100 for each language, to assess multilingual visual spelling accuracy, and (iii) leveraging the latest step-aware preference learning approach to enhance the visual aesthetic quality.
With the combination of these techniques, we deliver a powerful customized multilingual text encoder, Glyph-ByT5-v2, and a strong aesthetic graphic generation model, Glyph-SDXL-v2, that can support accurate spelling in $\sim10$ different languages. We perceive our work as a significant advancement, considering that the latest \dalle and \ideogram still struggle with the multilingual visual text rendering task.
\end{abstract}
\section{Introduction}
\label{sec:intro}

Visual text rendering has attracted huge interest and become a critical evaluation aspect in various latest commercial-level text-to-image generation models like \dalle, \midjourney, and \ideogram. The research community has also observed significant progress in the visual text generation task since the development of the more effective rectified flow approach in Stable Diffusion 3~\cite{esser2024scaling} or the customized text encoder Glyph-ByT5~\cite{liu2024glyph}.
However, all these efforts still focus on a single language, i.e., English. Visual text rendering for other languages, especially Chinese, Japanese, and Korean are still very challenging for existing approaches due to the scarcity of high-quality data.
Figure~\ref{fig:dalle_ideogram} illustrates the generated images with multilingual visual text based on \dalle and \ideogram, respectively. We can observe that most of their visual spelling accuracy is nearly zero, highlighting that the multilingual visual text rendering task poses a fundamental challenge for both leading commercial text-to-image generation models.

\begin{figure}[htbp]
\centering
\includegraphics[width=\linewidth]{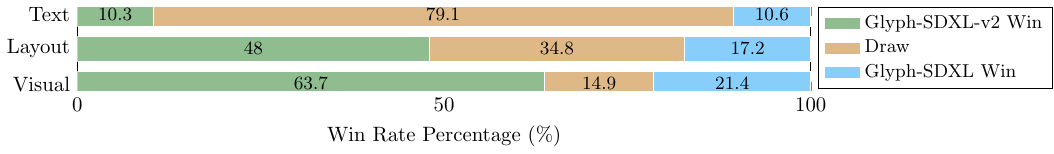} 
\caption{\footnotesize{Glyph-SDXL-v2 vs. Glyph-SDXL in graphic design images in terms of multilingual visual text spelling accuracy, layout quality, and visual aesthetics win-rates evaluated by human evaluator preferences. The only difference between Glyph-SDXL-v2 and Glyph-SDXL is that we replace SDXL with Albedo SDXL + SPO.}}
\label{fig:winrate_spo_sdxl_vs_sdxl}
\end{figure}

\begin{table}[!t]
\begin{minipage}[t]{1\linewidth}
\centering
\tablestyle{1pt}{1.5}
\resizebox{1.0\linewidth}{!}
{
\begin{tabular}{l|c|c|cccc}
\multirow{2}{*}{Language} & \multirow{2}{*}{\# unique chars} & \multirow{2}{*}{\# font types (\# OFL)\footnote{\scriptsize{The number of open-source font types licensed under the SIL Open Font License that permits commercial usage.}})} & \multicolumn{4}{c}{Precision ($\%$)} \\\cline{4-7}
& & & $\le$20 chars & $\le$20-50 chars & $\le$50-100 chars & $\ge$100 chars  \\
\shline
English & $52$ & $512$ ($305$) & ${95.87}$ & ${95.55}$ & ${93.94}$ & ${96.76}$ \\
French & $52$+$28$ & $495$ ($294$) & ${93.1}$ & ${90.67}$ & ${93.69}$ & ${88.05}$ \\
Spanish & $52$+$14$ & $473$ ($286$) & ${92.33}$ & ${88.75}$ & ${88.59}$ & ${91.33}$ \\
German &  $52$+$8$ & $471$ ($279$) & ${91.13}$ & ${87.65}$ & ${88.98}$ & ${90.97}$ \\
Portuguese & $52$+$24$ & $497$ ($295$) & ${82.27}$ & ${87.00}$ & ${89.92}$ & ${90.01}$ \\
Italian & $52$+$20$ & $498$ ($295$) & ${90.33}$ & ${88.28}$ & ${86.55}$ & ${93.30}$ \\ 
Russian & $66$ & $160$ ($102$) & ${91.67}$ & ${87.85}$ & ${85.19}$ & ${85.54}$ \\\hline
Chinese & $5,000$ & $143$ ($61$) & ${97.56}$ & ${97.51}$ & ${91.84}$ & ${83.94}$ \\
Japanese & $1,148$ & $211$ ($111$) & ${89.11}$ & ${90.82}$ & ${85.76}$ & ${82.38}$ \\
Korean & $617$ & $122$ ($92$) & ${91.16}$ & ${92.73}$ & ${86.54}$ & ${79.91}$ \\
\end{tabular}
}
\vspace{2mm}
\caption{
\small{Illustrating the multilingual visual text rendering results achieved with our approach across a varying number of characters. Performance is demonstrated by evaluating word-level precision for the seven languages listed above and character-level precision for the three languages listed below. All results are based on a single model rather than multiple models tailored for each language.
}}
\label{tab:teaser_table}
\end{minipage}
\end{table}

\begin{figure*}
\centering
\begin{minipage}{0.19\textwidth}
{\includegraphics[width=\textwidth]{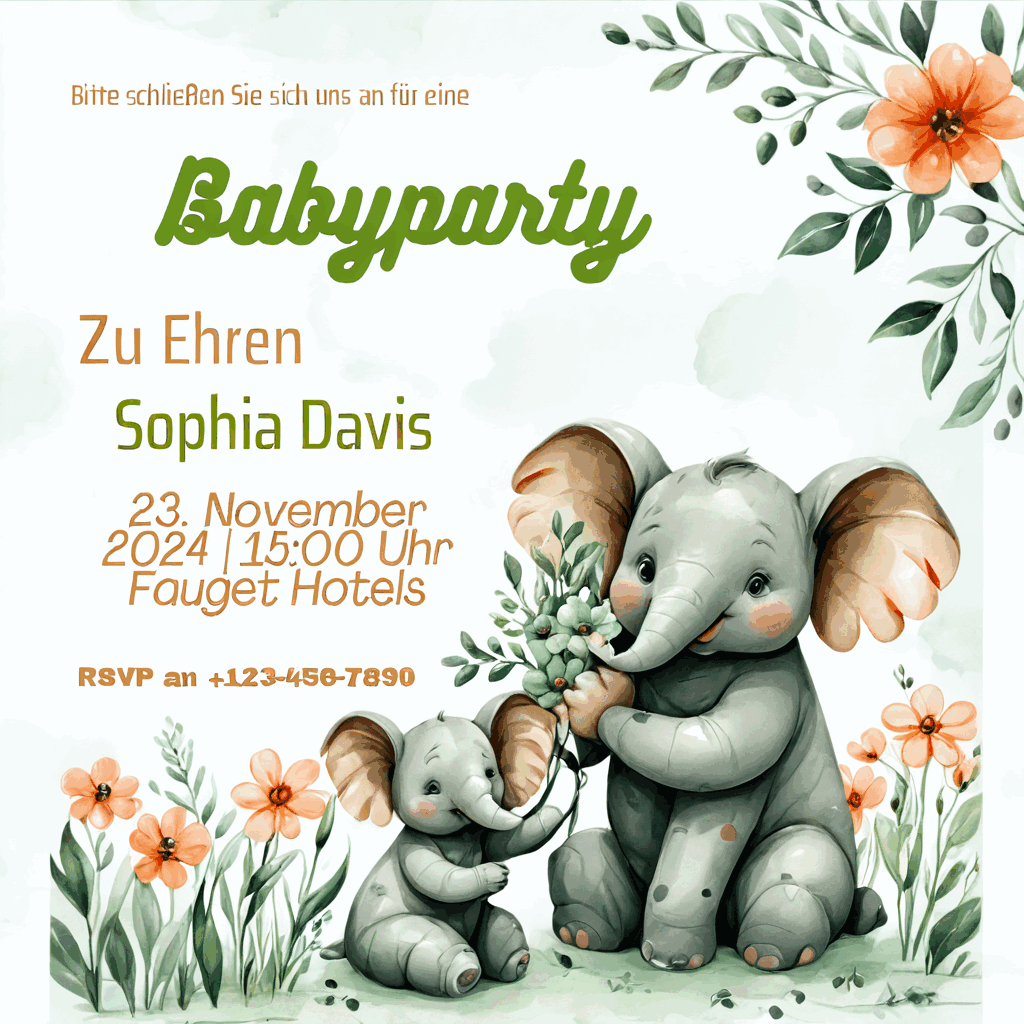}}
\vspace{-3mm}
\end{minipage}
\begin{minipage}{0.19\textwidth}
{\includegraphics[width=\textwidth]{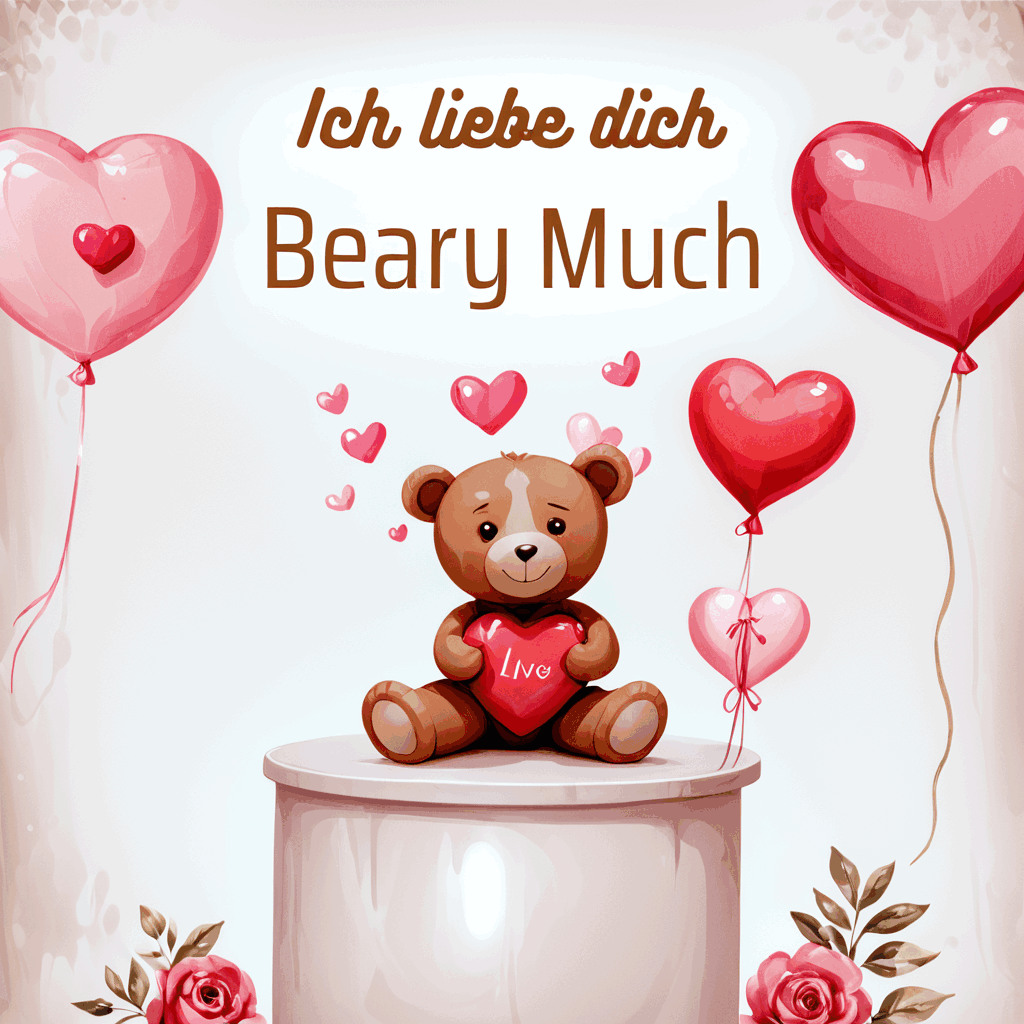}}
\vspace{-3mm}
\end{minipage}
\begin{minipage}{0.19\textwidth}
{\includegraphics[width=\textwidth]{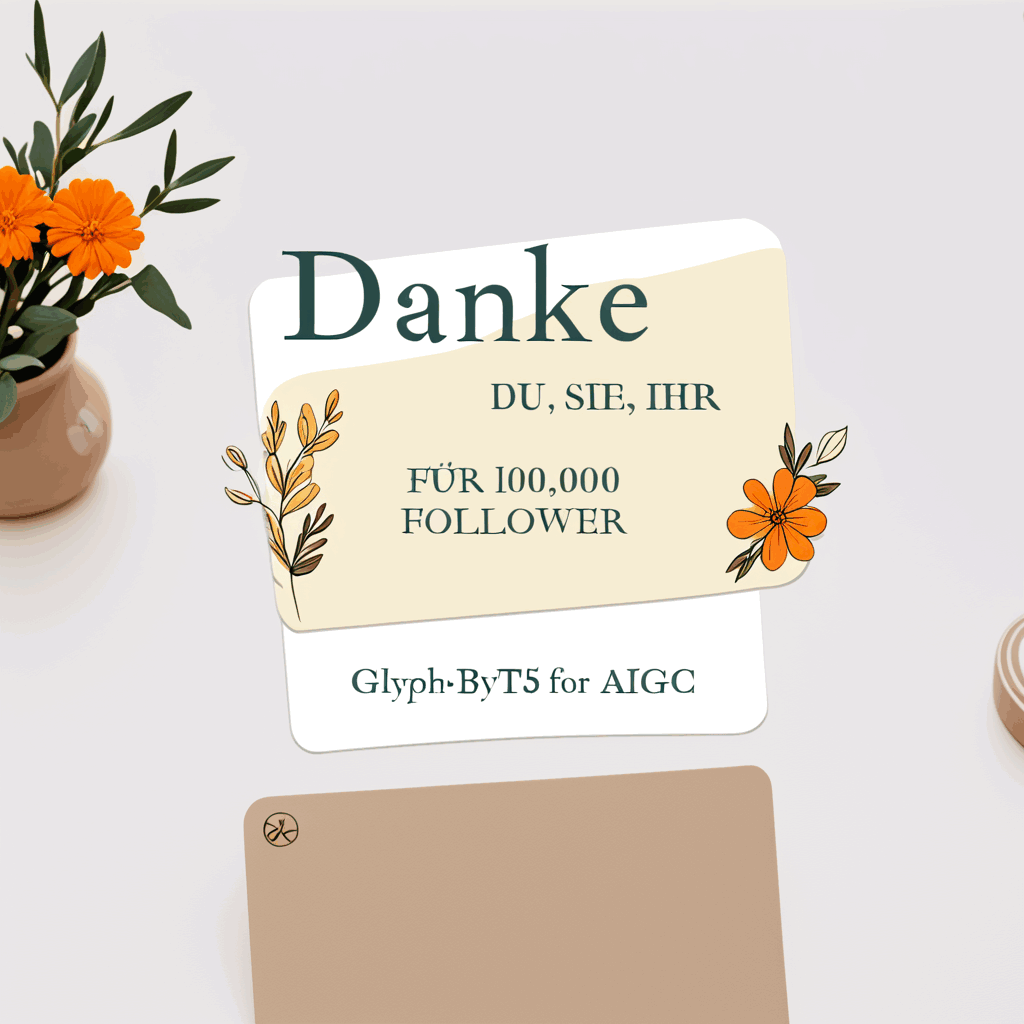}}
\vspace{-3mm}
\end{minipage}
\begin{minipage}{0.19\textwidth}
{\includegraphics[width=\textwidth]{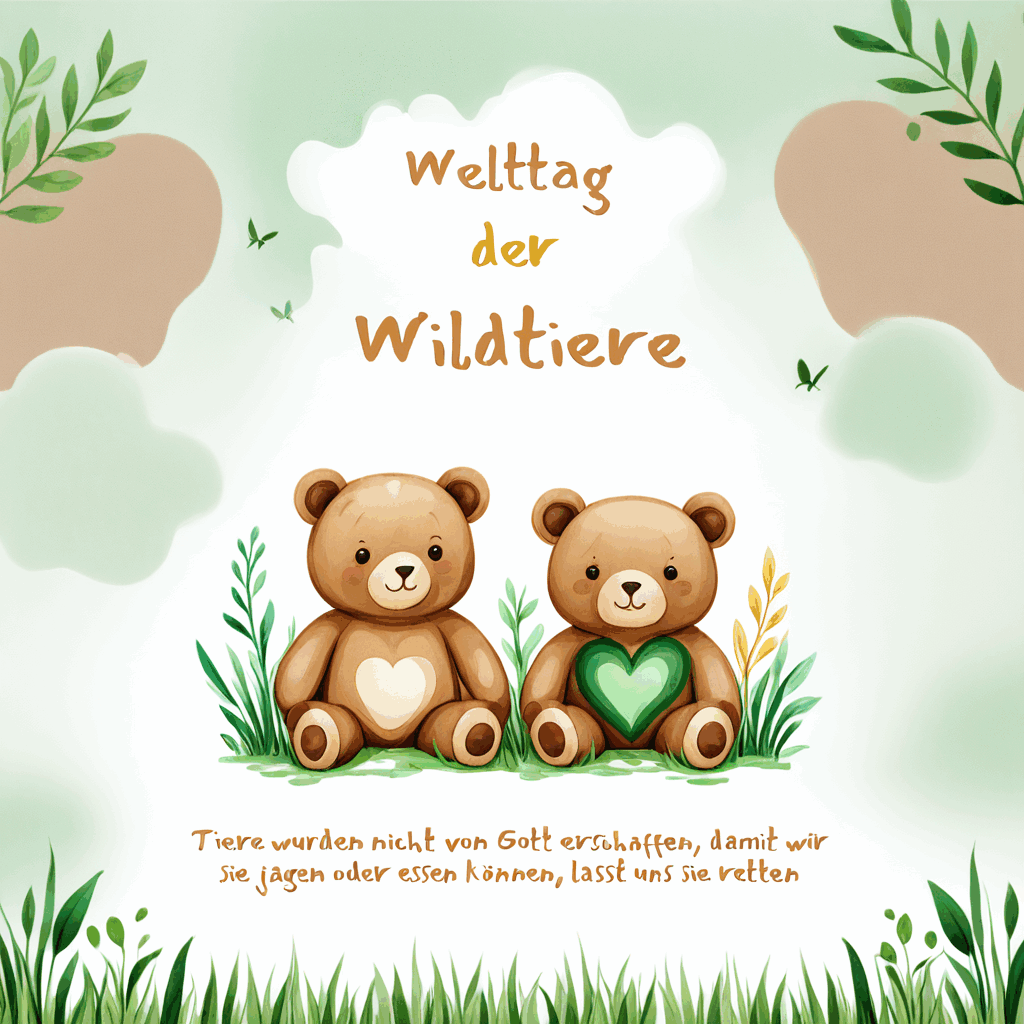}}
\vspace{-3mm}
\end{minipage}
\begin{minipage}{0.19\textwidth}
{\includegraphics[width=\textwidth]{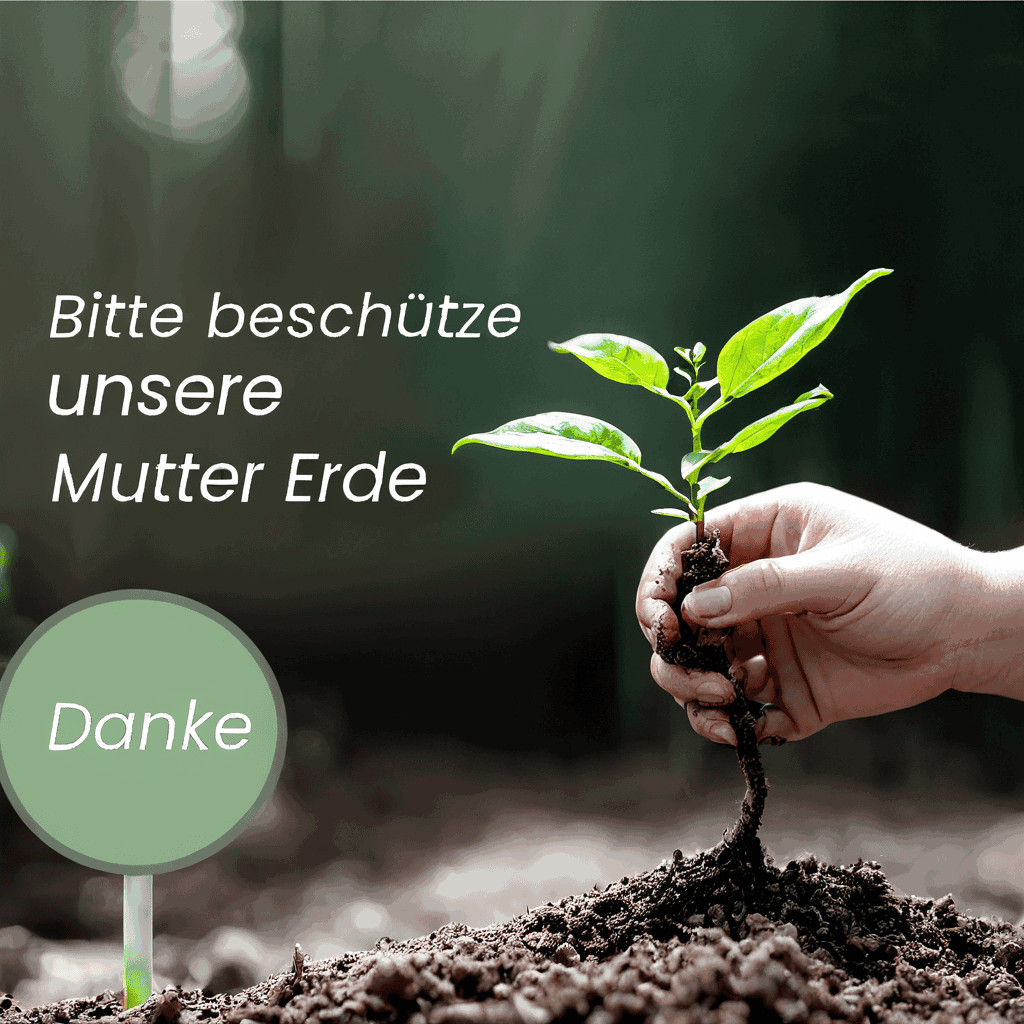}}
\vspace{-3mm}
\end{minipage}\\
\begin{minipage}{0.19\textwidth}
{\includegraphics[width=\textwidth]{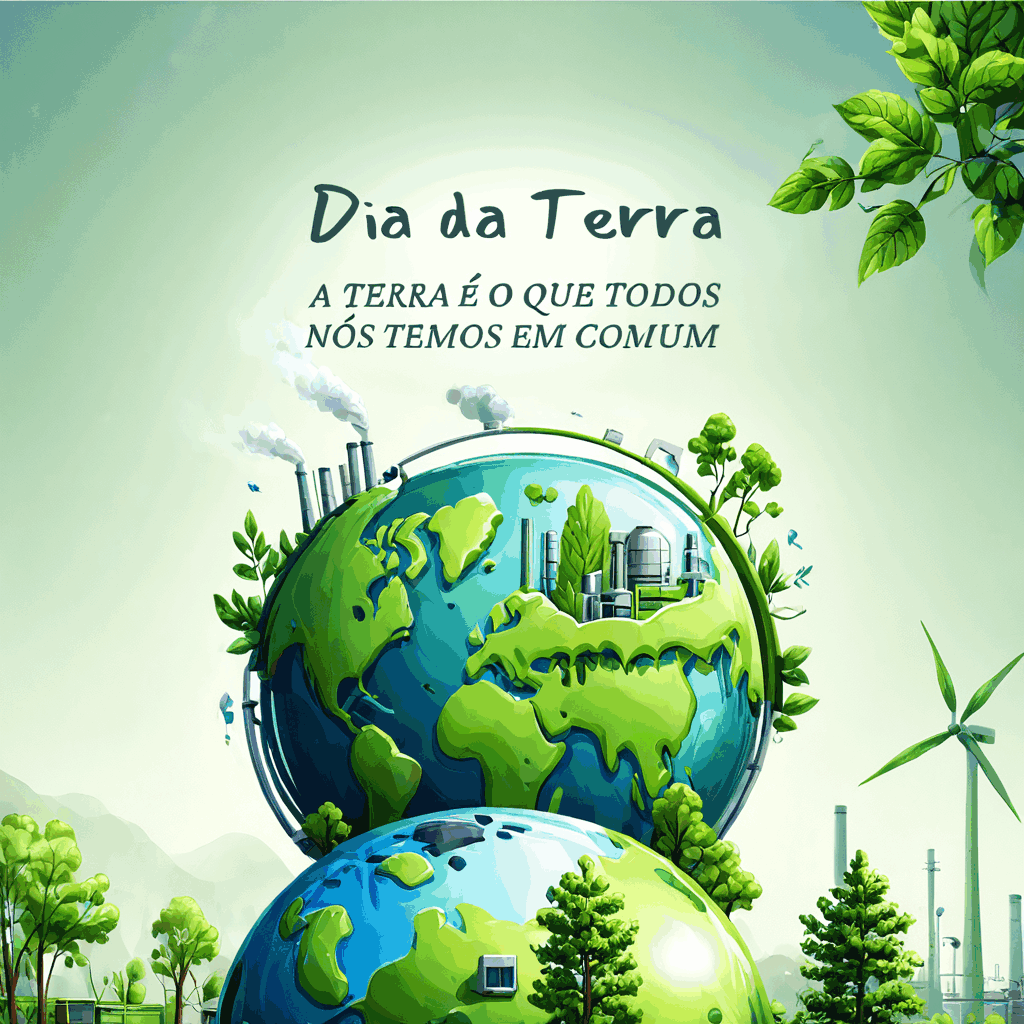}}
\vspace{-3mm}
\end{minipage}
\begin{minipage}{0.19\textwidth}
{\includegraphics[width=\textwidth]{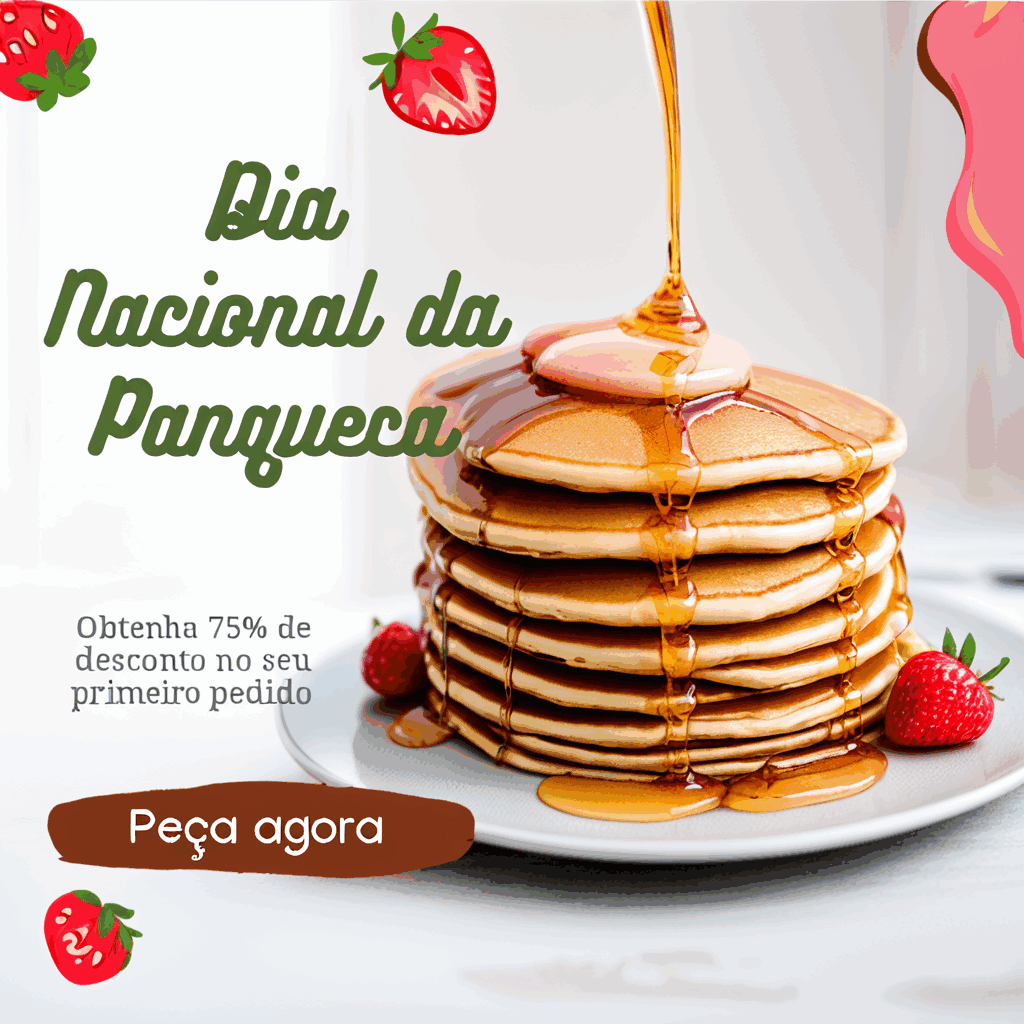}}
\vspace{-3mm}
\end{minipage}
\begin{minipage}{0.19\textwidth}
{\includegraphics[width=\textwidth]{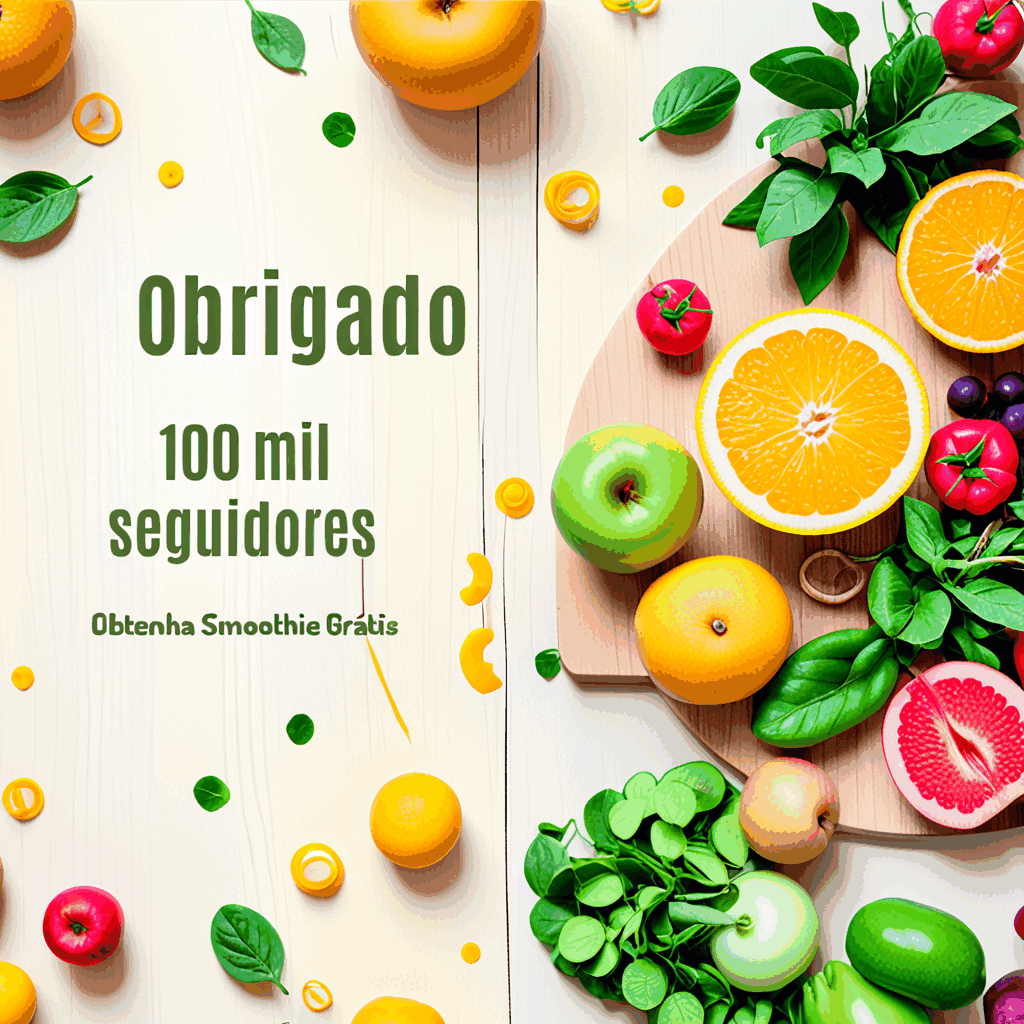}}
\vspace{-3mm}
\end{minipage}
\begin{minipage}{0.19\textwidth}
{\includegraphics[width=\textwidth]{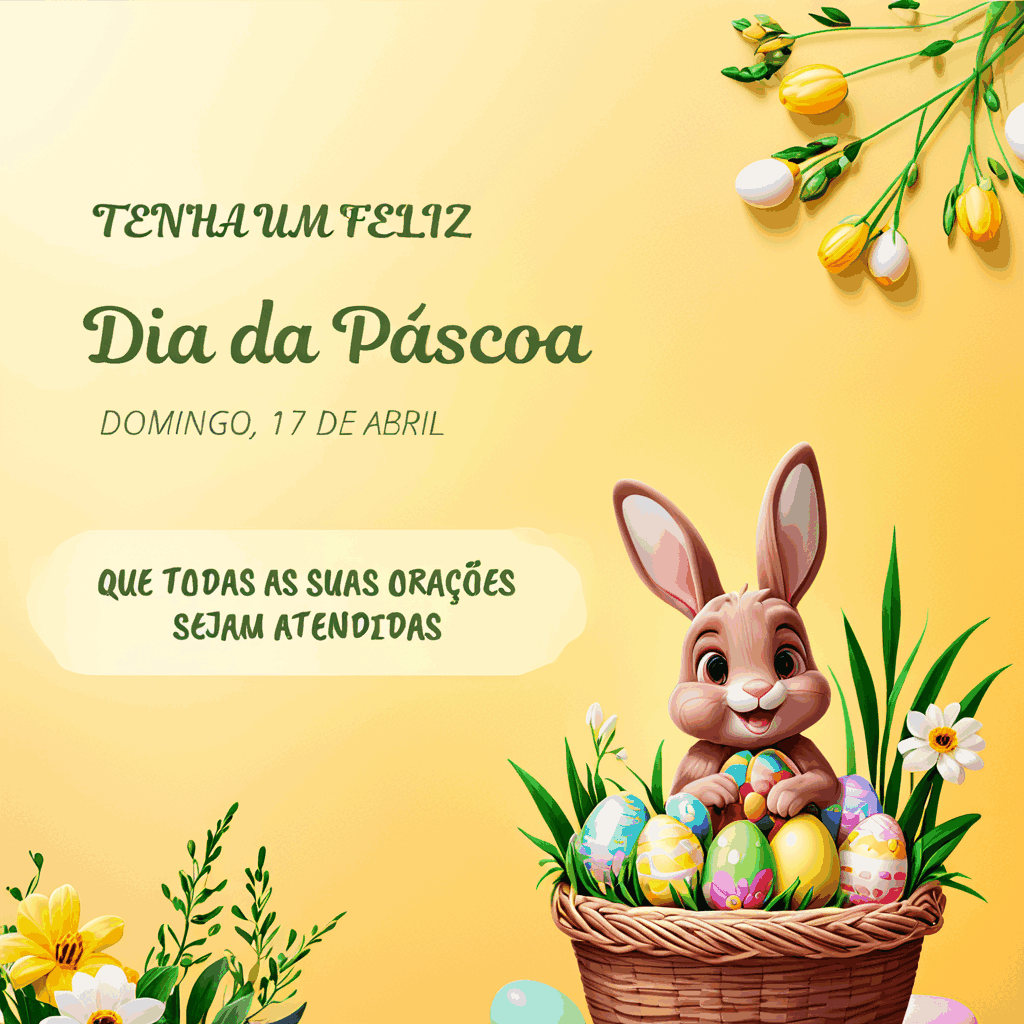}}
\vspace{-3mm}
\end{minipage}
\begin{minipage}{0.19\textwidth}
{\includegraphics[width=\textwidth]{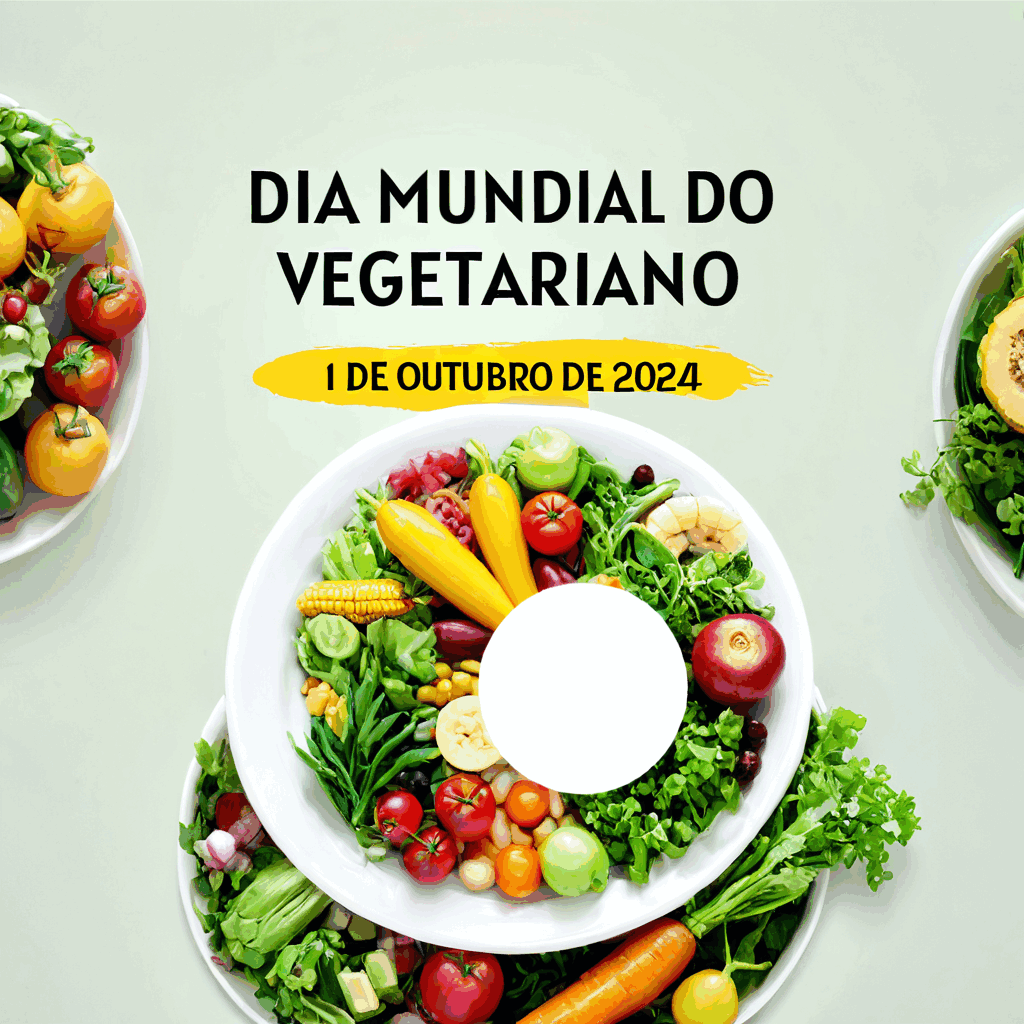}}
\vspace{-3mm}
\end{minipage}\\
\begin{minipage}{0.19\textwidth}
{\includegraphics[width=\textwidth]{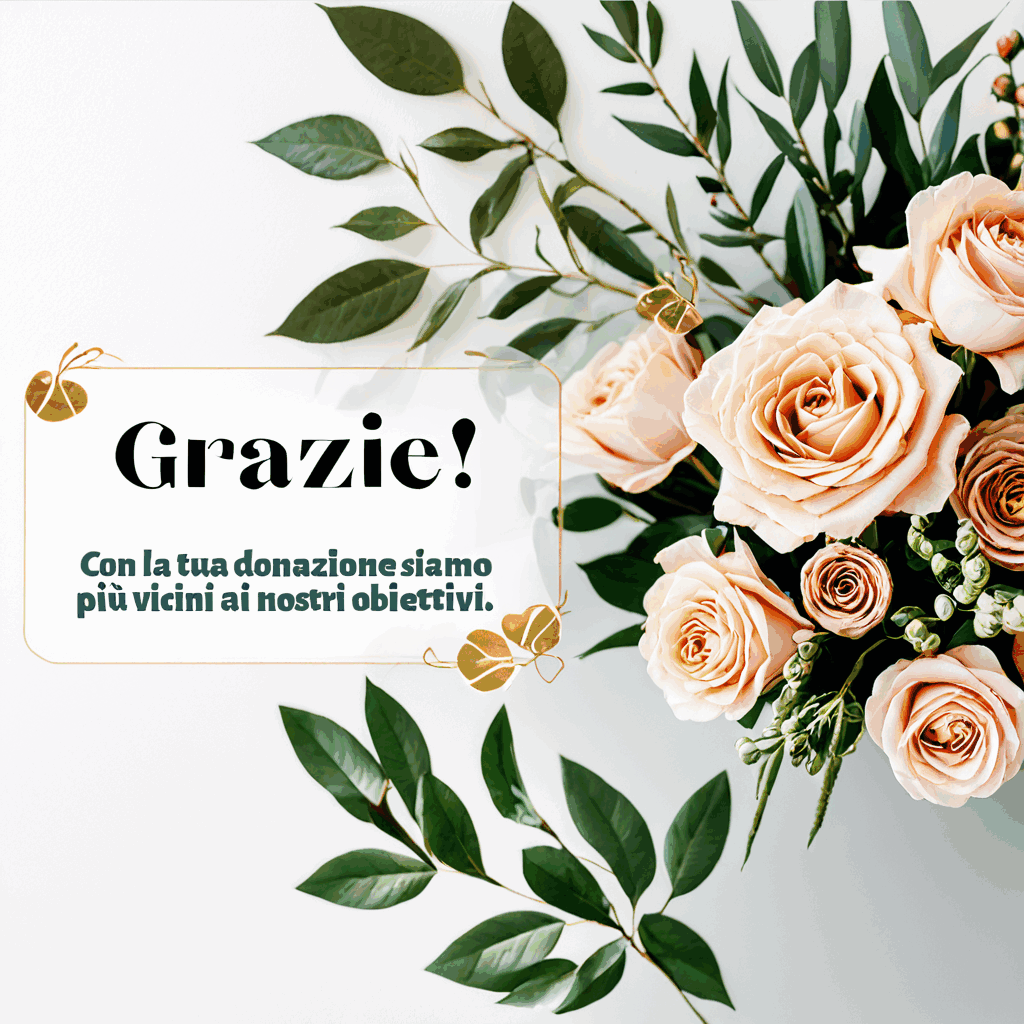}}
\vspace{-3mm}
\end{minipage}
\begin{minipage}{0.19\textwidth}
{\includegraphics[width=\textwidth]{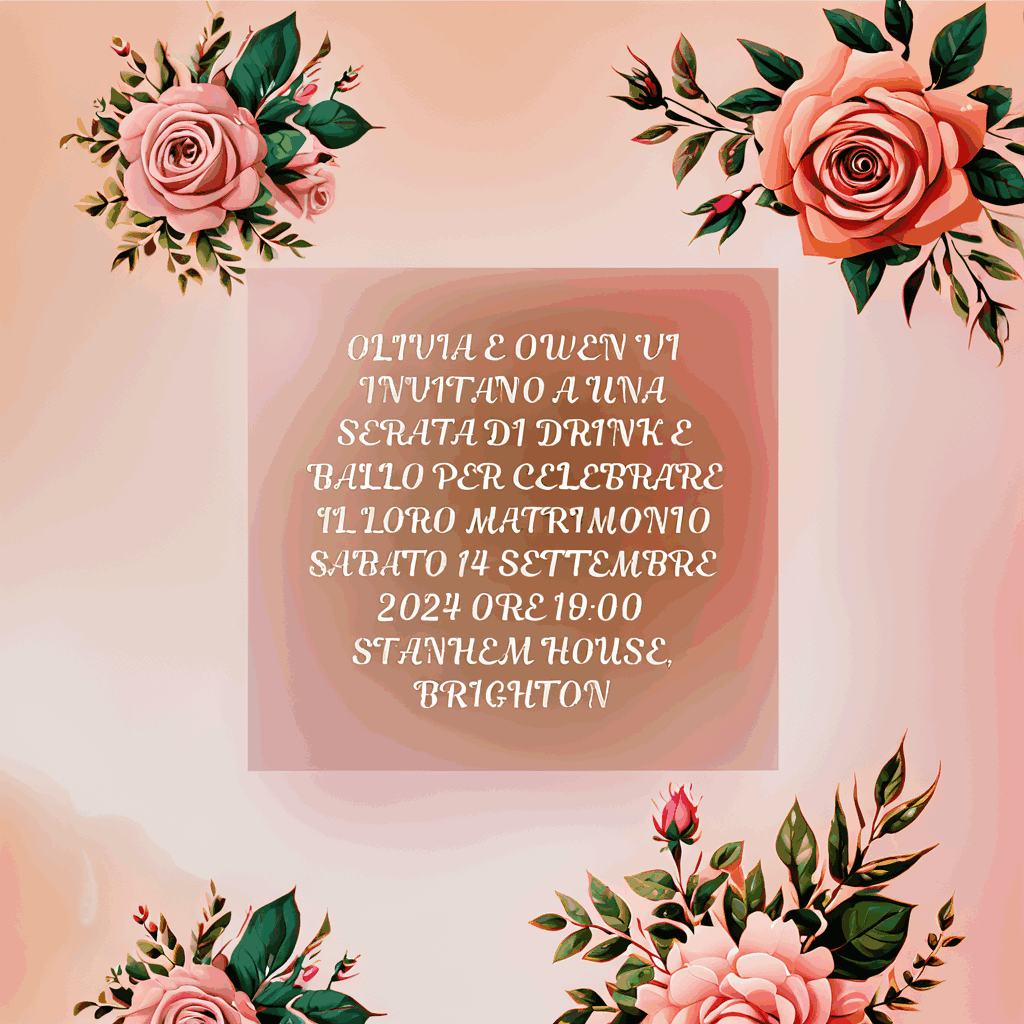}}
\vspace{-3mm}
\end{minipage}
\begin{minipage}{0.19\textwidth}
{\includegraphics[width=\textwidth]{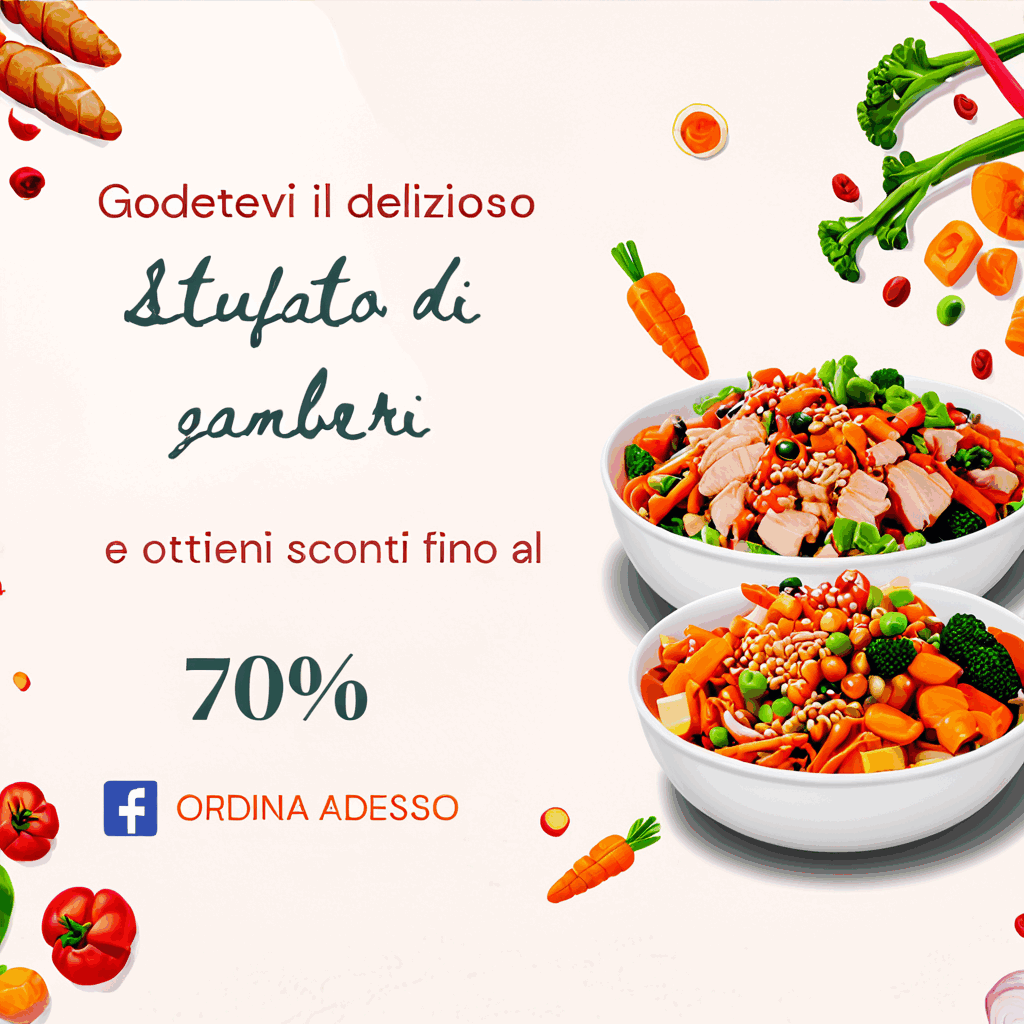}}
\vspace{-3mm}
\end{minipage}
\begin{minipage}{0.19\textwidth}
{\includegraphics[width=\textwidth]{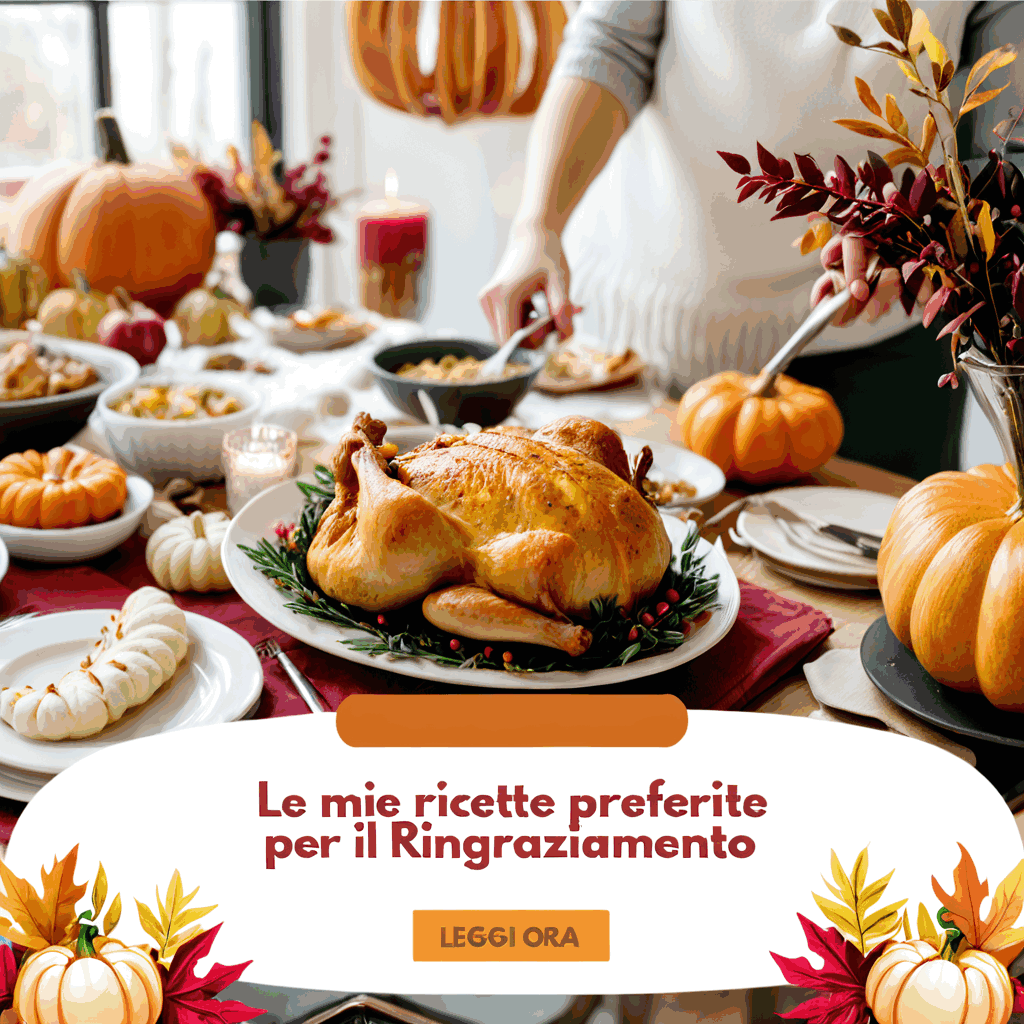}}
\vspace{-3mm}
\end{minipage}
\begin{minipage}{0.19\textwidth}
{\includegraphics[width=\textwidth]{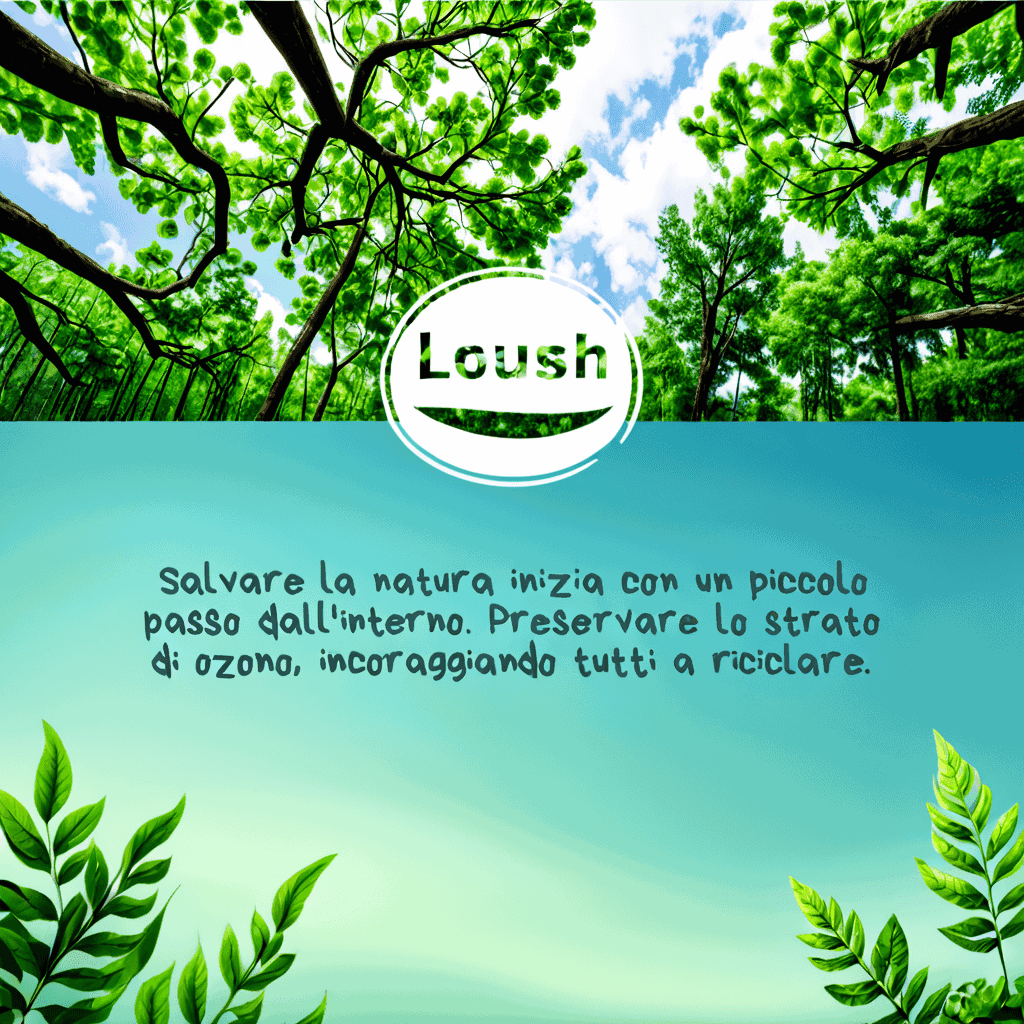}}
\vspace{-3mm}
\end{minipage}\\
\begin{minipage}{0.19\textwidth}
{\includegraphics[width=\textwidth]{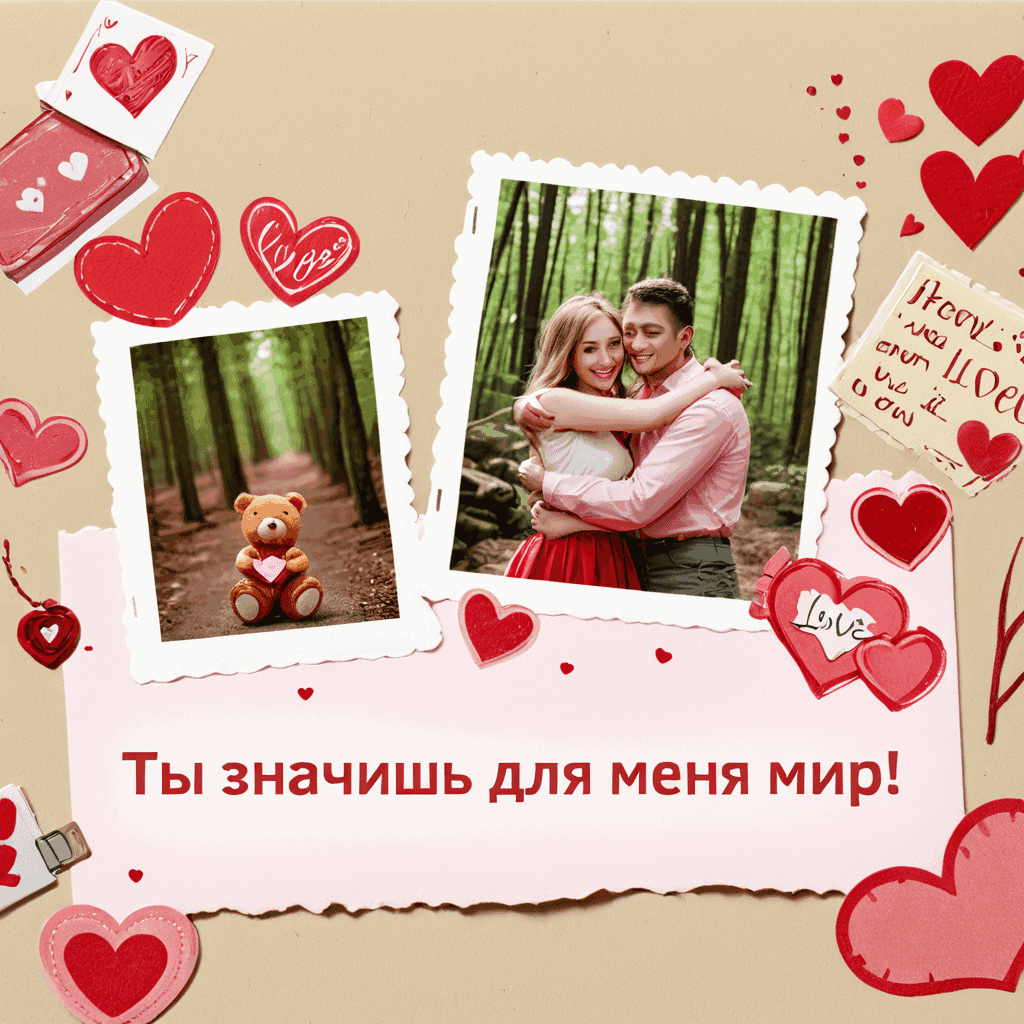}}
\vspace{-3mm}
\end{minipage}
\begin{minipage}{0.19\textwidth}
{\includegraphics[width=\textwidth]{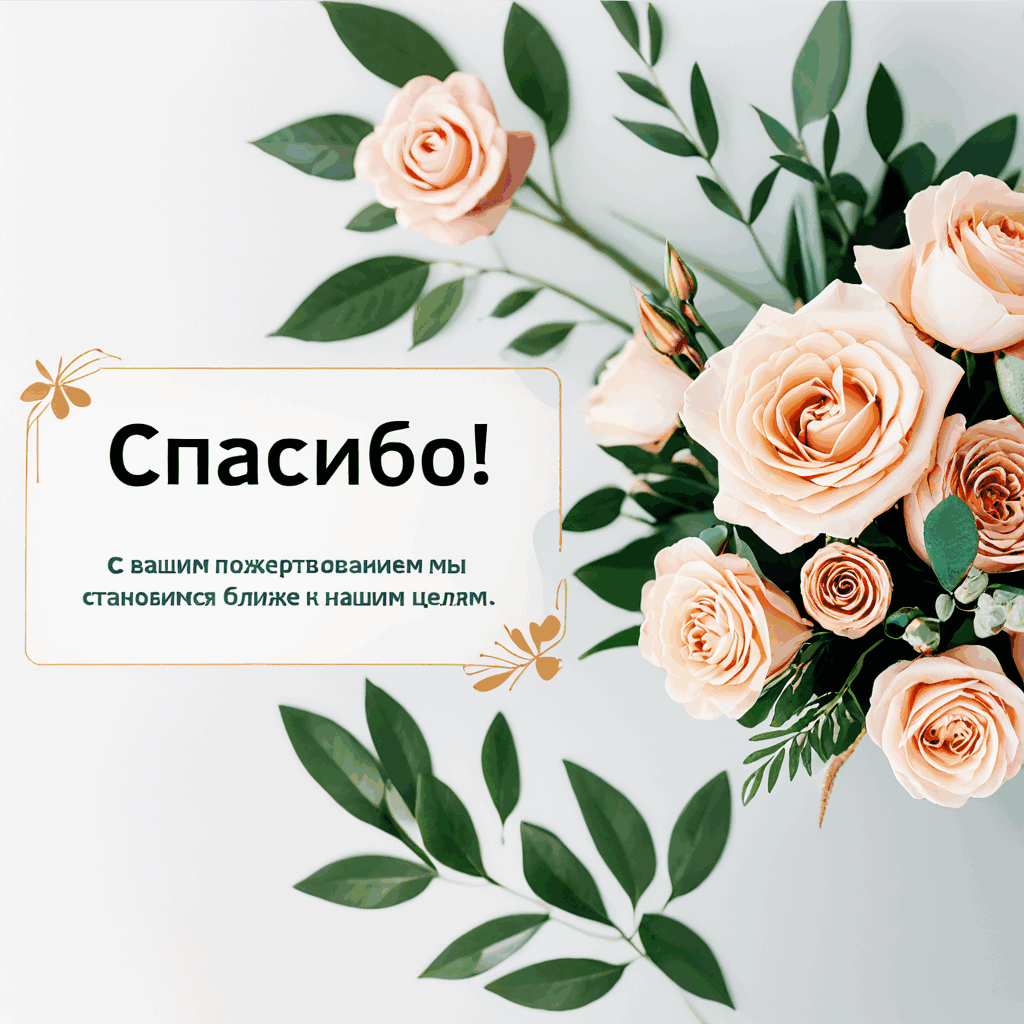}}
\vspace{-3mm}
\end{minipage}
\begin{minipage}{0.19\textwidth}
{\includegraphics[width=\textwidth]{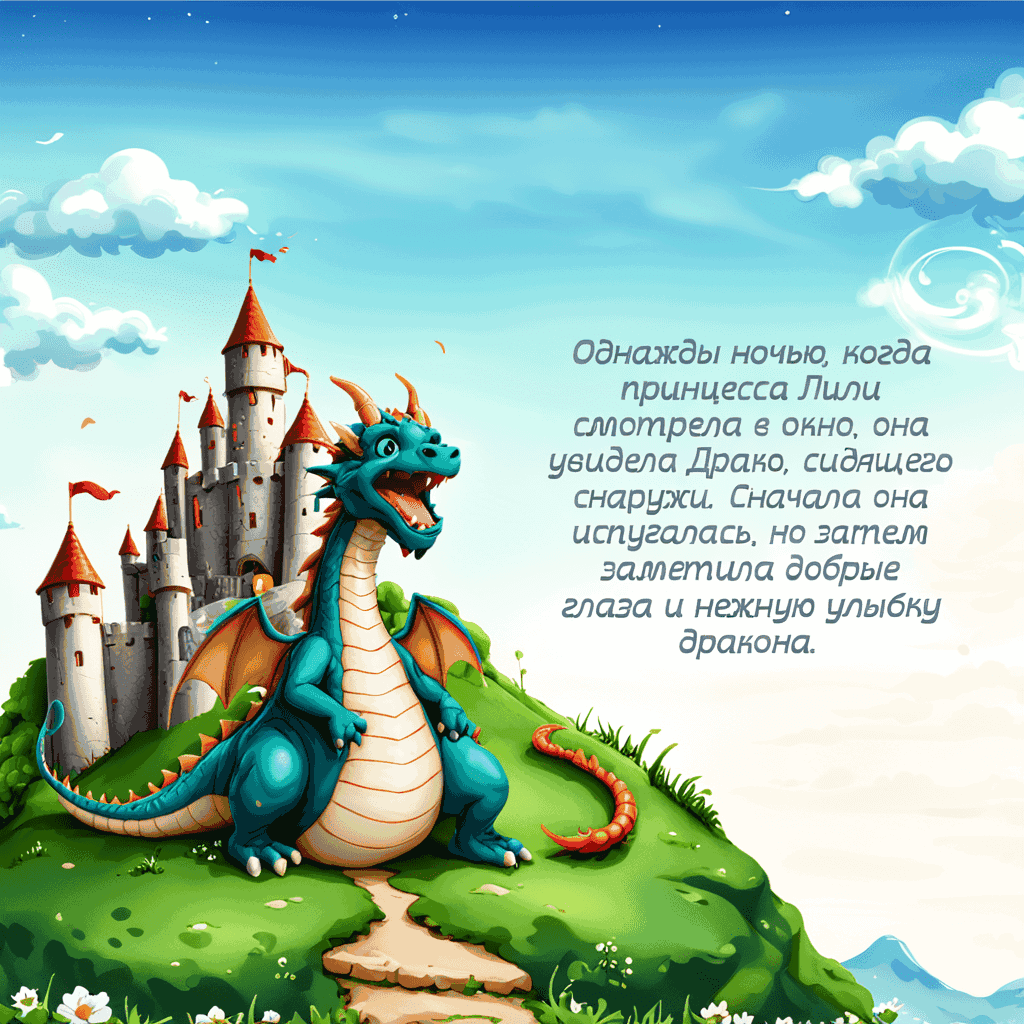}}
\vspace{-3mm}
\end{minipage}
\begin{minipage}{0.19\textwidth}
{\includegraphics[width=\textwidth]{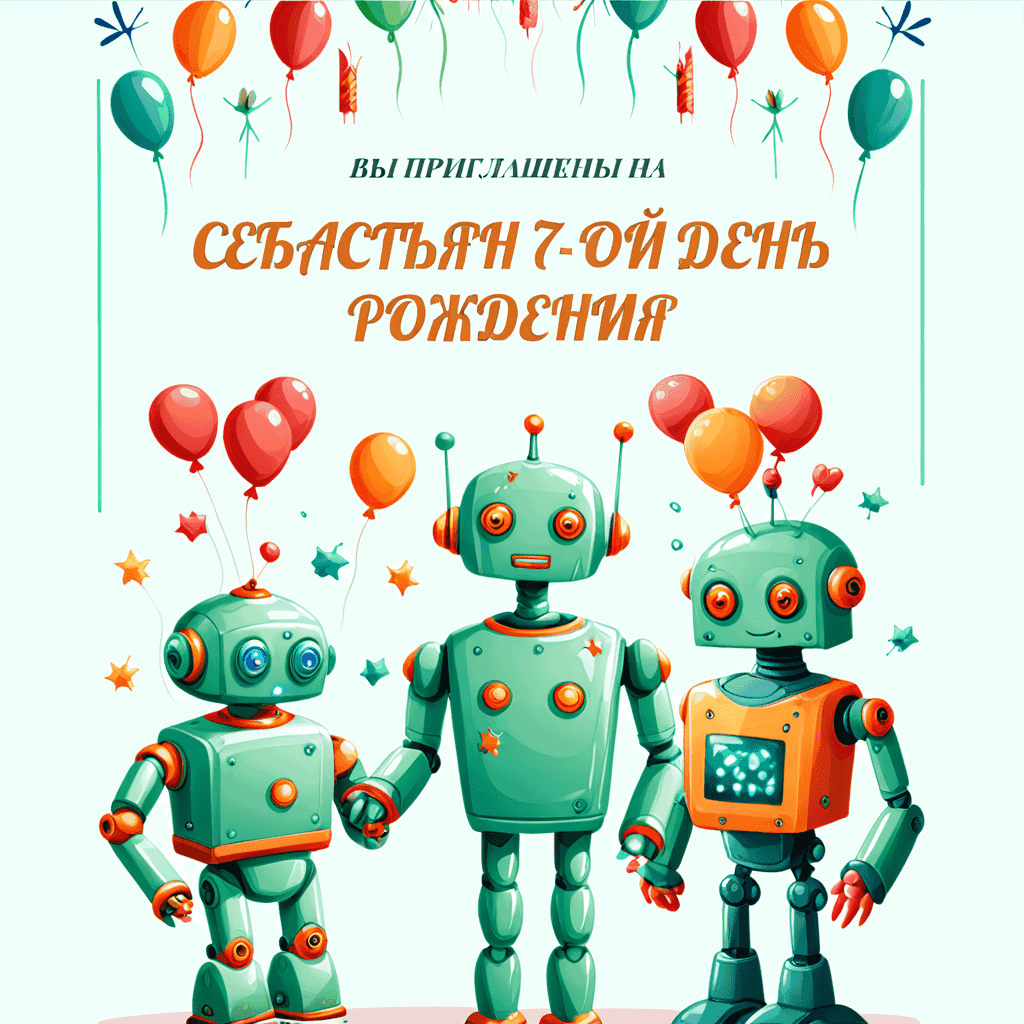}}
\vspace{-3mm}
\end{minipage}
\begin{minipage}{0.19\textwidth}
{\includegraphics[width=\textwidth]{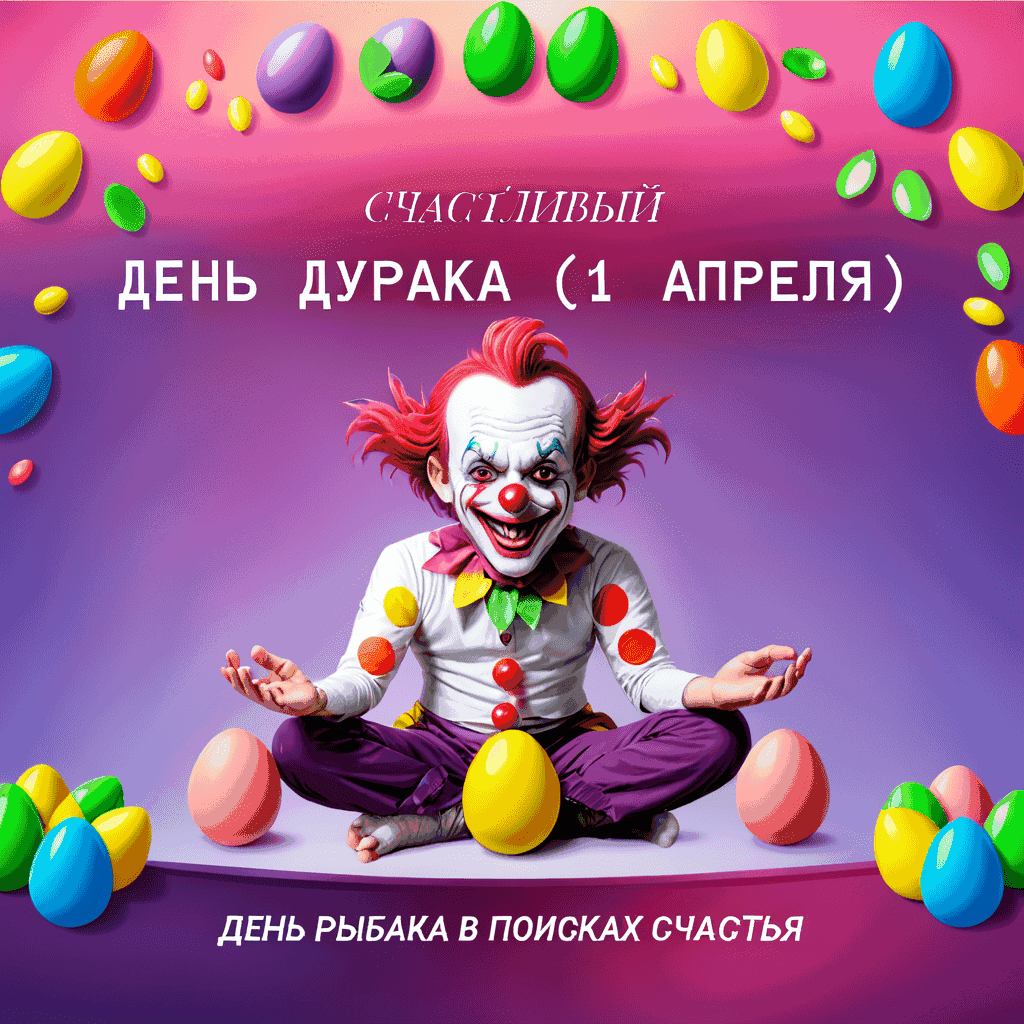}}
\vspace{-3mm}
\end{minipage}\\
\vspace{-3mm} 
,\captionof{figure}{\small{Illustrating the multilingual visual text rendering results with our approach. We show the German, Portuguese, Italian, and Russian visual text results in the 1st, 2nd, 3rd, and 4th rows, respectively.}}
\label{fig:teaser_2}
\end{figure*}

We address the multilingual visual text rendering task by building high-quality and scalable multilingual training datasets.The key challenge is that glyph images and graphic design images associated with other languages are much sparser and harder to collect while being more complex than those in English. To overcome this challenge, we design a simple translation-based approach to transform English glyph images and graphic design images into ones with other languages. We show a representative case in Figure~\ref{fig:multilingual_glyph_text}. {To ensure coherent layout arrangement across different languages, we keep the total number of characters close to the number of characters in English.}
With this design, we first create a multilingual paired text-glyph dataset consisting of $1$ million glyph images with black backgrounds to train a multilingual Glyph-ByT5-v2 text encoder. This encoder bridges the gap between multilingual glyph images and multilingual text prompts. Then, we create a multilingual graphic design dataset consisting of $10$ million images to train the multilingual Glyph-SDXL-v2.

In addition, we improve the visual aesthetics of the generated graphic design images by leveraging the latest step-aware preference optimization technique~\cite{liang2024step} and the albedo technique~\cite{albedo}. We empirically verified that SDXL fine-tuned with these post-training techniques can significantly enhance the visual aesthetics of the generated graphic design images.

To demonstrate the effectiveness of our approach, we not only construct a multilingual \textsc{VisualParagraphy} benchmark to assess visual spelling accuracy but also conduct a thorough user study to compare the overall quality of the graphic design images generated with \dalle and our Glyph-SDXL-v2.
As shown in Figure~\ref{fig:winrate_spo_sdxl_vs_sdxl},
the graphic design images generated with Glyph-SDXL-v2 are preferred $63.7\%$ of the time over those generated with the previous Glyph-SDXL in terms of visual aesthetics while ensuring closely accurate visual spelling performance.
We also have reported the OCR precision of our approach on multilingual \textsc{VisualParagraphy} benchmark in Table~\ref{tab:teaser_table}.
In addition, we illustrate the generated multilingual visual text results in Figure~\ref{fig:teaser} and Figure~\ref{fig:teaser_2}.
we expect our approach to provide a strong aesthetic baseline for the accurate multilingual visual text rendering task and inspire more efforts to push progress along this path.

\begin{figure*}[t]
\centering
\begin{minipage}[t]{0.235\textwidth}
\begin{subfigure}[t]{\textwidth}
{\includegraphics[width=\textwidth]{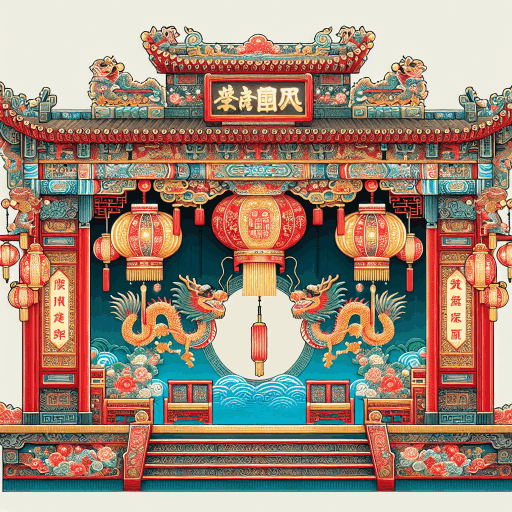}}
{\includegraphics[width=\textwidth]{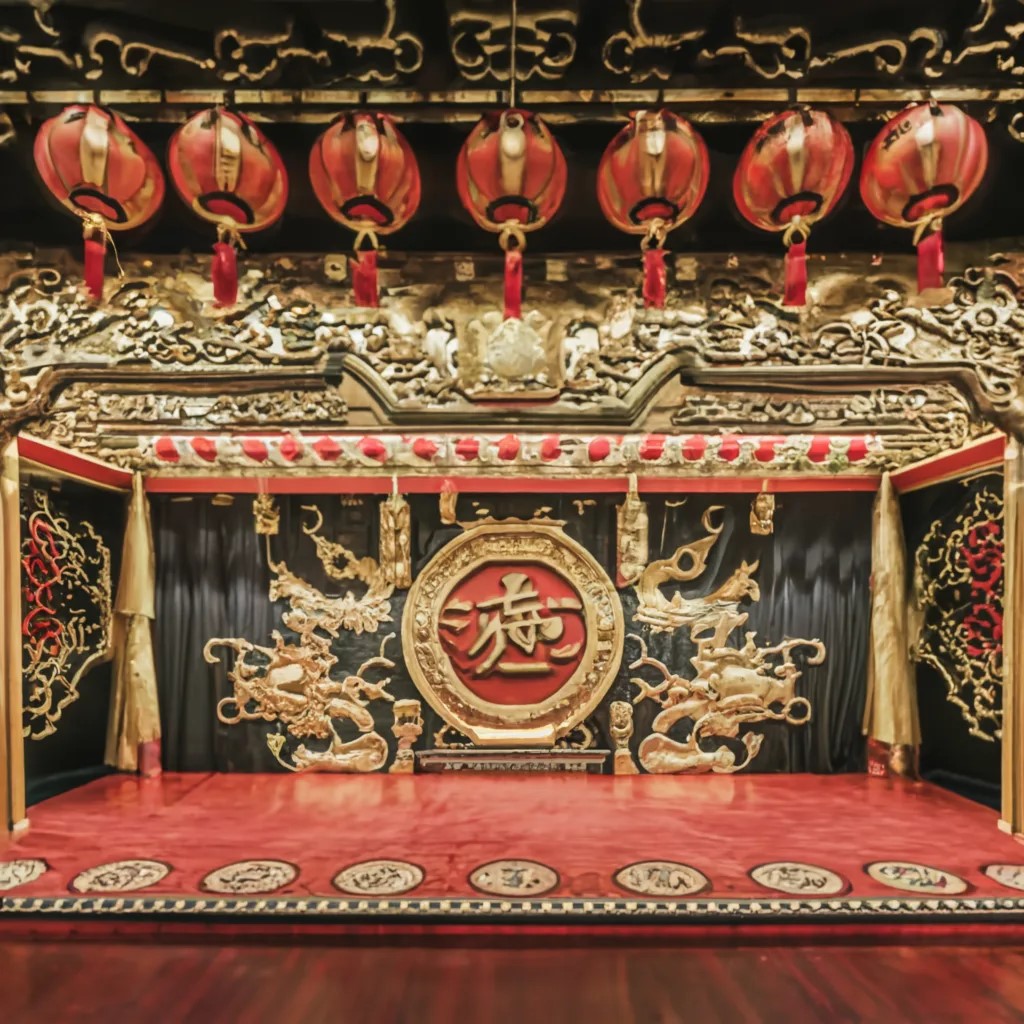}}
\caption{Chinese: \redtext{"\begin{CJK}{UTF8}{gbsn}根据国家法定假期的规定，现将有关放假事宜通知如下：2024年元旦1月1日至1月3日放假，共3天，特此通知。\end{CJK}", "\begin{CJK}{UTF8}{gbsn}预祝大家假期愉快！\end{CJK}", "\begin{CJK}{UTF8}{gbsn}放假通知\end{CJK}"}, OCR precision \dalle: 0\%, \ideogram: 0\%.}
\end{subfigure}
\end{minipage}
\begin{minipage}[t]{0.235\textwidth}
\begin{subfigure}[t]{\textwidth}
{\includegraphics[width=\textwidth]{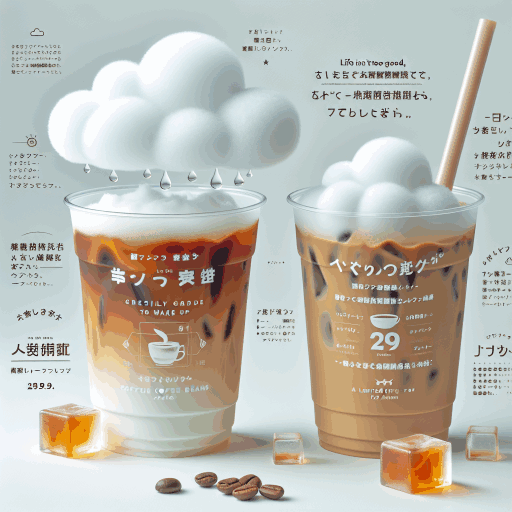}}
{\includegraphics[width=\textwidth]{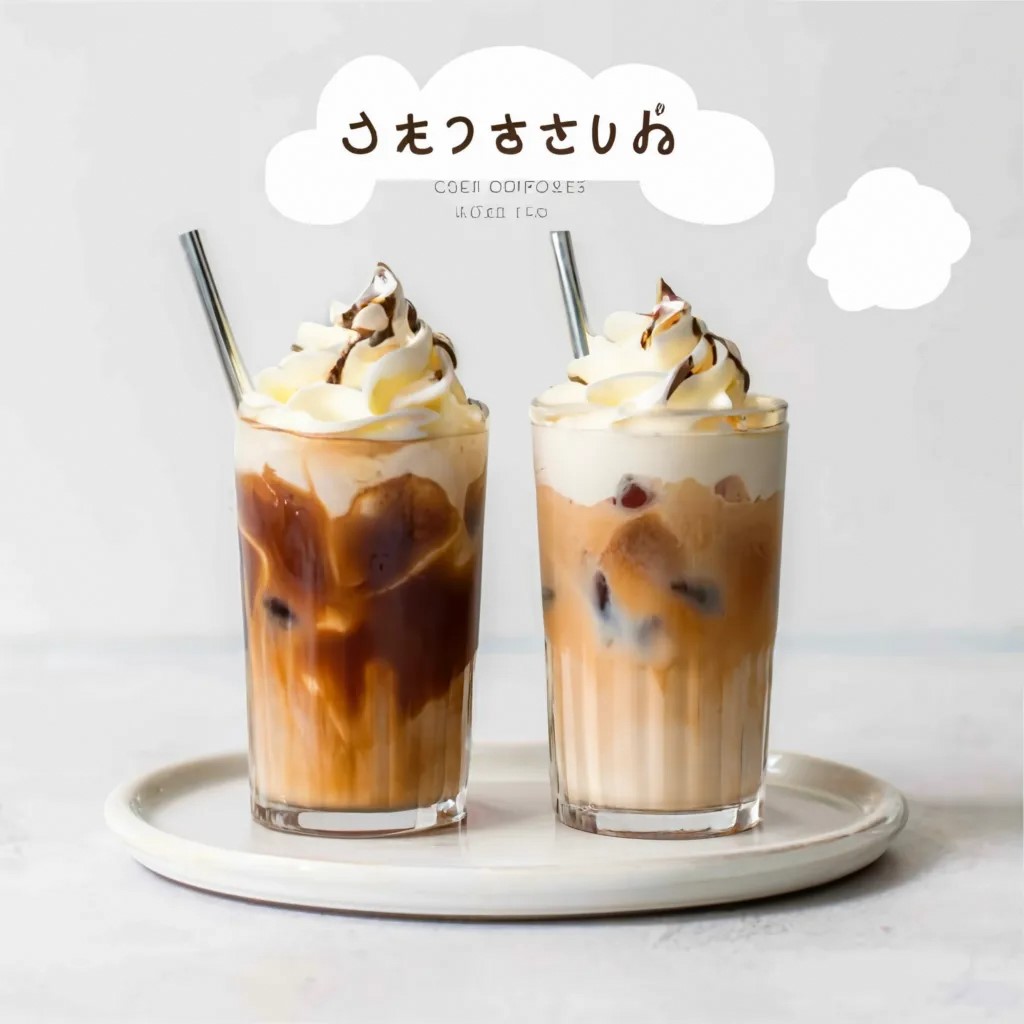}}
\caption{Japanese: \redtext{"\begin{CJK}{UTF8}{min}早起は困難な人々 目を覚ますために助けが必要です\end{CJK}", "\begin{CJK}{UTF8}{min}毎日は二度目のチャンスです 一日一日を大切にするために\end{CJK}", "\begin{CJK}{UTF8}{min}人生は良すぎず、コーヒーは悪すぎません\end{CJK}", "\begin{CJK}{UTF8}{min}29.9/カップ \end{CJK}", "\begin{CJK}{UTF8}{min}挽きたてのコーヒー豆\end{CJK}", "\begin{CJK}{UTF8}{min}{秋季限定}\end{CJK}", "\begin{CJK}{UTF8}{min}生活は苦しいですが カフェラテを選びましょう\end{CJK}"}, OCR precision \dalle: 5.1\%, \ideogram: 0\%.}
\end{subfigure}
\end{minipage}
\begin{minipage}[t]{0.235\textwidth}
\begin{subfigure}[t]{\textwidth}
{\includegraphics[width=\textwidth]{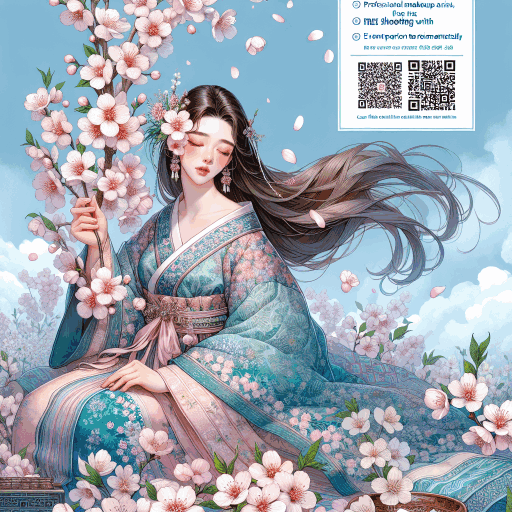}}
{\includegraphics[width=\textwidth]{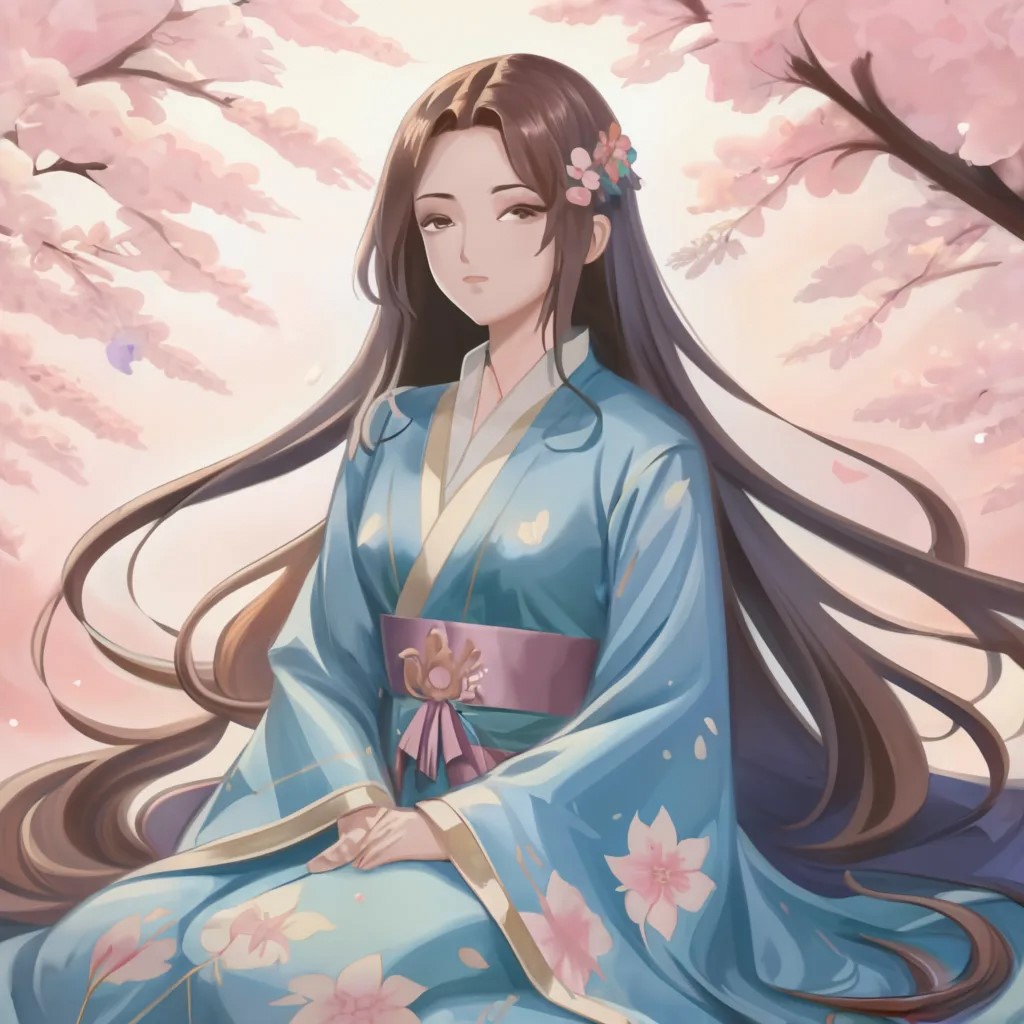}}
\caption{Korean: \redtext{"\begin{CJK}{UTF8}{mj}QR 코드를 스캔하여 즉시 등록하세요\end{CJK}", "\begin{CJK}{UTF8}{mj}전문 메이크업 아티스트 아름다운 한복 무료 촬영\end{CJK}", "\begin{CJK}{UTF8}{mj}【행사 기간】 5월 6일-5월 8일 【행사 장소】 상사호 고전 마을\end{CJK}", "\begin{CJK}{UTF8}{mj}한복 동호회\end{CJK}", "\begin{CJK}{UTF8}{mj}한복 체험 / 국조 문화 창작전\end{CJK}"}, OCR precision \dalle: 0\%, \ideogram: 0\%.}
\end{subfigure}
\end{minipage}
\begin{minipage}[t]{0.235\textwidth}
\begin{subfigure}[t]{\textwidth}
{\includegraphics[width=\textwidth]{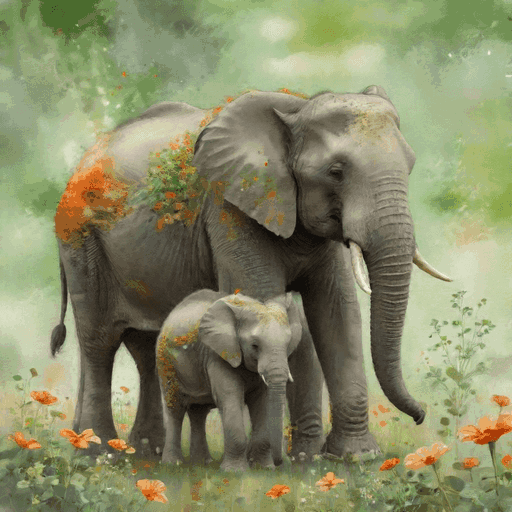}}
{\includegraphics[width=\textwidth]{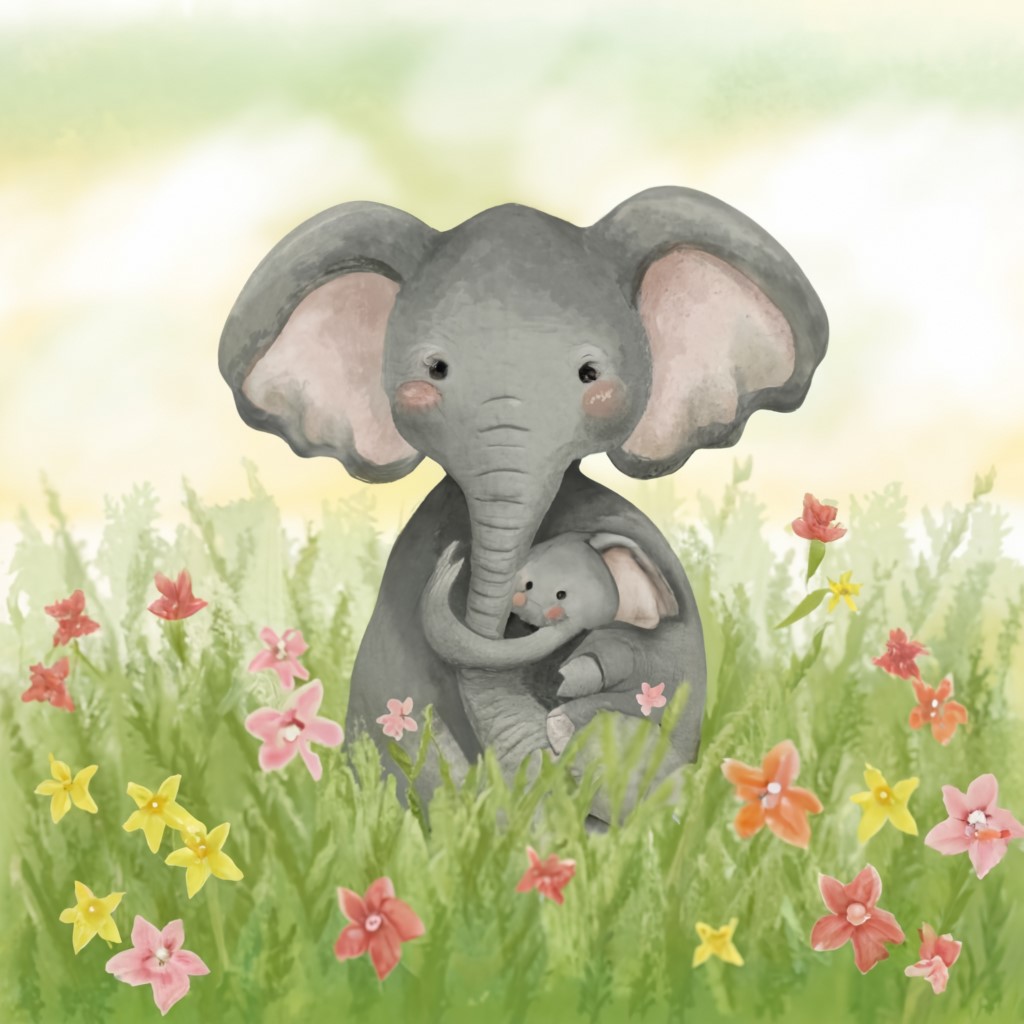}}
\caption{German: \redtext{"Sophia Davis", "Babyparty", "Bitte schließen Sie sich uns an für eine", "Zu Ehren", "23. November 2024 | 15:00 Uhr Fauget Hotels", "RSVP an +123-456-7890"}, OCR precision \dalle: 0\%, \ideogram: 0\%.}
\end{subfigure}
\end{minipage}\\
\begin{minipage}[t]{0.235\textwidth}
\begin{subfigure}[t]{\textwidth}
{\includegraphics[width=\textwidth]{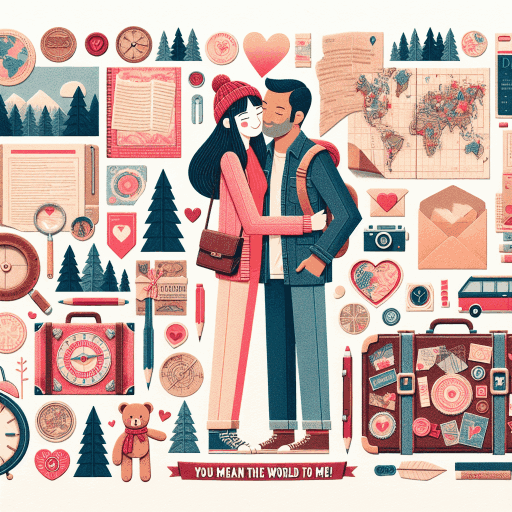}}
{\includegraphics[width=\textwidth]{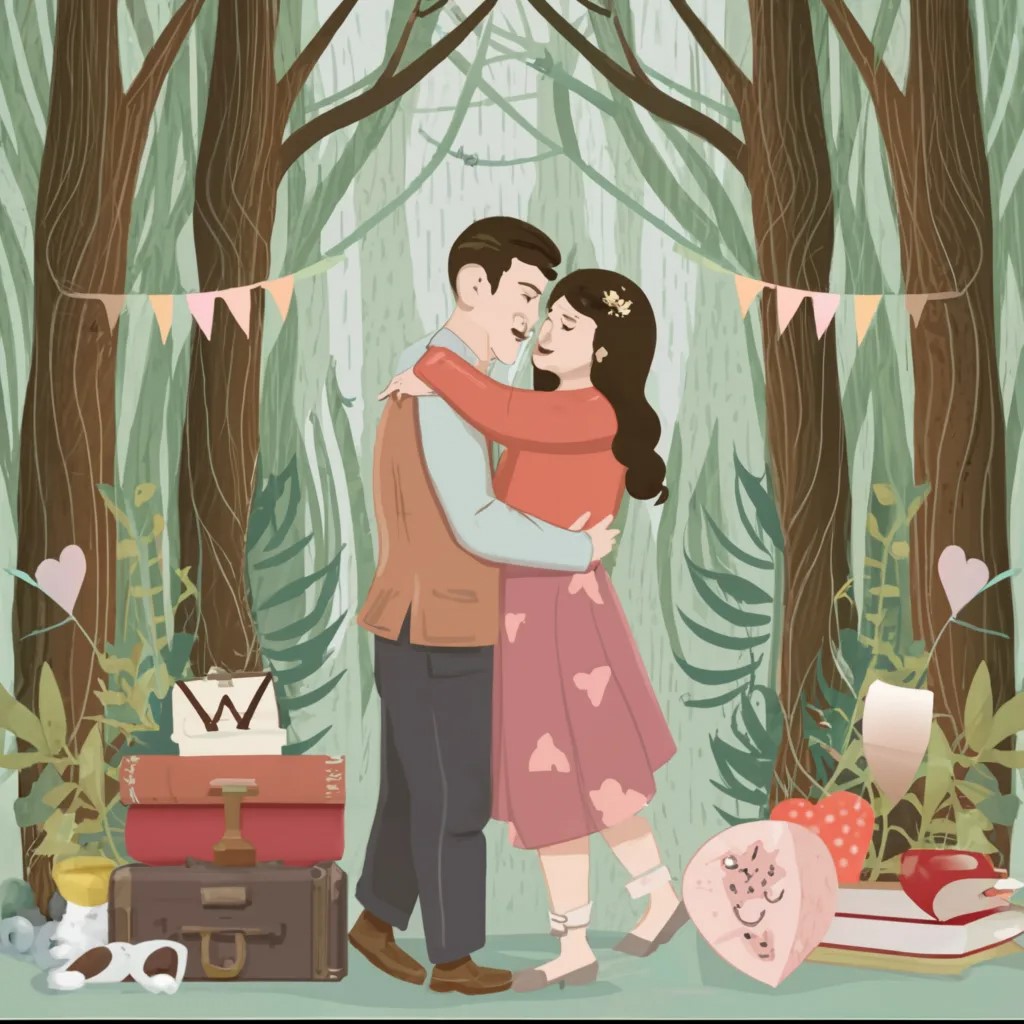}}
\caption{Russian: \redtext{"\foreignlanguage{russian}{Ты значишь для меня мир!}"}, OCR precision \dalle: 0\%, \ideogram: 0\%.}
\end{subfigure}
\end{minipage}
\begin{minipage}[t]{0.235\textwidth}
\begin{subfigure}[t]{\textwidth}
{\includegraphics[width=\textwidth]{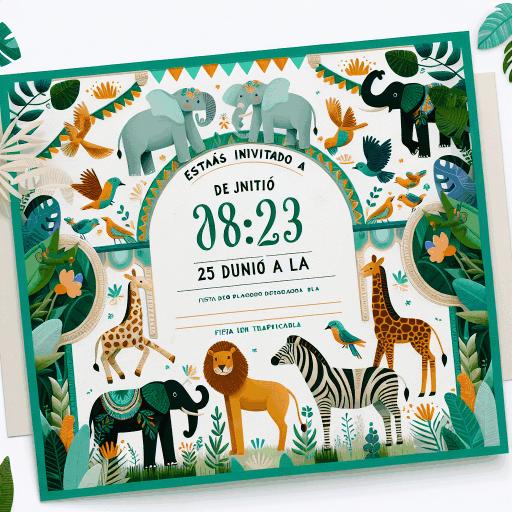}}
{\includegraphics[width=\textwidth]{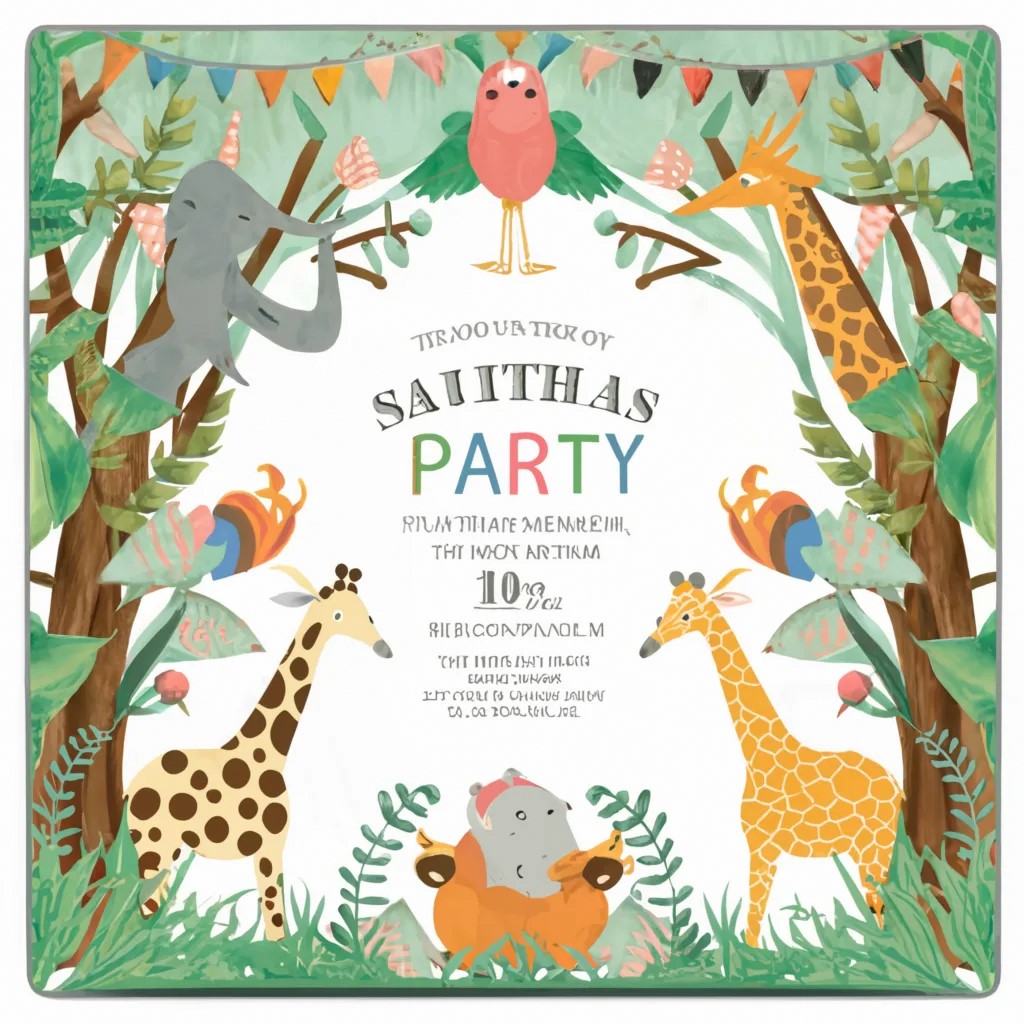}}
\caption{Spanish: \redtext{"08:00AM", "25 DE JUNIO DE 2024", "Estás invitado a", "FIESTA EN LA", "SELVA"}, OCR precision \dalle: 21.1\%, \ideogram: 0\%.}
\end{subfigure}
\end{minipage}
\begin{minipage}[t]{0.235\textwidth}
\begin{subfigure}[t]{\textwidth}
{\includegraphics[width=\textwidth]{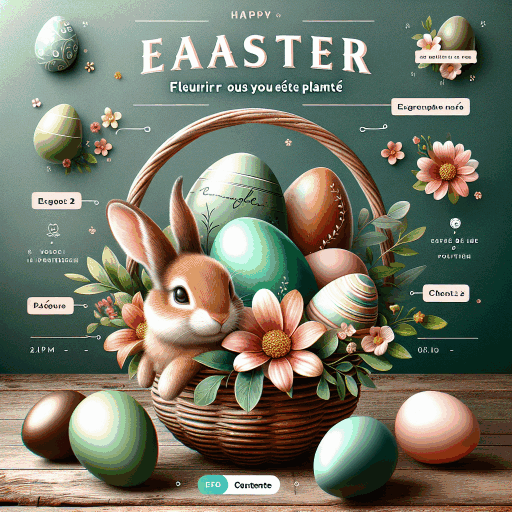}}
{\includegraphics[width=\textwidth]{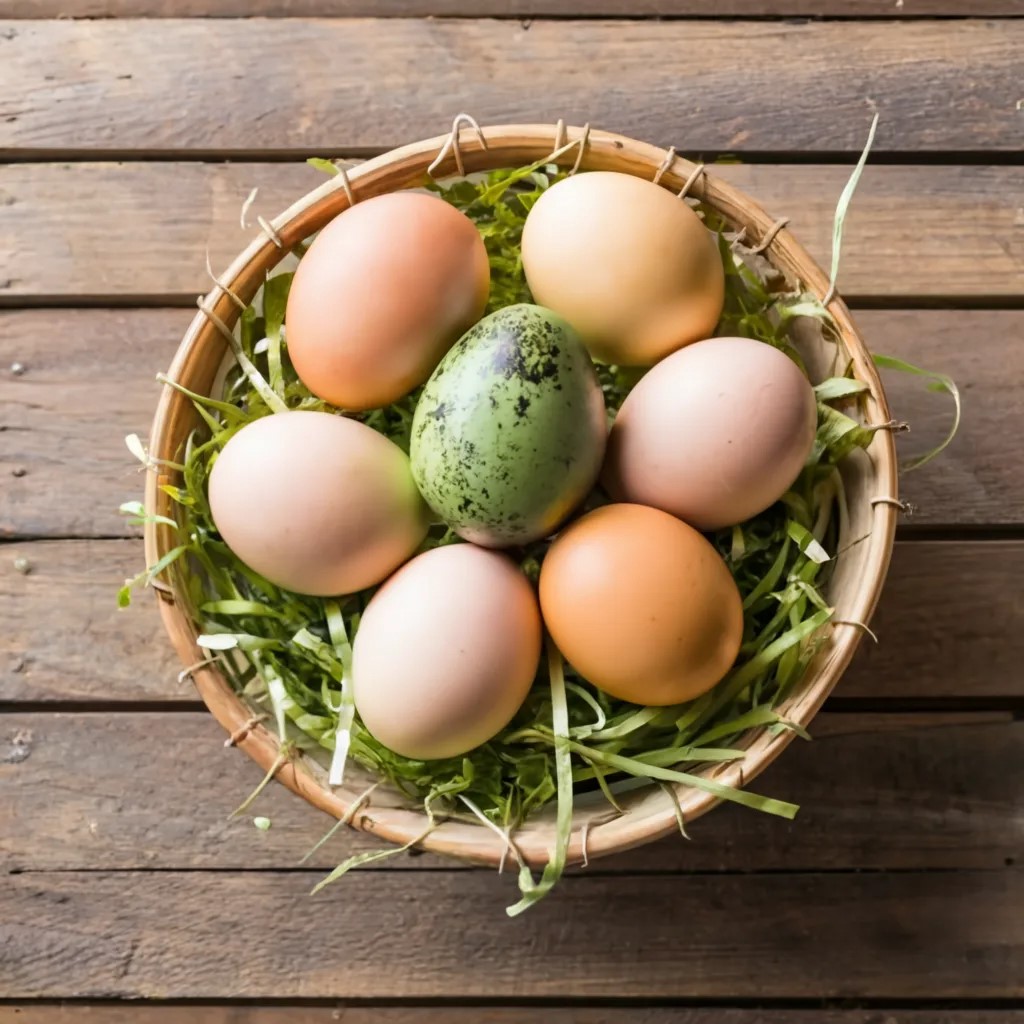}}
\caption{French: \redtext{"fleurir où vous êtes planté", "CONTENT.E", "PÂQUES"}, \dalle OCR precision: 0\%, \ideogram OCR precision: 0\%.}
\end{subfigure}
\end{minipage}
\begin{minipage}[t]{0.235\textwidth}
\begin{subfigure}[t]{\textwidth}
{\includegraphics[width=\textwidth]{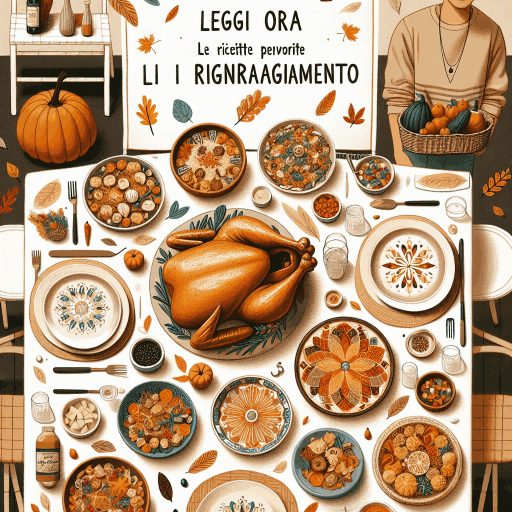}}
{\includegraphics[width=\textwidth]{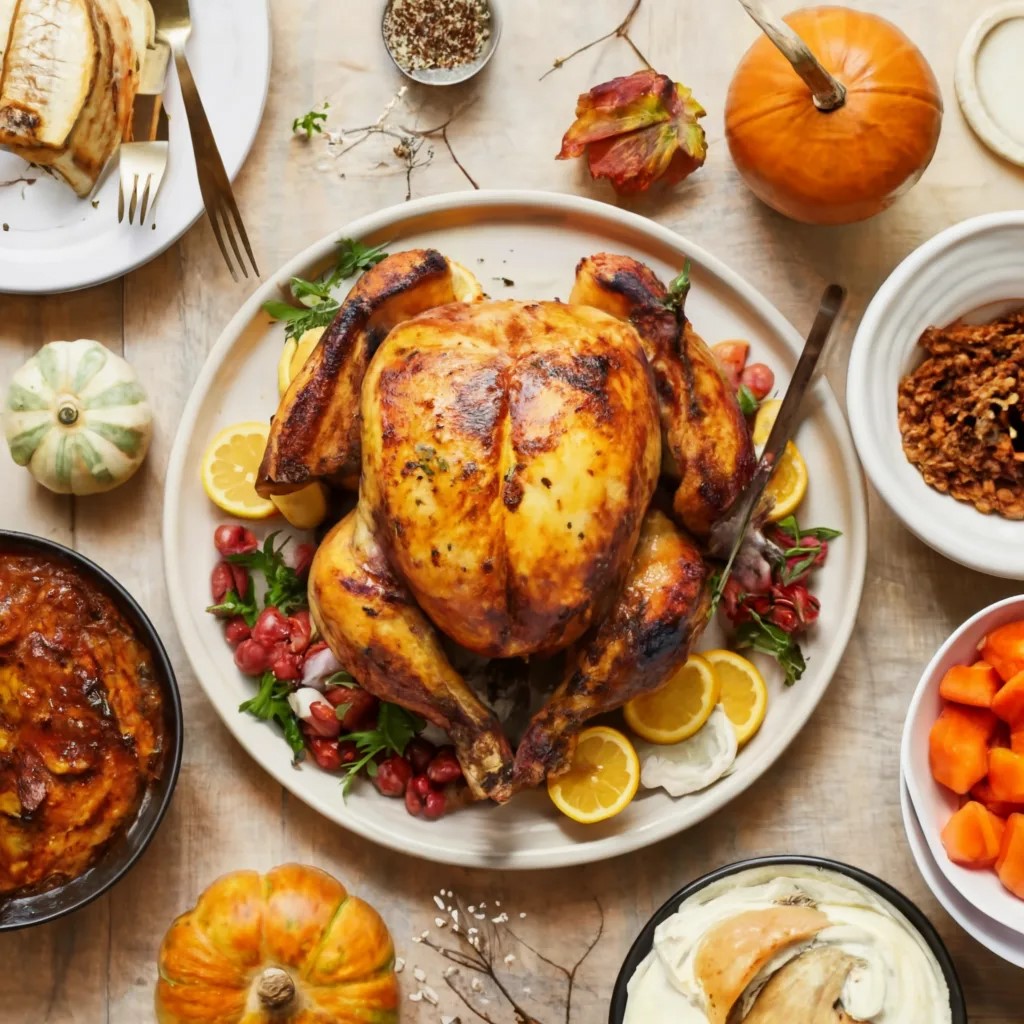}}
\caption{Italian: \redtext{"LEGGI ORA", "Le mie ricette preferite per il Ringraziamento"}, OCR precision \dalle: 37.5\%, \ideogram, 0\%.}
\end{subfigure}
\end{minipage}
\captionof{figure}{\small{Visualization of multilingual generation results by \dalle and \ideogram.}}
\vspace{-5mm}
\label{fig:dalle_ideogram}
\end{figure*}

\section{Related Work}
\label{sec:related_work}

Most existing efforts~\cite{yang2024glyphcontrol,ma2023glyphdraw,Liu2022CharacterAwareMI,chen2023textdiffuser,chen2023textdiffuser2,liu2024glyph} focus on accurate visual text rendering of English. AnyText~\cite{tuo2023anytext} further shows very modest visual text rendering results for other languages like Chinese, Japanese, and Korean due to the difficulty of collecting high-quality data, training the models on only 10,000 images for five different languages. This is far from addressing the multilingual visual text rendering task, considering the number of characters for these non-English languages is much larger than the number of characters in English. In addition, we also empirically find that all the latest commercial image generation models like \dalle, Imagen3, Stable Diffusion 3~\cite{esser2024scaling}, and \ideogram\footnote{\url{https://about.ideogram.ai/1.0}} perform relatively poorly on multilingual visual text rendering tasks.

\vspace{2mm}
\noindent\textbf{Our Contribution} Our work mainly extends the previous Glyph-ByT5~\cite{liu2024glyph} for accurate multilingual visual text rendering tasks. The primary contribution of our work lies in presenting a scalable high-quality dataset of $1$ million multilingual glyph-text pairs and $10$ million multilingual graphic design images, training a powerful multilingual Glyph-ByT5-v2 that aligns multilingual text into the glyph image space, and training a multilingual Glyph-SDXL-v2 that supports accurate visual text rendering in ten different languages.

\section{Our Approach}
\label{sec:approach}

We build the multilingual visual text rendering approach based on the previous Glyph-ByT5 and Glyph-SDXL~\cite{liu2024glyph} by making the following improvements. We keep the design of the glyph text encoder, glyph vision encoder, and box-level contrastive loss unchanged by default. We illustrate more details of the improvements as follows.

\subsection{Multilingual Glyph-ByT5}

\vspace{1mm}
\noindent\textbf{Multilingual Glyph-Text Dataset}
The first key contribution of this work is to build a scalable and accurate multilingual (paragraph-)glyph-text dataset $\mathcal{D}^\mathrm{glyph}_\mathrm{multilingual}$ and the multilingual graphic design dataset$\mathcal{D}_\mathrm{multilingual}^\mathrm{design}$.
We illustrate the statistics of the constructed glyph-text dataset in the first column Table~\ref{table:multilingual_data_statistics}.

\begin{figure*}[t]
\centering
\begin{subfigure}[b]{0.16\textwidth}
{\includegraphics[width=\textwidth]{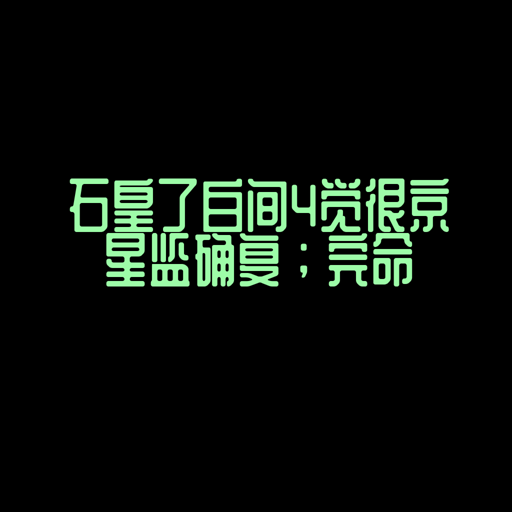}}
\caption*{Original}
\end{subfigure}
\begin{subfigure}[b]{0.16\textwidth}
{\includegraphics[width=\textwidth]{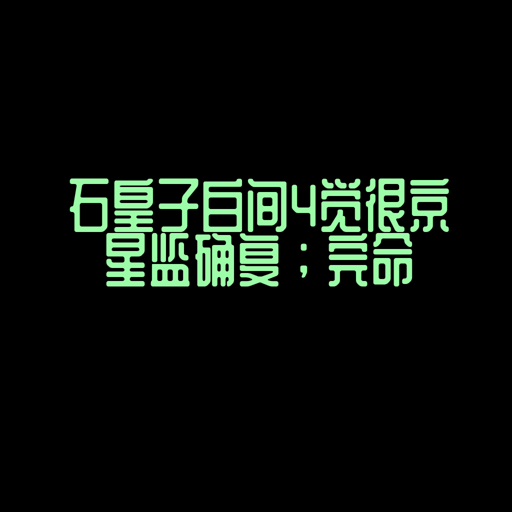}}
\caption*{\begin{CJK}{UTF8}{gbsn}了\end{CJK} $\rightarrow$ \begin{CJK}{UTF8}{gbsn}子\end{CJK}}
\end{subfigure}
\begin{subfigure}[b]{0.16\textwidth}
{\includegraphics[width=\textwidth]{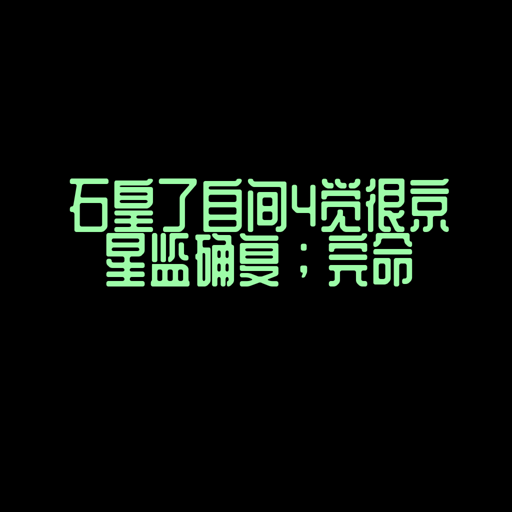}}
\caption*{\begin{CJK}{UTF8}{gbsn}白\end{CJK} $\rightarrow$ \begin{CJK}{UTF8}{gbsn}自\end{CJK}}
\end{subfigure}
\begin{subfigure}[b]{0.16\textwidth}
{\includegraphics[width=\textwidth]{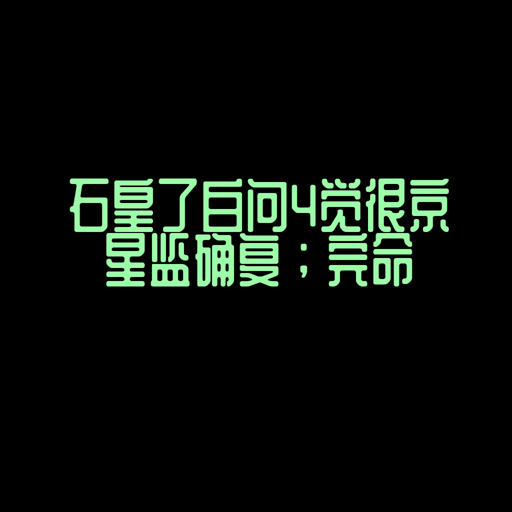}}
\caption*{\begin{CJK}{UTF8}{gbsn}间\end{CJK} $\rightarrow$ \begin{CJK}{UTF8}{gbsn}问\end{CJK}}
\end{subfigure}
\begin{subfigure}[b]{0.16\textwidth}
{\includegraphics[width=\textwidth]{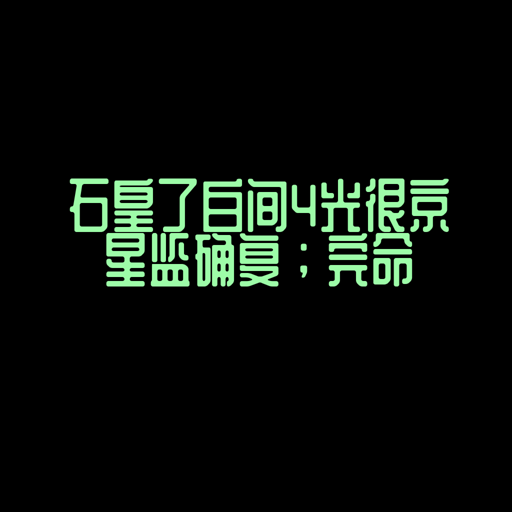}}
\caption*{\begin{CJK}{UTF8}{gbsn}觉\end{CJK} $\rightarrow$ \begin{CJK}{UTF8}{gbsn}光\end{CJK}}
\end{subfigure}
\begin{subfigure}[b]{0.16\textwidth}
{\includegraphics[width=\textwidth]{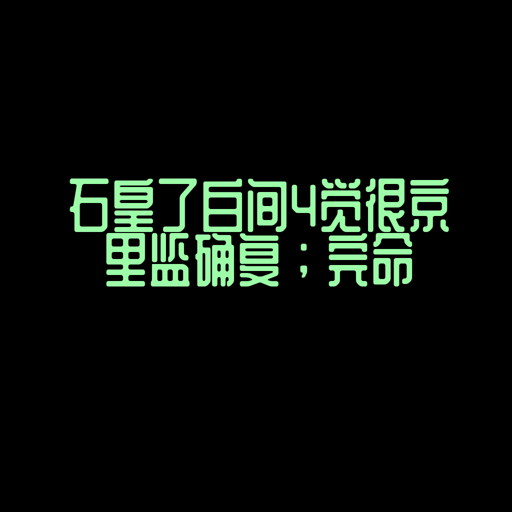}}
\caption*{\begin{CJK}{UTF8}{gbsn}星\end{CJK} $\rightarrow$ \begin{CJK}{UTF8}{gbsn}里\end{CJK}}
\end{subfigure}
\captionof{figure}{\small{Illustrating the similarly-shaped character replacement strategy for Chinese glyph augmentation.}}
\label{fig:glyph_augmentation}
\vspace{2mm}
\centering
\begin{subfigure}[b]{0.195\textwidth}
{\includegraphics[width=\textwidth]{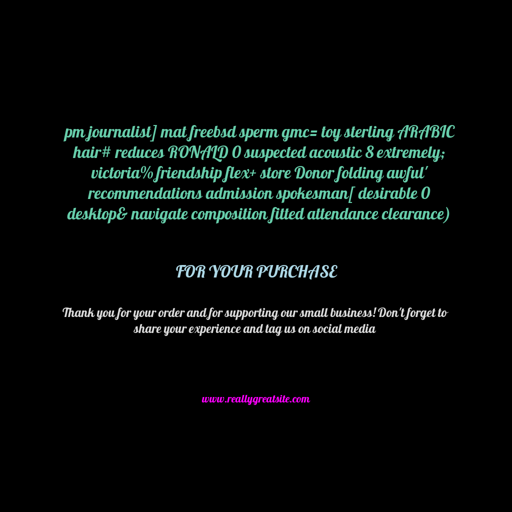}}
\caption*{\scriptsize{English}}
\end{subfigure}
\begin{subfigure}[b]{0.195\textwidth}
{\includegraphics[width=\textwidth]{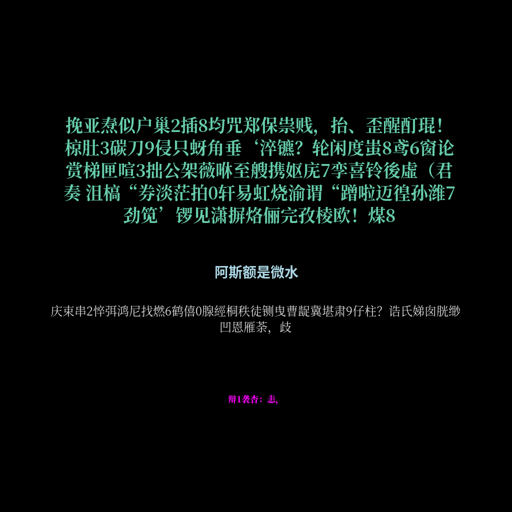}}
\caption*{\scriptsize{Chinese}}
\end{subfigure}
\begin{subfigure}[b]{0.195\textwidth}
{\includegraphics[width=\textwidth]{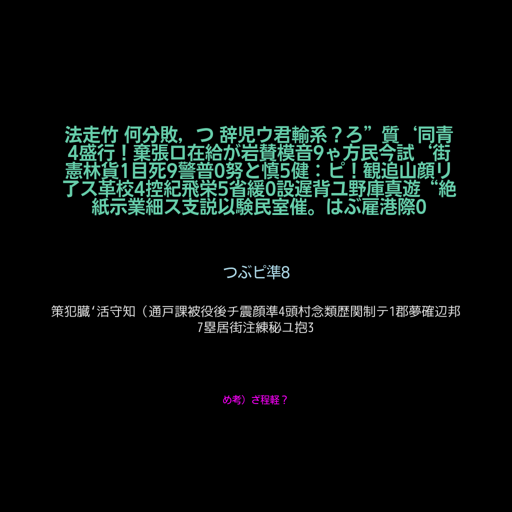}}
\caption*{\scriptsize{Japanese}}
\end{subfigure}
\begin{subfigure}[b]{0.195\textwidth}
{\includegraphics[width=\textwidth]{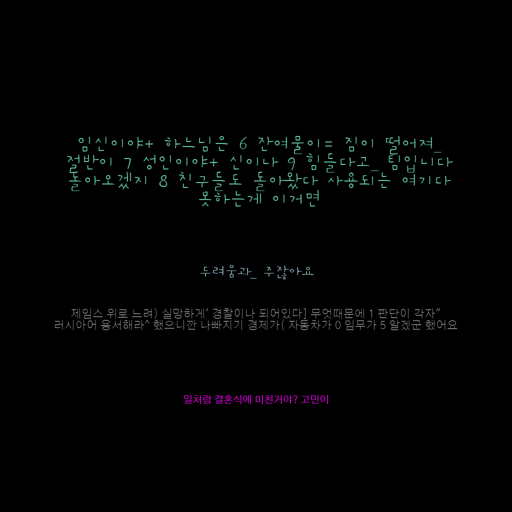}}
\caption*{\scriptsize{Korean}}
\end{subfigure}
\begin{subfigure}[b]{0.195\textwidth}
{\includegraphics[width=\textwidth]{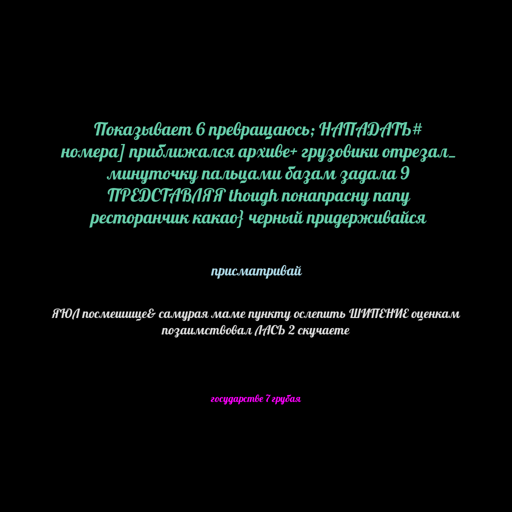}}
\caption*{\scriptsize{Russian}}
\end{subfigure} \\
\begin{subfigure}[b]{0.195\textwidth}
{\includegraphics[width=\textwidth]{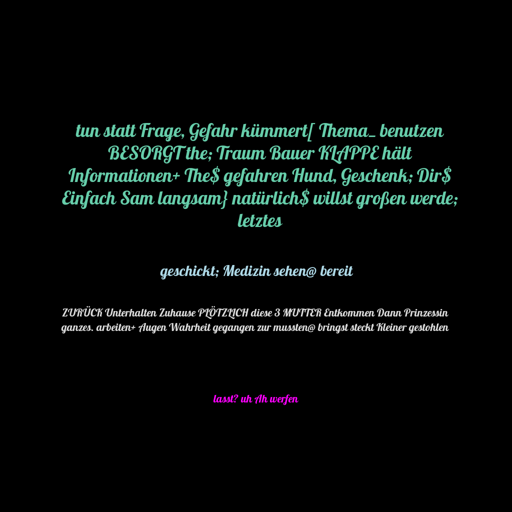}}
\caption*{\scriptsize{German}}
\end{subfigure}
\begin{subfigure}[b]{0.195\textwidth}
{\includegraphics[width=\textwidth]{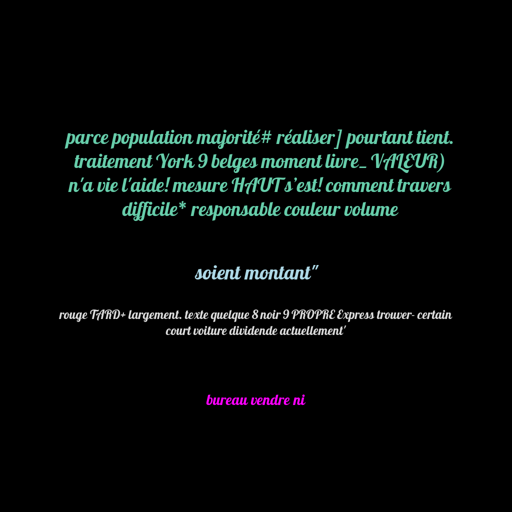}}
\caption*{\scriptsize{French}}
\end{subfigure}
\begin{subfigure}[b]{0.195\textwidth}
{\includegraphics[width=\textwidth]{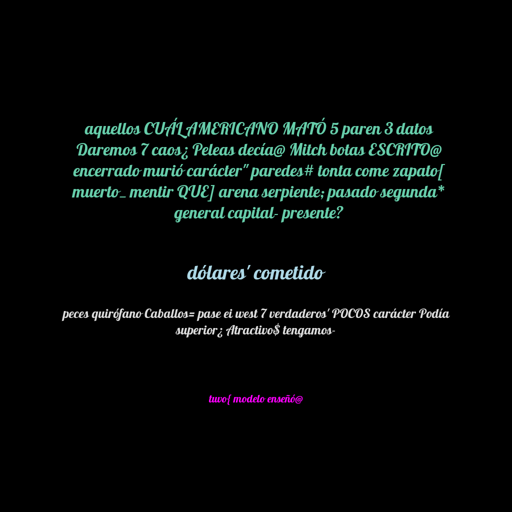}}
\caption*{\scriptsize{Spanish}}
\end{subfigure}
\begin{subfigure}[b]{0.195\textwidth}
{\includegraphics[width=\textwidth]{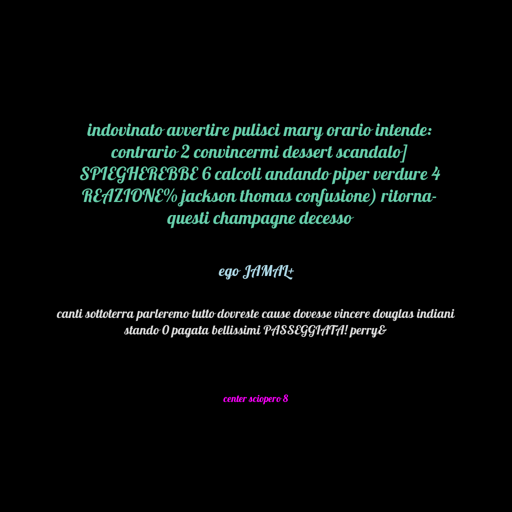}}
\caption*{\scriptsize{Italian}}
\end{subfigure}
\begin{subfigure}[b]{0.195\textwidth}
{\includegraphics[width=\textwidth]{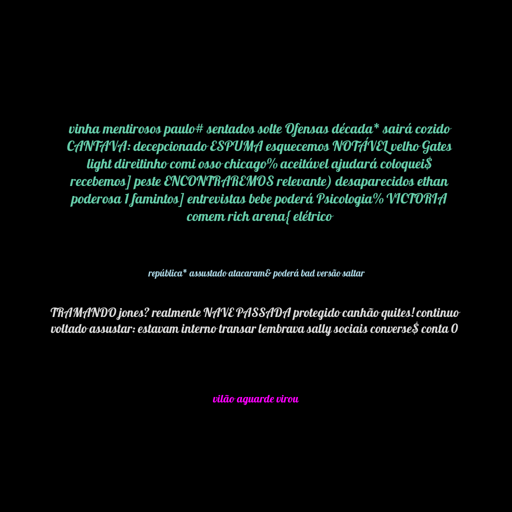}}
\caption*{\scriptsize{Portuguese}}
\end{subfigure} \\
\captionof{figure}{\small{Illustration of using the translation-based approach to transform the English glyph image into glyph images of other languages. }}
\label{fig:multilingual_glyph_text}
\vspace{7mm}
\centering
\begin{subfigure}[b]{0.195\textwidth}
{\includegraphics[width=\textwidth]{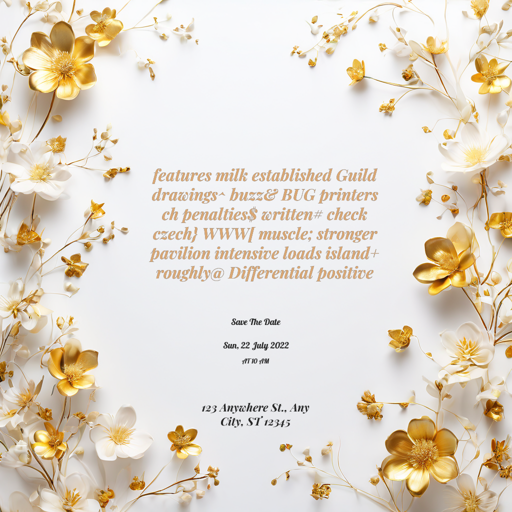}}
\caption*{\scriptsize{English}}
\end{subfigure}
\begin{subfigure}[b]{0.195\textwidth}
{\includegraphics[width=\textwidth]{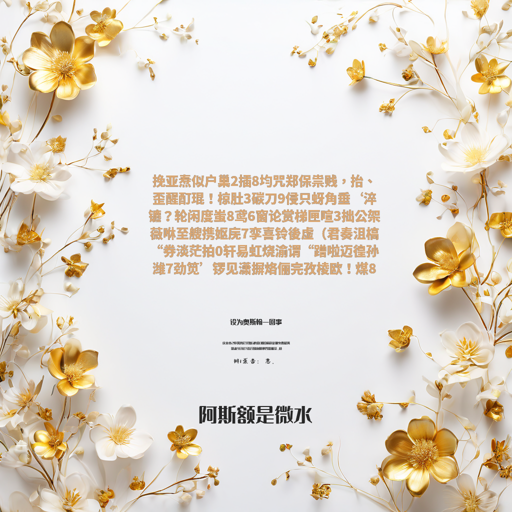}}
\caption*{\scriptsize{Chinese}}
\end{subfigure}
\begin{subfigure}[b]{0.195\textwidth}
{\includegraphics[width=\textwidth]{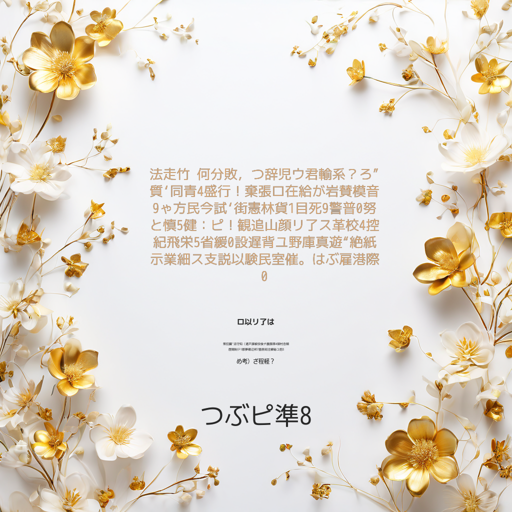}}
\caption*{\scriptsize{Japanese}}
\end{subfigure}
\begin{subfigure}[b]{0.195\textwidth}
{\includegraphics[width=\textwidth]{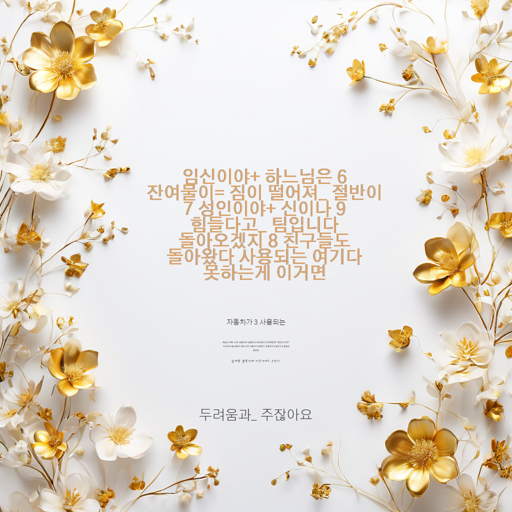}}
\caption*{\scriptsize{Korean}}
\end{subfigure}
\begin{subfigure}[b]{0.195\textwidth}
{\includegraphics[width=\textwidth]{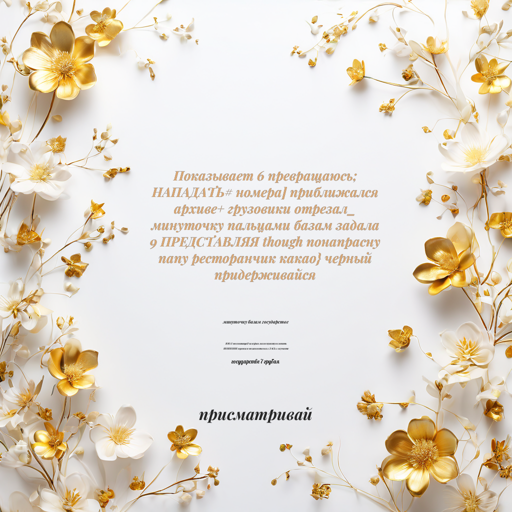}}
\caption*{\scriptsize{Russian}}
\end{subfigure} \\
\begin{subfigure}[b]{0.195\textwidth}
{\includegraphics[width=\textwidth]{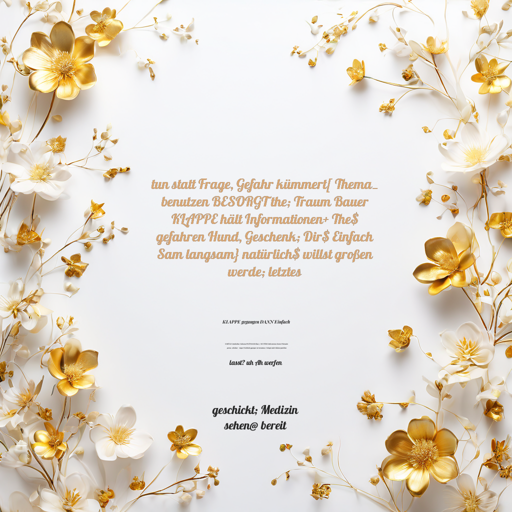}}
\caption*{\scriptsize{German}}
\end{subfigure}
\begin{subfigure}[b]{0.195\textwidth}
{\includegraphics[width=\textwidth]{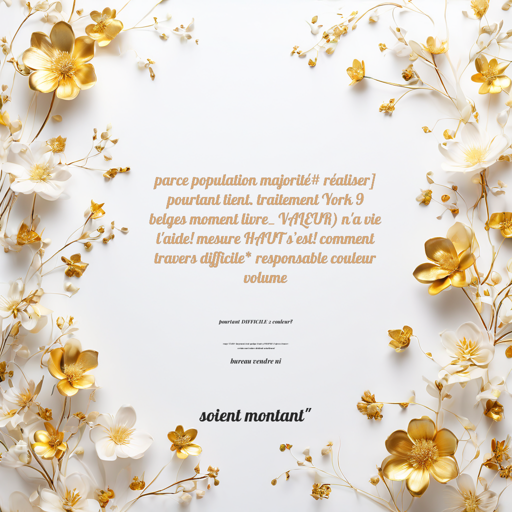}}
\caption*{\scriptsize{French}}
\end{subfigure}
\begin{subfigure}[b]{0.195\textwidth}
{\includegraphics[width=\textwidth]{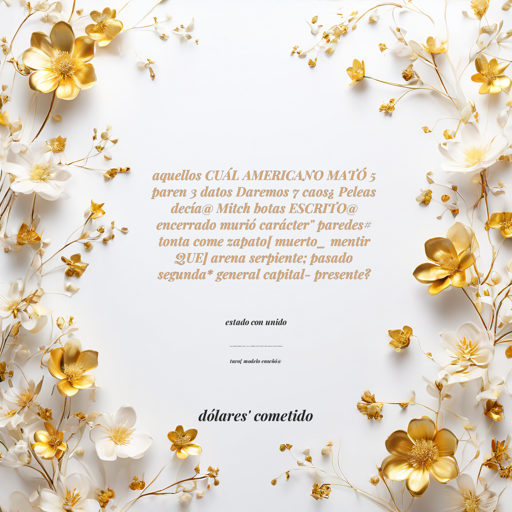}}
\caption*{\scriptsize{Spanish}}
\end{subfigure}
\begin{subfigure}[b]{0.195\textwidth}
{\includegraphics[width=\textwidth]{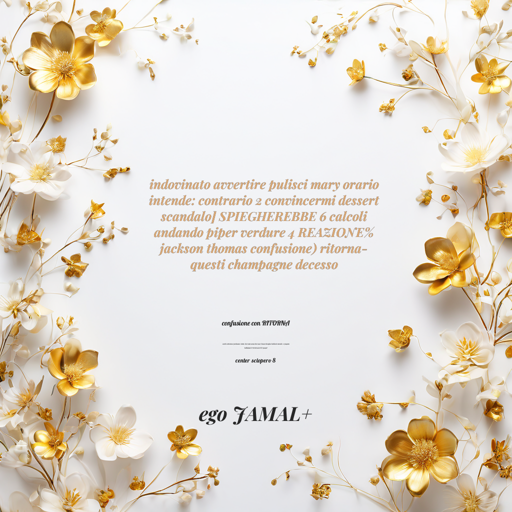}}
\caption*{\scriptsize{Italian}}
\end{subfigure}
\begin{subfigure}[b]{0.195\textwidth}
{\includegraphics[width=\textwidth]{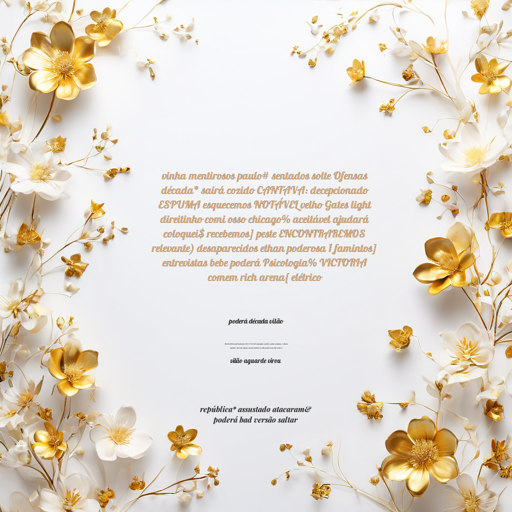}}
\caption*{\scriptsize{Portuguese}}
\end{subfigure} \\
\captionof{figure}{\small{Illustration of using the translation-based approach to transform the English design image into design images of other languages.}}
\label{fig:multilingual_design}
\end{figure*}

Given that the number of glyph images and graphic designs corresponding to the other nine languages is much smaller compared to English, it is impractical to collect a large number of high-quality graphic design images for these languages. To address this challenging issue, we propose a simple yet effective translation-based approach to generate a large number of multilingual glyph-text and paragraph-glyph-text pairs. This is achieved by transforming the high-quality English glyph-text dataset, created using the graphic renderer based on \cite{jia2023cole}, into datasets for other languages. We explain how we construct the multilingual dataset as follows.
We construct a multilingual vocabulary set consisting of the $5,000$ most frequently used words in each language. We also generate a large number of multilingual glyph-text pairs using a random strategy that transforms the original English characters into characters of other languages. Additionally, we need to consider the differences in font types across different languages. Figure~\ref{fig:multilingual_glyph_text} illustrates representative cases of transforming an English glyph image into glyph images of nine other languages.

\begin{table}[!t]
\begin{minipage}[t]{1\linewidth}
\centering
\tablestyle{15pt}{1.35}
\resizebox{1\linewidth}{!}
{
\begin{tabular}{l|c|c}
\multirow{2}{*}{Language} & Glyph-ByT5-v2 & Glyph-SDXL-v2 \\\cline{2-3}
& \# of glyph-text pairs & \# of design-image-text pairs \\
\shline
English & $100$K & $1$M \\
Chinese & $300$K & $3$M \\
French & $20$K & $200$K \\
German & $20$K & $200$K \\
Spanish & $20$K & $200$K \\
Portuguese & $20$K & $200$K \\
Italian & $20$K & $200$K \\
Russian & $100$K & $1$M \\
Japanese & $100$K & $1$M \\
Korean & $100$K & $1$M \\
Mixed & $200$K & $2$M \\
\end{tabular}
}
\vspace{-2mm}
\caption{\footnotesize Illustrating the details of multilingual training dataset of Glyph-ByT5-v2 and Glyph-SDXL-v2.}
\label{table:multilingual_data_statistics}
\end{minipage}
\end{table}

\vspace{1mm}
\noindent\textbf{Multilingual Glyph Augmentation}
Similar to Glyph-ByT5, we apply glyph augmentation during the glyph-alignment pre-training stage. For alphabetic languages, we use the same augmentation strategies as for English, namely \emph{glyph replacement}, \emph{glyph repeat}, \emph{glyph drop}, and \emph{glyph add} at both the character level and word level.
For character-based languages like Chinese, Japanese, and Korean, we apply \emph{glyph repeat} and \emph{glyph drop} only at the character level, as it is difficult to define word-level combinations for these languages.
In addition, to overcome the challenge of modeling the complex structures of various Chinese characters, we design a similarly-shaped character replacement strategy that replaces selected characters with the most similar ones based on their shape.
To help users understand these designs, we provide a typical example of the similarly-shaped character replacement scheme in Figure~\ref{fig:glyph_augmentation}, where we randomly select a character and replace it with a structurally similar character for each augmented case.

\vspace{1mm}
\noindent\textbf{Other Settings}
We chose to build the multilingual Glyph-ByT5-v2 text encoder from the ByT5-Small text encoder (217M parameters), implemented the visual encoder based on DINOv2 with ViT-B/14 (86M parameters), and applied the original box-level contrastive loss directly. We treated all images of different languages equally without applying separate box-level contrastive loss for each language independently.
In addition, we also apply the hard-negative contrastive loss based on glyph augmentation to further improve the visual spelling accuracy.

\subsection{Multilingual Glyph-SDXL}

We build the multilingual Glyph-SDXL-v2 by integrating the multilingual Glyph-ByT5-v2 with a improved SDXL~\cite{liang2024step} by following the region-wise multi-head cross-attention scheme proposed in~\cite{liu2024glyph}.

\vspace{1mm}
\noindent\textbf{Multilingual Design Dataset}
The second key challenge of this work is also at the lack of high-quality multilingual graphic design datasets for training a multilingual design-text generation model.
We follow the similar translation based manner to transform the previous English graphic design images into the ones of other language and illustrate a representative example in Figure~\ref{fig:multilingual_design}.

We use the same English background captions, generated with LLaVA~\cite{liu2024visual} based on Llama 2-13B~\cite{touvron2023llama}, and apply different languages only for the prompts sent to the multilingual Glyph-ByT5-v2 text encoder. We also follow the region-wise multi-text-encoder fusion scheme to integrate the glyph-aware capabilities of our multilingual Glyph-ByT5-v2 text encoder with the strengths of the two original CLIP text encoders as described in~\cite{liu2024glyph}.
Table~\ref{table:multilingual_data_statistics} illustrates the statistics of the constructed multilingual design dataset on the second column.
\section{Experiment}
\label{sec:experiment}

\subsection{Training Settings}
We illustrate the hyperparameters for training Glyph-ByT5-v2 and Glyph-SDXL-v2 in Table \ref{table:glyph_hparam} and Table \ref{table:sdxl_hparam}, respectively. We train Glyph-ByT5-v2 with $4\times$ A100 GPUs and Glyph-SDXL-v2 with $32 \times$ A100 GPUs.
To evaluate our method's ability to generate accurate multilingual design text in graphic design images, we developed the multilingual \textsc{VisualParagraphy} benchmark. We compared our method against the previous Glyph-SDXL and commercial products such as \dalle in the design-text generation task. Our evaluation includes objective OCR metrics and a subjective user study to assess visual quality from multiple perspectives.

\begin{table}[!t]
\begin{minipage}[t]{0.45\linewidth}
\centering
\tablestyle{1pt}{1.5}
\resizebox{0.99\linewidth}{!}
{
\begin{tabular}{l|c}
Hyperparameter &            Glyph-ByT5-Small  \\
\shline
Text Encoder &    ByT5-Small \\
Vision Encoder &     DINOv2 ViT-B/14 \\
Peak Learning-rate &    5.00E-04  \\
Batch Size & 1536 \\
Epochs &  5 \\
Warmup Iterations &  100  \\
Weight Decay &     0.2  \\
Text-Encoder Dropout &      0.1  \\
\end{tabular}
}
\vspace{-2mm}
\caption{\footnotesize Glyph-ByT5-v2 pre-training hyper-parameters.}
\label{table:glyph_hparam}
\end{minipage}
\begin{minipage}[t]{0.5\linewidth}
\centering
\tablestyle{1pt}{1.5}
\resizebox{0.99\linewidth}{!}
{
\begin{tabular}{l|c}
Hyperparameter     &        Glyph-SDXL-Small  \\
\shline
Text Encoder  &    Glyph-ByT5-Small \\
UNet Learning-rate &            5.00E-05\\
Text Enoder Learning-rate &            1.00E-04  \\
Batch Size &                256 \\
Iterations &                  10000   \\
Weight Decay &                0.01 \\
Text-Encoder Weight Decay &                0.2 \\
Text-Encoder Dropout &                   0.1  \\
Gradient Clipping &                   1.0 \\
\end{tabular}
}
\vspace{-2mm}
\caption{\footnotesize Glyph-SDXL-v2 model training hyper-parameters.}
\label{table:sdxl_hparam}
\end{minipage}
\vspace{-3mm}
\end{table}

\subsection{Evaluation Metrics}

We categorize the ten different languages into two groups: (i) Alphabetic languages, consisting of English, French, German, Spanish, Italian, Portuguese, and Russian, use words as the basic unit of each sentence. These languages employ alphabets to form words, which can be easily identified using spaces. (ii) Character-based languages, consisting of Chinese, Japanese, and Korean, use characters as the basic unit of each sentence.
Considering the differences between these languages, we adopt case-sensitive word-level precision for the alphabetic languages and character-level precision for the character-based languages. Additionally, we recruited 10 users with design backgrounds to evaluate and compare the quality of images generated by Glyph-SDXL-v2 vs. Glyph-SDXL, and Glyph-SDXL-v2 vs. \dalle.

\begin{figure}[htbp]
\centering
\includegraphics[width=\linewidth]{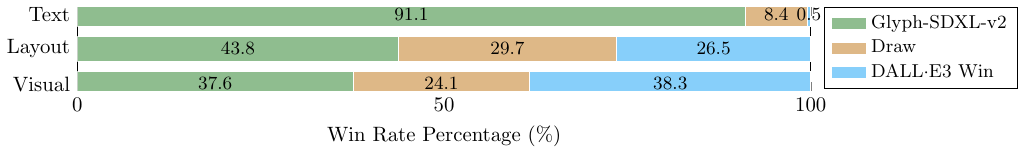} 
\caption{\footnotesize{Glyph-SDXL-v2 vs. \dalle in graphic design images in terms of multilingual visual text spelling accuracy, layout quality, and visual aesthetics win-rates evaluated by human evaluator preferences.}}
\label{fig:winrate_dalle}
\vspace{-3mm}
\end{figure}

\subsection{Multilingual \textsc{VisualParagraphy} Benchmark}
We constructed a multilingual benchmark for the design-text generation task, named Multilingual \textsc{VisualParagraphy}, amassing $100$ prompts for each language. For each language, we collected prompts with varying numbers of characters and difficulty levels: rendering fewer than $20$ characters, rendering $20$ to $50$ characters, rendering $50$ to $100$ characters, and rendering more than $100$ characters.

\begin{figure*}[t]
\begin{minipage}[t]{1\linewidth}
\centering
\begin{subfigure}[b]{0.16\textwidth}
\includegraphics[width=\textwidth]{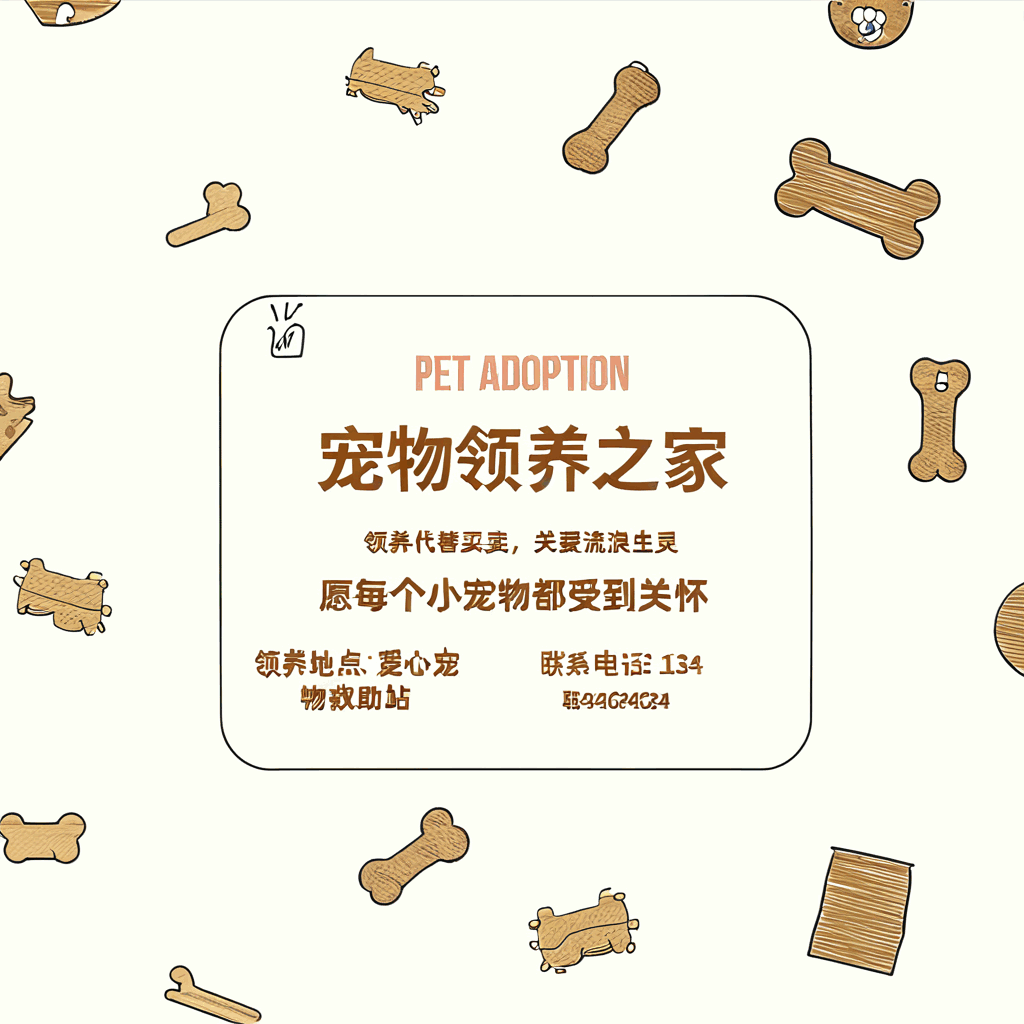}
\vspace{-3mm}
\end{subfigure}
\begin{subfigure}[b]{0.16\textwidth}
\includegraphics[width=\textwidth]{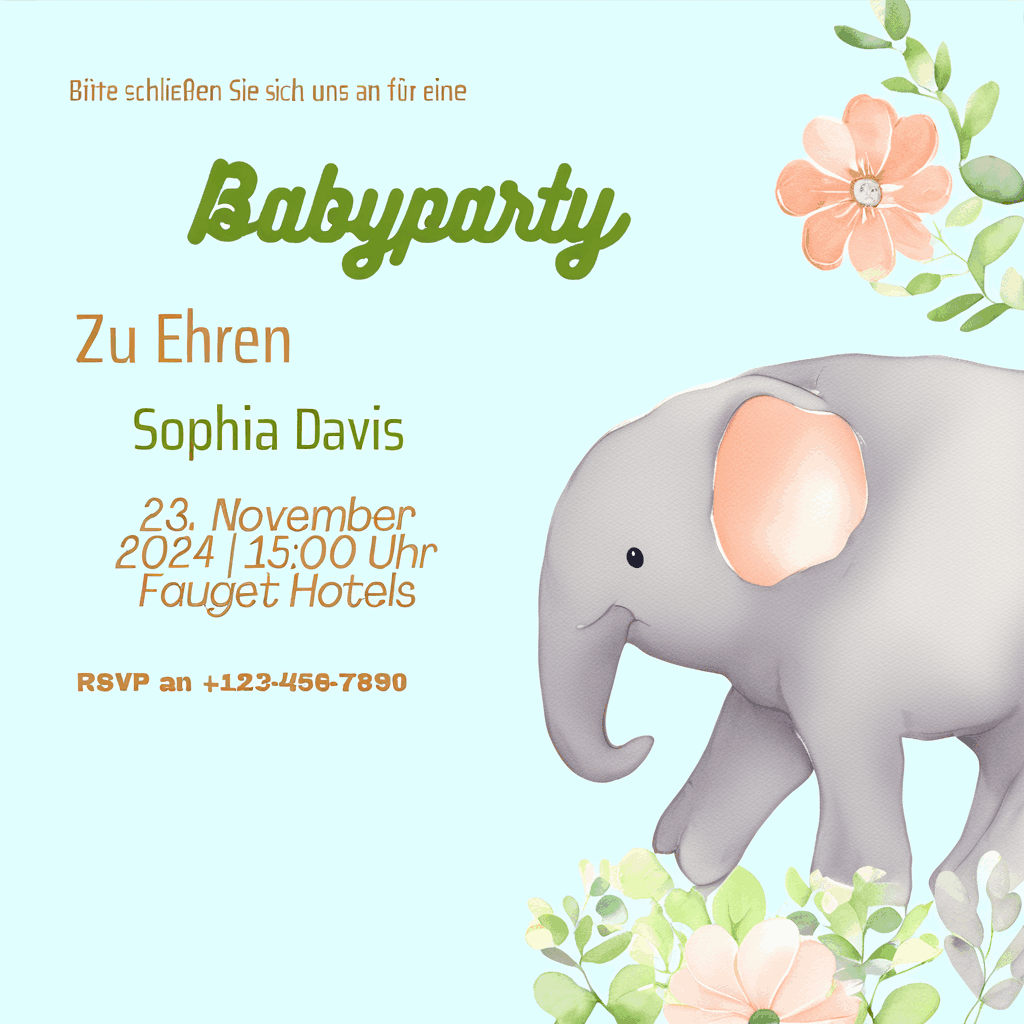}
\vspace{-3mm}
\end{subfigure}
\begin{subfigure}[b]{0.16\textwidth}
\includegraphics[width=\textwidth]{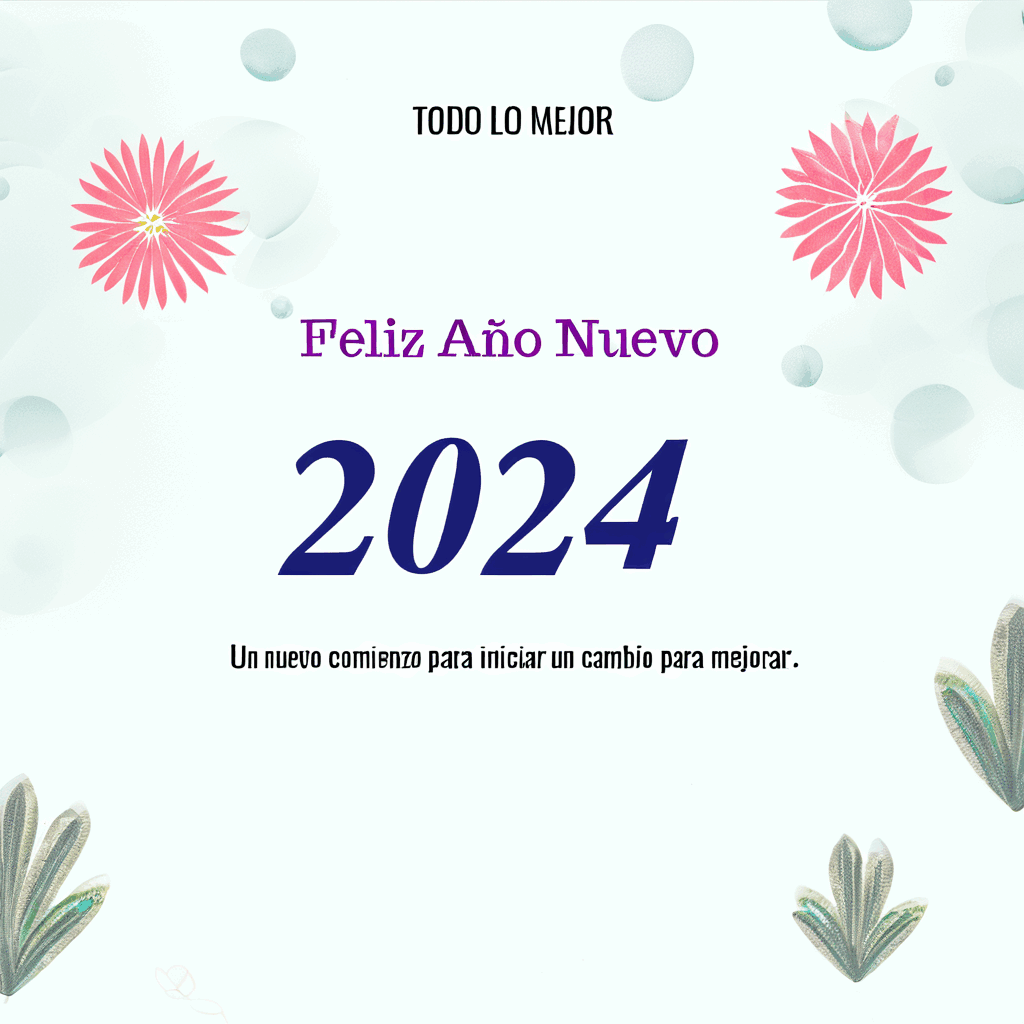}
\vspace{-3mm}
\end{subfigure}
\begin{subfigure}[b]{0.16\textwidth}
\includegraphics[width=\textwidth]{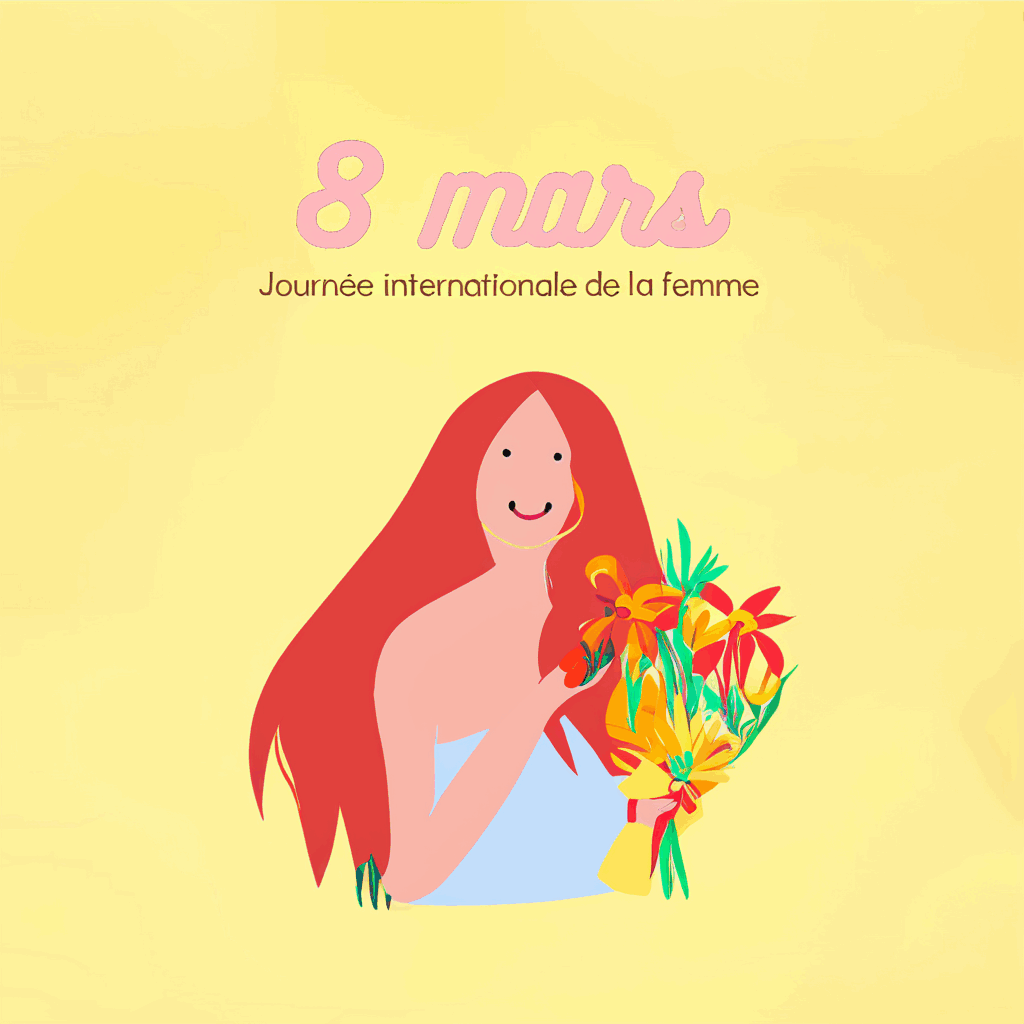}
\vspace{-3mm}
\end{subfigure}
\begin{subfigure}[b]{0.16\textwidth}
\includegraphics[width=\textwidth]{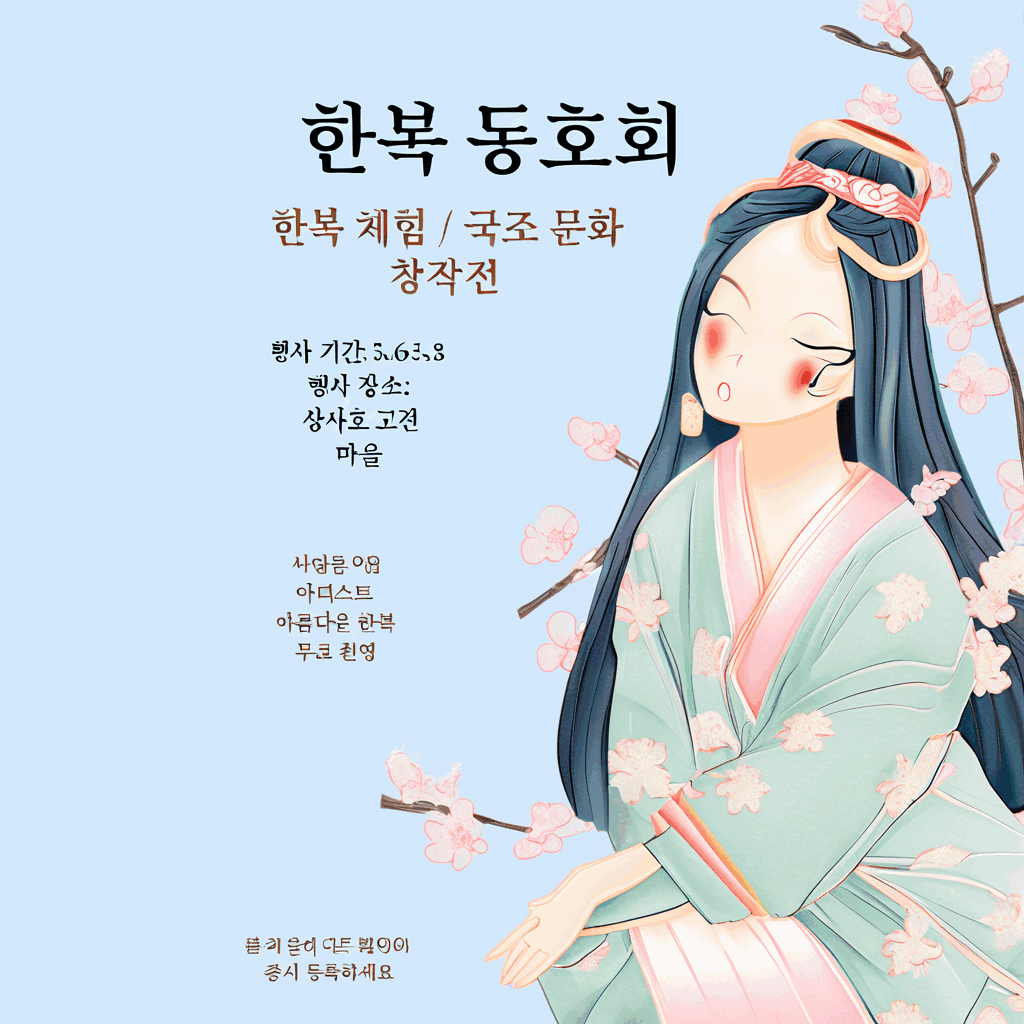}
\vspace{-3mm}
\end{subfigure}
\begin{subfigure}[b]{0.16\textwidth}
\includegraphics[width=\textwidth]{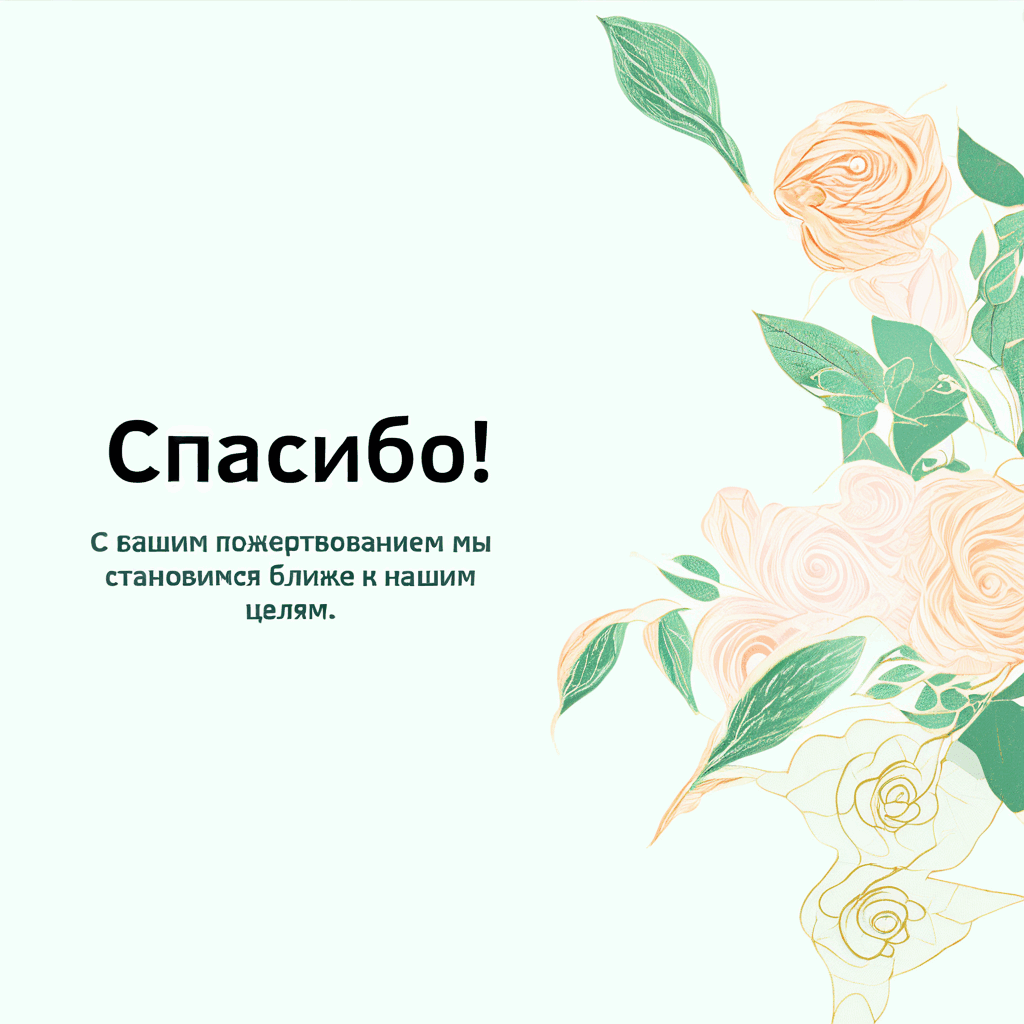}
\vspace{-3mm}
\end{subfigure}\\

\begin{subfigure}[b]{0.16\textwidth}
\includegraphics[width=\textwidth]{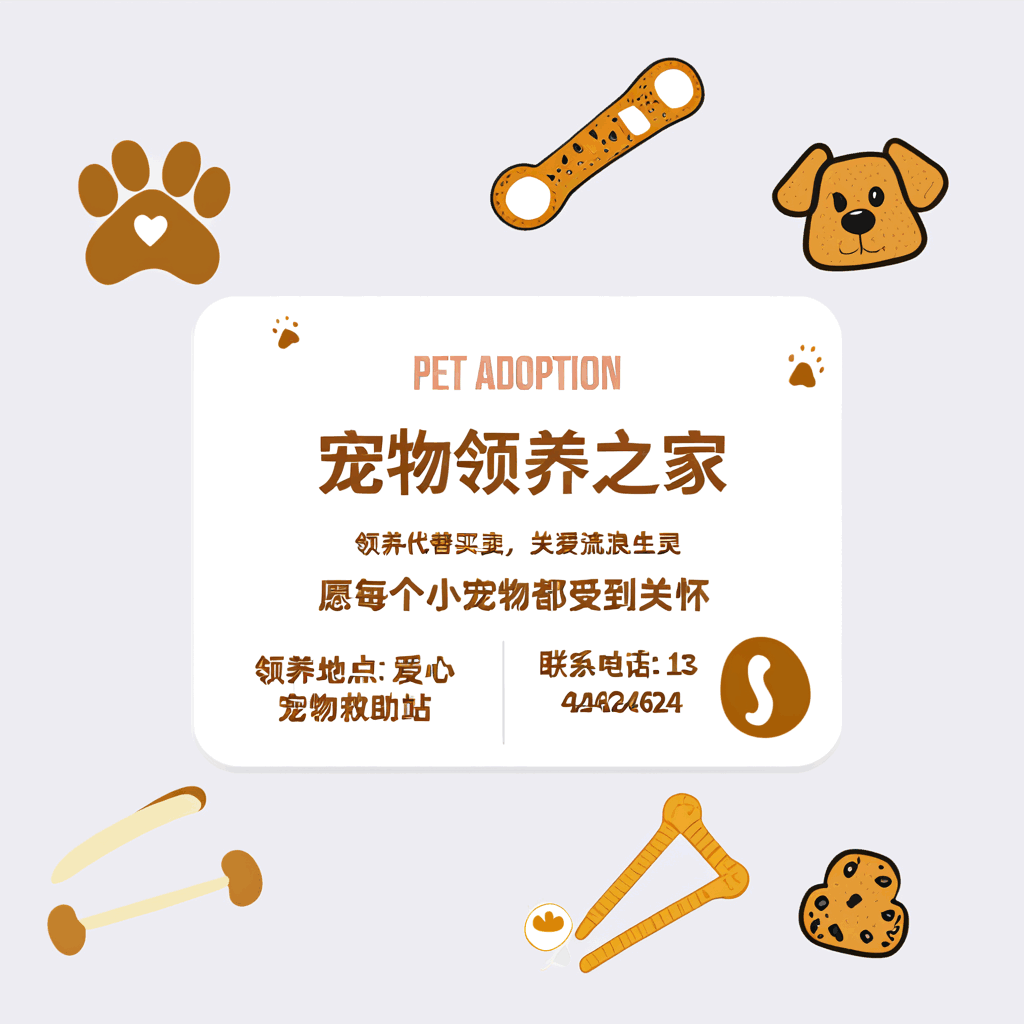}
\vspace{-3mm}
\end{subfigure}
\begin{subfigure}[b]{0.16\textwidth}
\includegraphics[width=\textwidth]{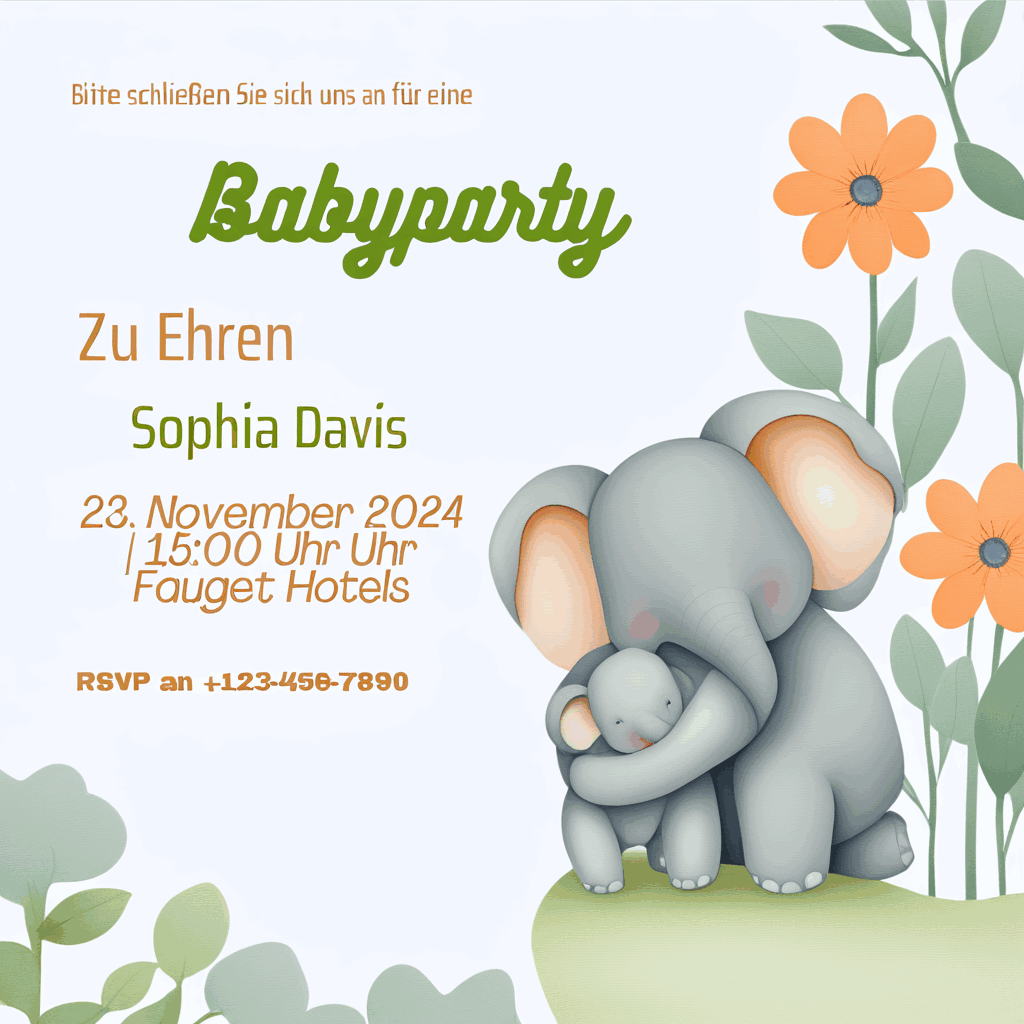}
\vspace{-3mm}
\end{subfigure}
\begin{subfigure}[b]{0.16\textwidth}
\includegraphics[width=\textwidth]{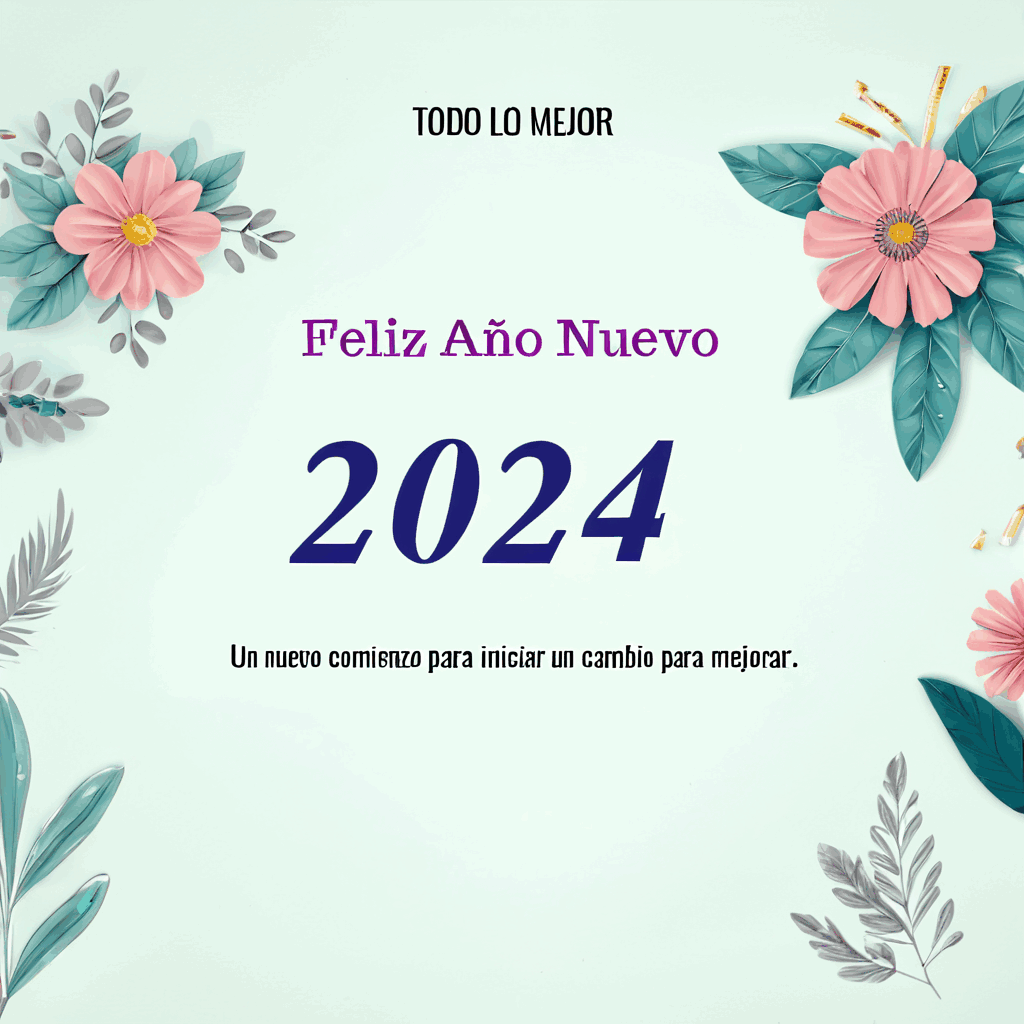}
\vspace{-3mm}
\end{subfigure}
\begin{subfigure}[b]{0.16\textwidth}
\includegraphics[width=\textwidth]{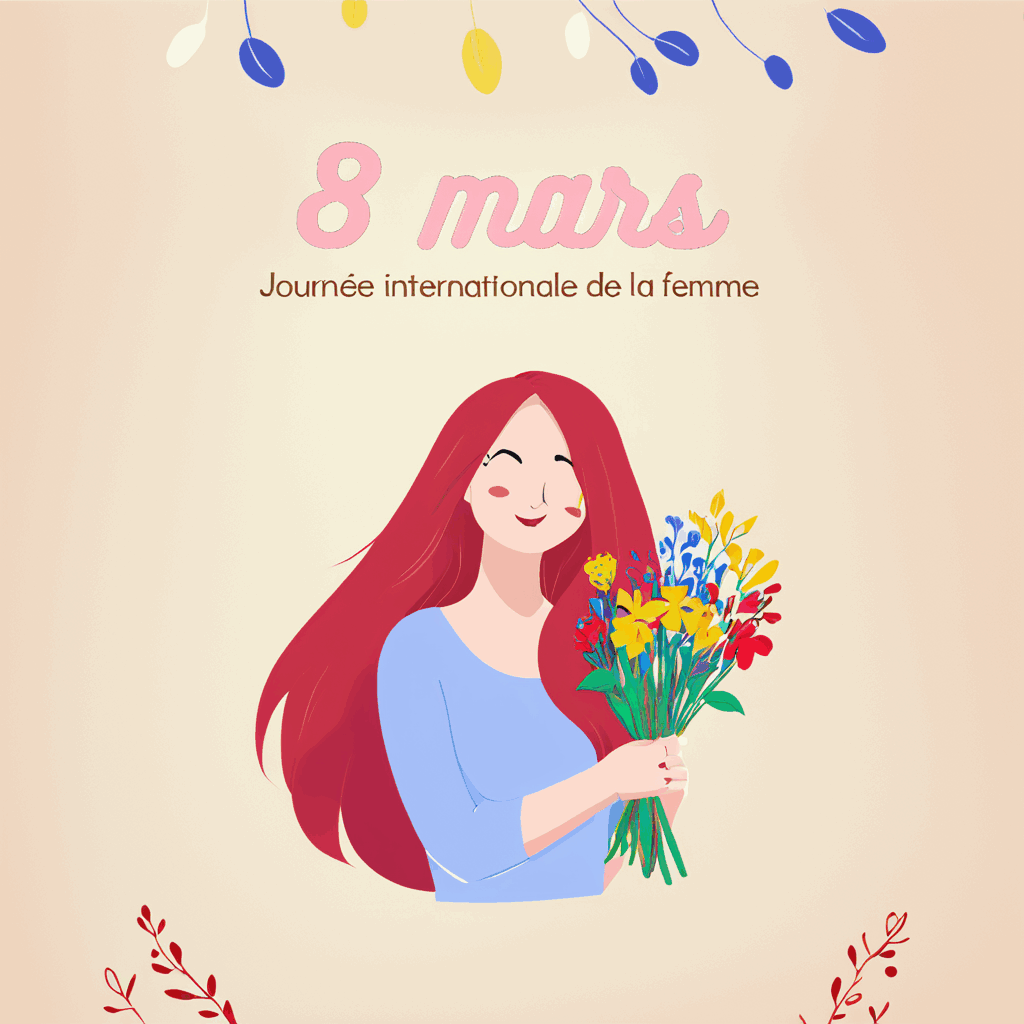}
\vspace{-3mm}
\end{subfigure}
\begin{subfigure}[b]{0.16\textwidth}
\includegraphics[width=\textwidth]{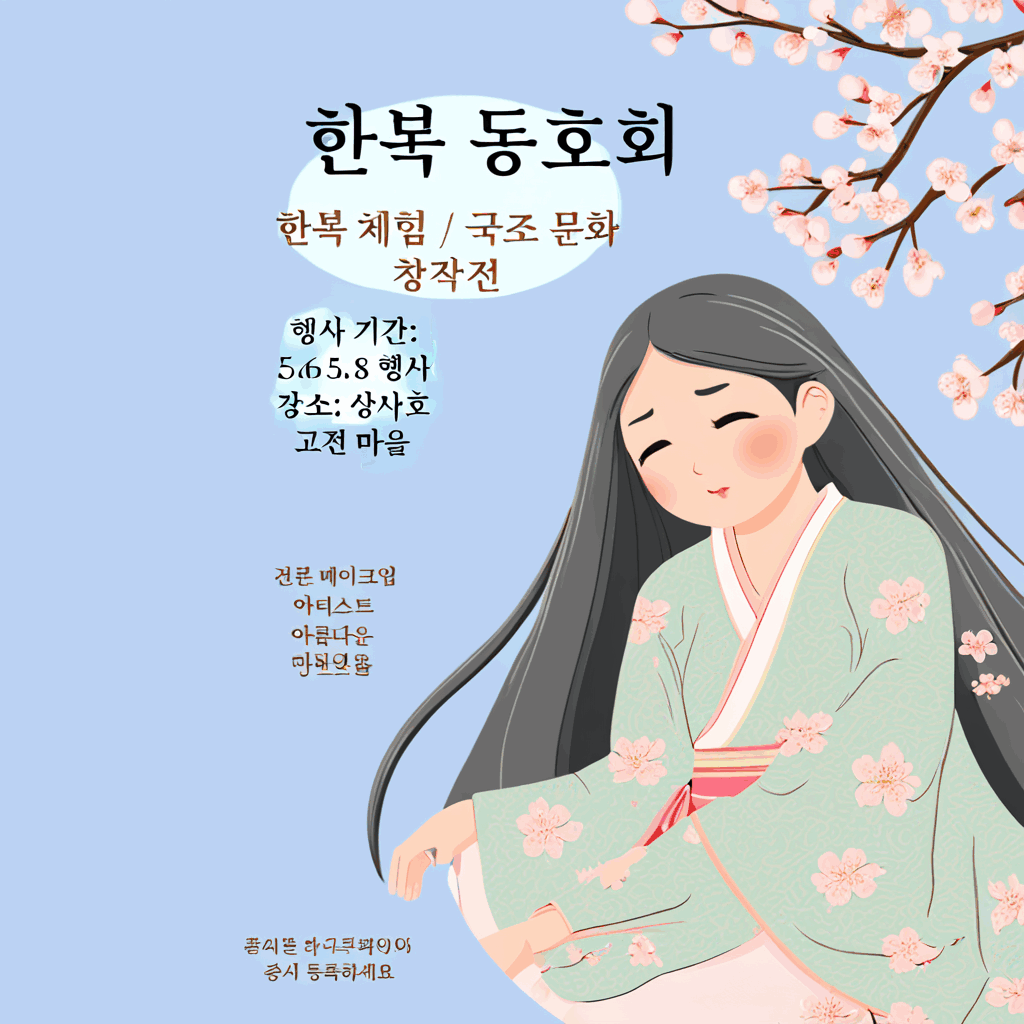}
\vspace{-3mm}
\end{subfigure}
\begin{subfigure}[b]{0.16\textwidth}
\includegraphics[width=\textwidth]{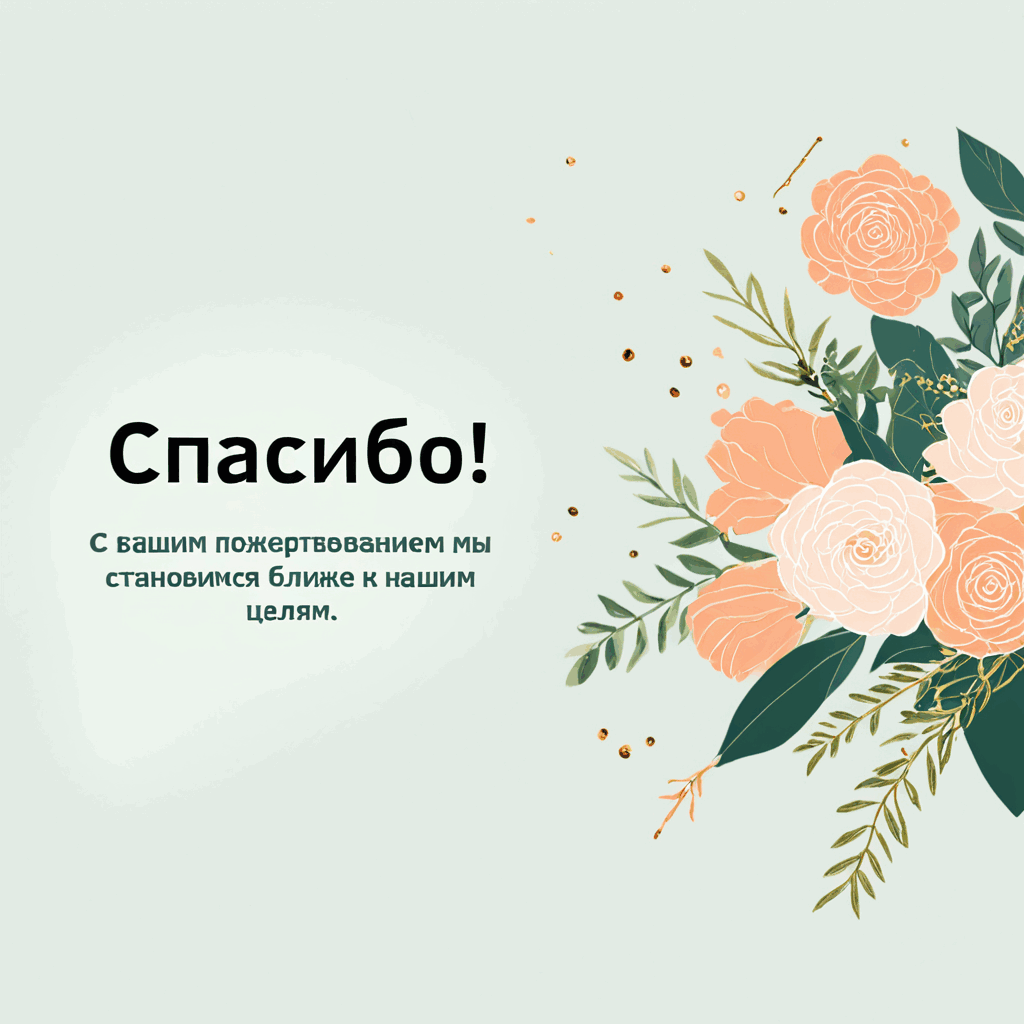}
\vspace{-3mm}
\end{subfigure}\\

\begin{subfigure}[b]{0.16\textwidth}
\includegraphics[width=\textwidth]{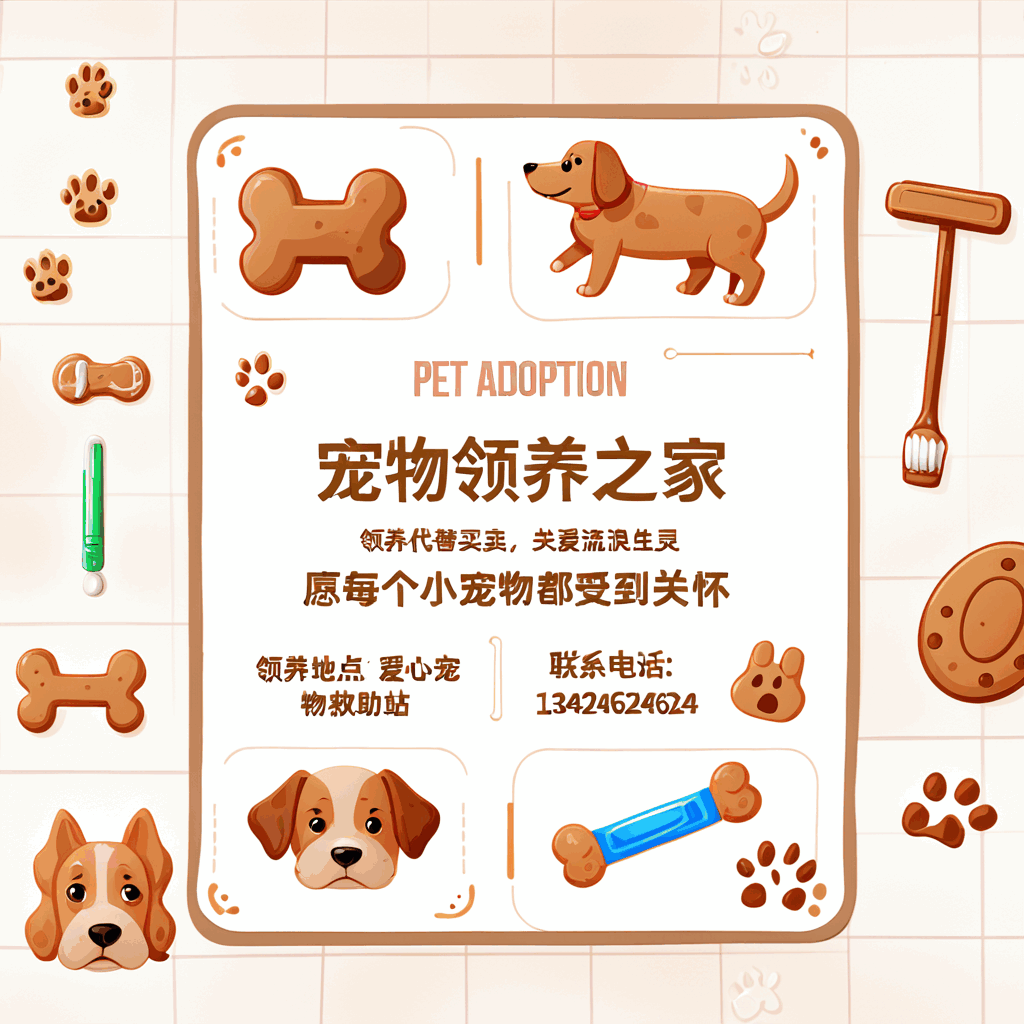}
\vspace{-3mm}
\end{subfigure}
\begin{subfigure}[b]{0.16\textwidth}
\includegraphics[width=\textwidth]{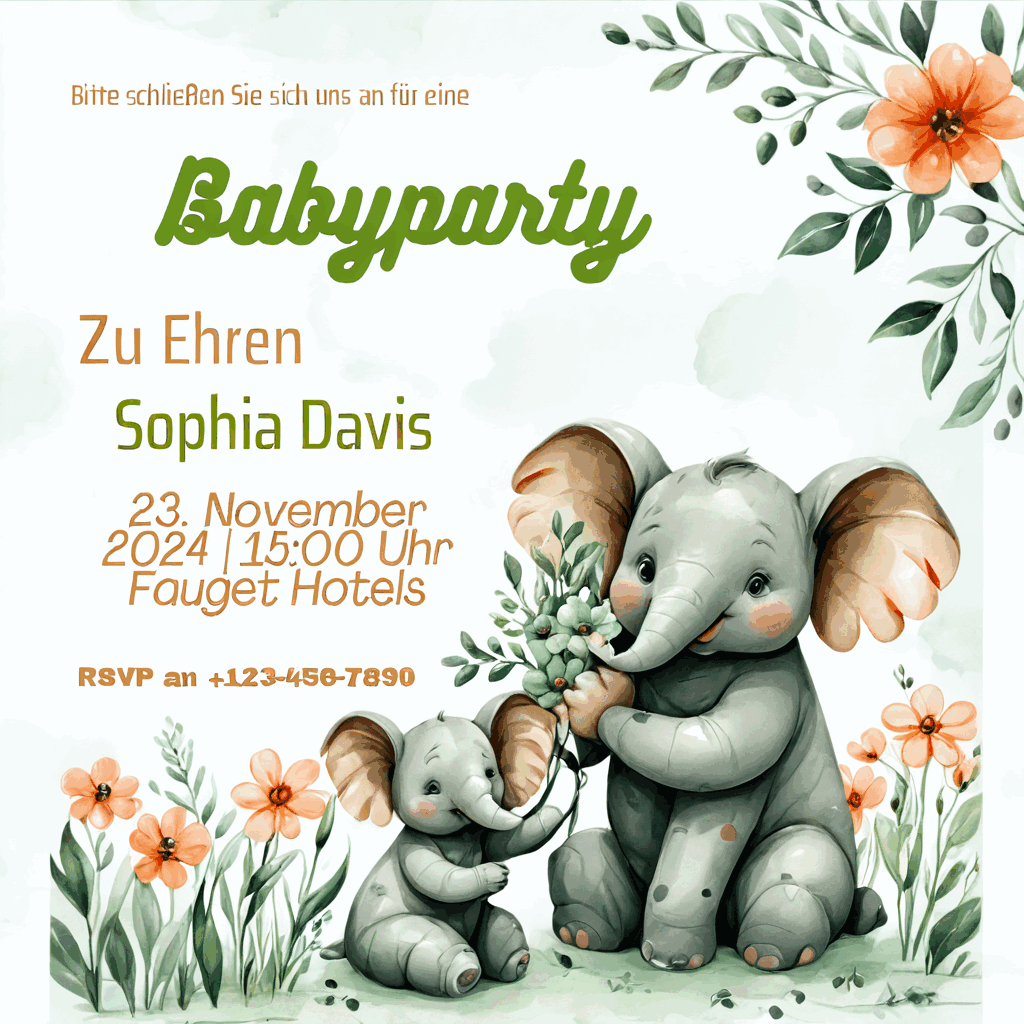}
\vspace{-3mm}
\end{subfigure}
\begin{subfigure}[b]{0.16\textwidth}
\includegraphics[width=\textwidth]{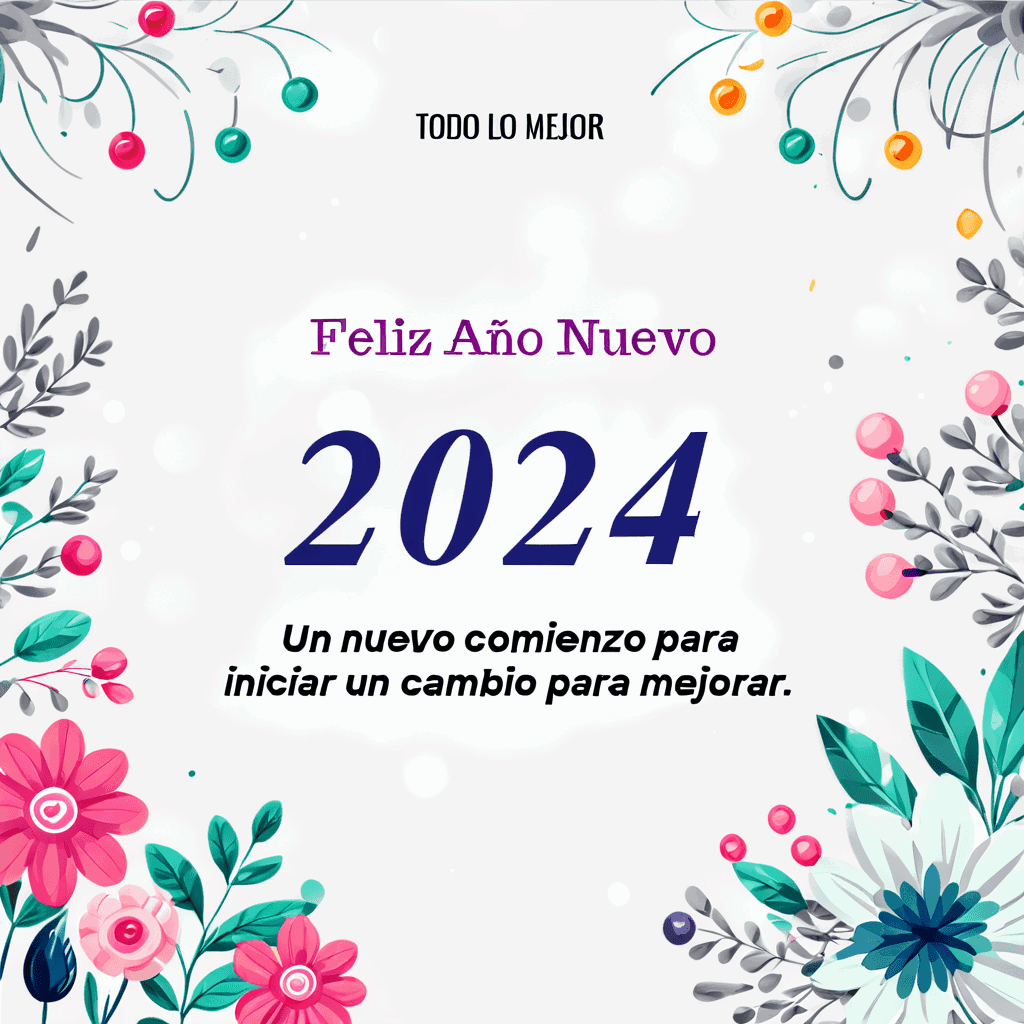}
\vspace{-3mm}
\end{subfigure}
\begin{subfigure}[b]{0.16\textwidth}
\includegraphics[width=\textwidth]{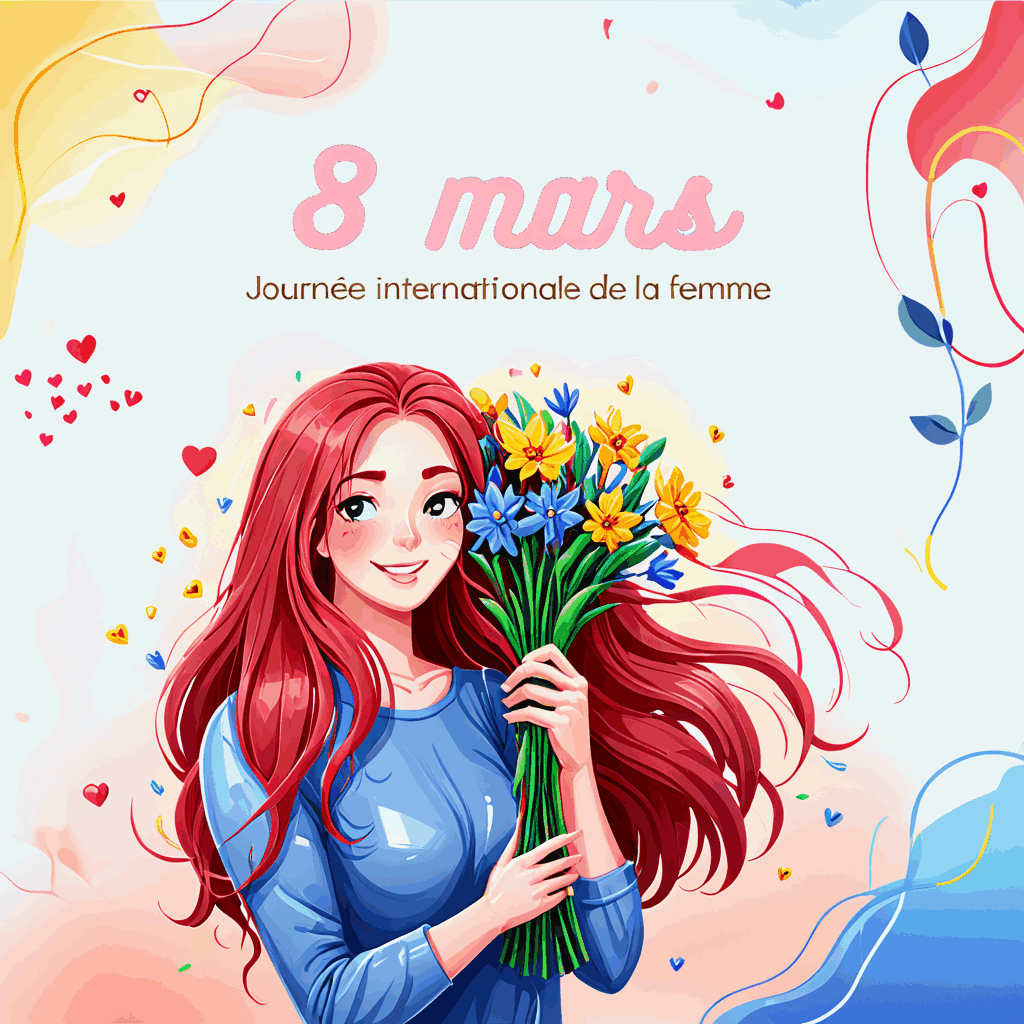}
\vspace{-3mm}
\end{subfigure}
\begin{subfigure}[b]{0.16\textwidth}
\includegraphics[width=\textwidth]{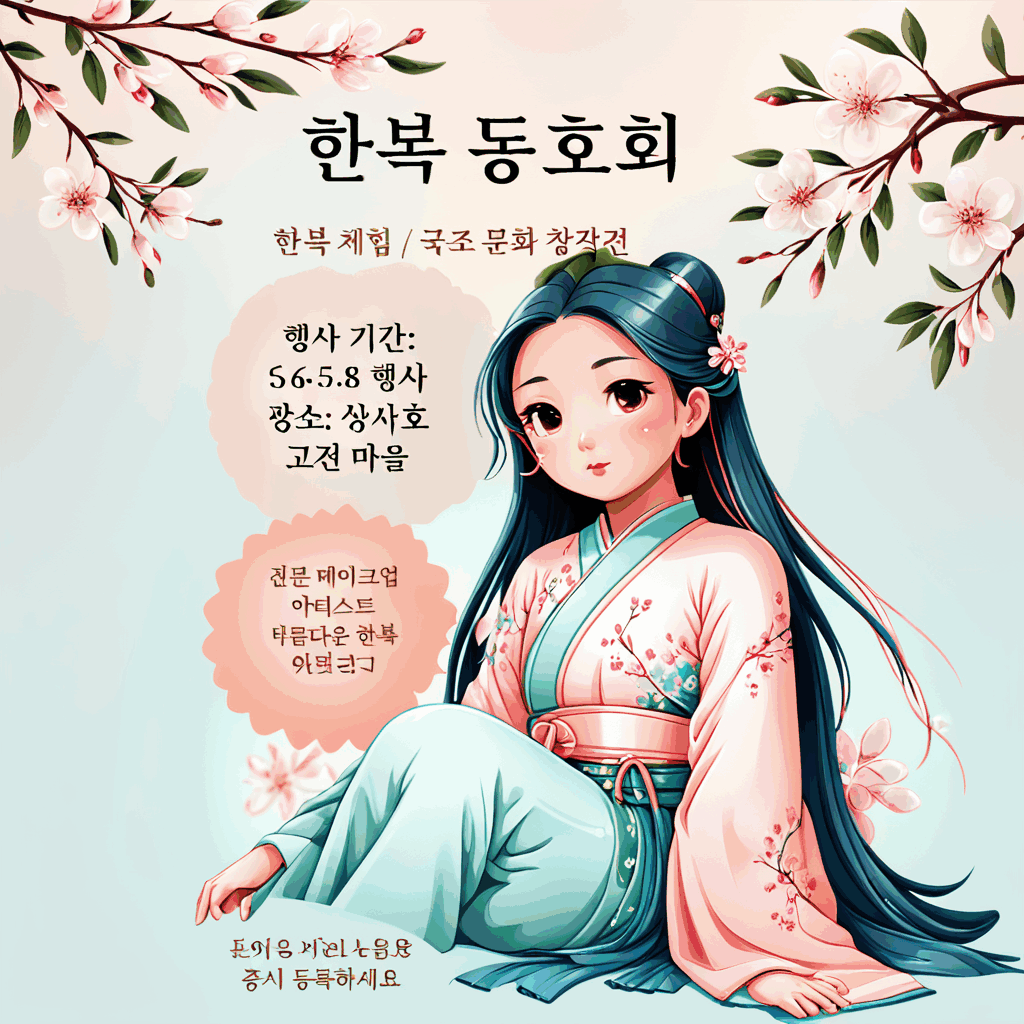}
\vspace{-3mm}
\end{subfigure}
\begin{subfigure}[b]{0.16\textwidth}
\includegraphics[width=\textwidth]{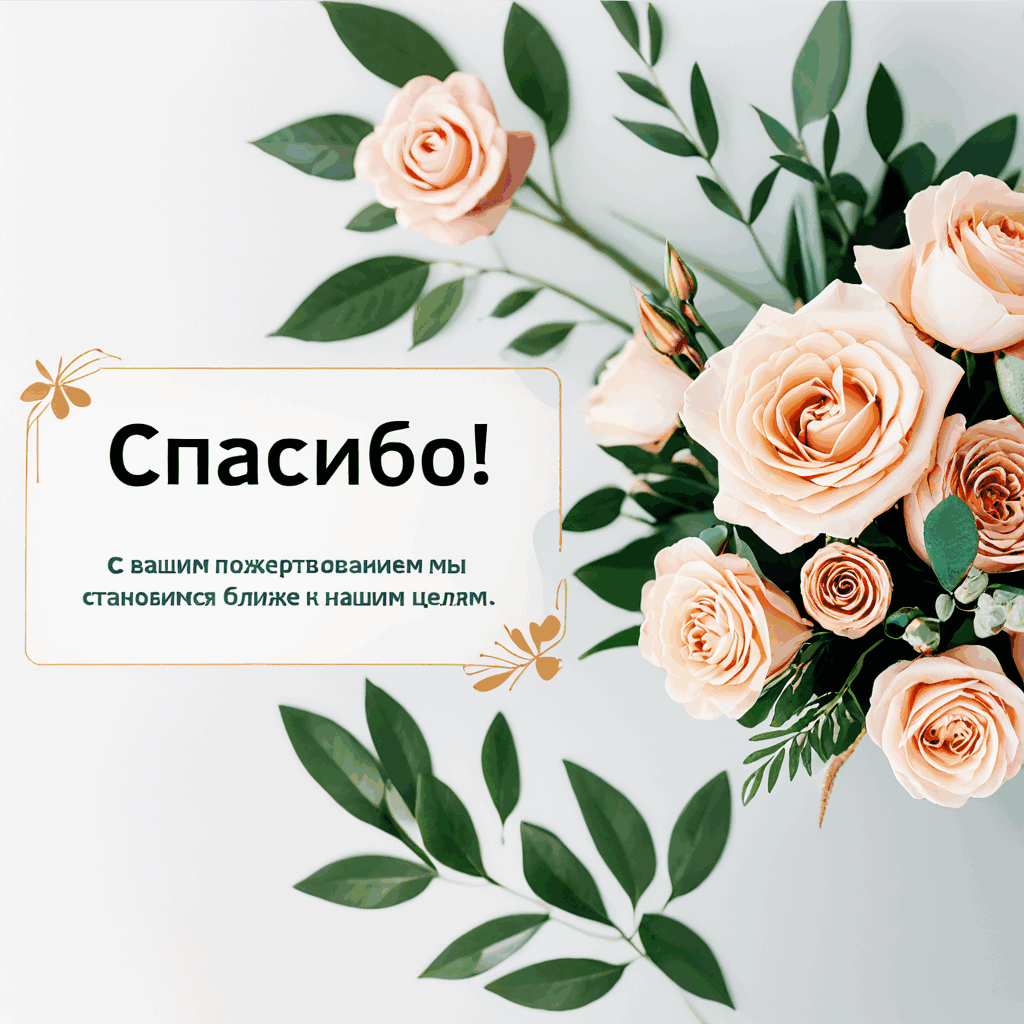}
\vspace{-3mm}
\end{subfigure}
\caption{\small{\textbf{Illustrating the effect of applying step-aware preference optimization (SPO) post-training.} Displayed in sequence are the images generated by: Glyph-SDXL on the first row, Glyph-SDXL Albedo~\cite{albedo} on the second row, and finally, Glyph-SDXL Albedo + SPO~\cite{liang2024step}  on the last row.}}
\label{fig:effect_spo}
\end{minipage}
\vspace{-2mm}
\end{figure*}

\subsection{Improving Aesthetics with SPO-SDXL}
Unlike the original Glyph-SDXL, which uses the default SDXL~\cite{podell2023sdxl}, we opted to use the improved SDXL after post-training, specifically the SPO-SDXL fine-tuned with a step-aware preference learning scheme~\cite{liang2024step}. Figure~\ref{fig:effect_spo} illustrates the improvements brought by applying the SPO post-training scheme.
We can observe that Glyph-SDXL Albedo + SPO generates the images of the best quality, which siginificantly outperforms the ones generated with the original SDXL, as shown in Figure~\ref{fig:winrate_spo_sdxl_vs_sdxl}

\subsection{Comparison with \dalle}
Last, we compare our approach with the latest \dalle on the multilingual visual text rendering task and illustrate the user study results in Figure~\ref{fig:winrate_dalle}. According to the comparison results, our method is preferred $91\%$ of the time over \dalle in terms of visual text quality. Notably, our approach also achieves comparable visual aesthetics to \dalle by leveraging the benefits of the latest SPO post-training scheme.

\section{Conclusion}
\label{sec:conclusion}
In this paper, we present an improved customized multilingual text encoder for accurate multilingual visual text rendering. We have built scalable, high-quality multilingual glyph-text and graphic design datasets, enabling the training of our models, Glyph-ByT5-v2 and Glyph-SDXL-v2. We empirically show that replacing the original SDXL with post-trained versions optimized for human preferences significantly enhances visual aesthetics. The effectiveness of our approach is demonstrated through detailed comparisons and user studies.
\subsection*{Appendix}

The detailed prompts for the generated images shown in Figures 1 and 5 are provided in Table~\ref{tab:prompt_list_1}, Table~\ref{tab:prompt_list_2}, Table~\ref{tab:prompt_list_3}, and Table~\ref{tab:prompt_list_4}, respectively.

\begin{table*}[htbp]
\begin{minipage}[t]{1\linewidth}  
\centering  
\tablestyle{1pt}{1}  
\resizebox{1.0\linewidth}{!}  
{  
\begin{tabular}{l|>{\centering\arraybackslash}m{16cm}}  
Image & Prompt \\  
\shline  
\scriptsize{Row 1, Col1} & \scriptsize{Background: Instagram Posts. The image features a woman sitting in a lotus position, also known as a yoga pose, with her legs crossed and her hands resting on her knees. She is surrounded by a serene environment, with trees in the background and a sun in the sky above her. The woman appears to be meditating or practicing yoga in a peaceful outdoor setting. Tags: WHITE, BROWN, BLUE, MODERN, meditation, exercise, fitness, yoga day, poster, health, illustration, international, position, concept, relaxation, yoga, woman Text: Text "passons à l'action pour trouver la santé" in $<$color-2$>$, $<$en-font-415$>$. Text "JOURNÉE YOGA" in $<$color-2$>$, $<$en-font-415$>$. Text "International" in $<$color-2$>$, $<$en-font-415$>$. } \\ \hline
\scriptsize{Row 1, Col2} & \scriptsize{Background: Instagram Posts. The image features a pink background with a white flower in the center. The flower is surrounded by a few leaves, creating a visually appealing arrangement. The flower and leaves are the main focus of the image, with the pink background providing a subtle contrast. Tags: ivory, brown, red, illustration, modern, memphis, workshop, design, graphic, work, creative, business, designer, creativity, company, presentation, education, website, fashion, graphic design Text: Text "GRATUIT!" in $<$color-4$>$, $<$en-font-188$>$. Text "ATELIERS " in $<$color-65$>$, $<$en-font-62$>$. Text "pour les débutants" in $<$color-4$>$, $<$en-font-340$>$. Text "CONCEPTION GRAPHIQUE" in $<$color-4$>$, $<$en-font-78$>$. } \\ \hline
\scriptsize{Row 1, Col3} & \scriptsize{Background: Posters. The image features a beautiful winter scene with snow-covered mountains in the background. There are several pine trees scattered throughout the scene, some of which are covered in snow. The mountains are adorned with a wreath, adding a festive touch to the landscape. The overall atmosphere of the image is serene and picturesque, capturing the essence of a snowy mountain range. Tags: Blue, green, Illustrated, colorful, Winter sale, sale, discount, advertising, promotion, special, offer, store, event, marketing, holiday, winter, blue, snow, illustration, template Text: Text "Soldes d'hiver" in $<$color-36$>$, $<$en-font-44$>$. Text "jusqu'à 60\% de réduction" in $<$color-67$>$, $<$en-font-44$>$. Text "Offre spéciale" in $<$color-67$>$, $<$en-font-44$>$. } \\ \hline
\scriptsize{Row 1, Col4} & \scriptsize{Background: Instagram Posts. The image features a basket filled with various eggs, including a green egg, a brown egg, and a white egg. The basket is placed on a wooden table, and the eggs are arranged in a visually appealing manner. In addition to the eggs, there are some flowers in the basket, adding a touch of color and natural beauty to the scene. The overall composition of the image is simple and elegant, showcasing the eggs and flowers as the main subjects. Tags: green, modern, easter day, happy easter day, easter, happy easter, decoration, happy, egg, spring, selebration, illustration, greeting, design, colorful, cute, template, bunny, day, festive Text: Text "fleurir où vous êtes planté" in $<$color-0$>$, $<$en-font-246$>$. Text "CONTENT.E" in $<$color-44$>$, $<$en-font-359$>$. Text "PÂQUES" in $<$color-44$>$, $<$en-font-408$>$. } \\ \hline
\scriptsize{Row 1, Col5} & \scriptsize{Background: Facebook Post. The image features a woman with long, flowing red hair, holding a bouquet of flowers in her hands. The bouquet consists of various types of flowers, including blue and yellow ones. The woman appears to be the main focus of the image, and she is surrounded by the vibrant colors of the flowers. Tags: pastel, colorful, cute, modern, feminime, abstract, illustration, international, woman, days, march 8, congratulation, greeting, girl, mother, eight, card, spring, social media Text: Text "Journée internationale de la femme" in $<$color-6$>$, $<$en-font-340$>$. Text "8 mars" in $<$color-68$>$, $<$en-font-234$>$. } \\ \hline

\scriptsize{Row 2, Col1} & \scriptsize{Background: Cards and invitations. The image features a bouquet of flowers, including roses, arranged in a vase. The bouquet is placed on a table, and the flowers are surrounded by green leaves. The flowers are of various sizes and are positioned in different directions, creating a visually appealing arrangement. The overall scene is a close-up view of the bouquet, showcasing its beauty and detail. Tags: gold, happy, minimalist, modern, illustration, colorful, social media, family, wellness, advocacy, love, boutique, small business, online shop, wylene, ylin creates, printable, message, personal, gratitude Text: Text "Con tu donación estamos más cerca de alcanzar nuestras metas." in $<$color-2$>$, $<$en-font-340$>$. Text "¡Gracias!" in $<$color-1$>$, $<$en-font-415$>$. } \\ \hline
\scriptsize{Row 2, Col2} & \scriptsize{Background: Instagram Posts. The image features a white background with a variety of colorful flowers and decorations. There are several pink flowers scattered throughout the scene, with some positioned closer to the top and others near the bottom. A blue flower can also be seen in the middle of the image. The overall composition creates a visually appealing and vibrant display. Tags: grey, navy, purple, pink, teal, colorful, illustration, happy, celebration, post, party, year, new, event, celebrate, happy new year, new year, countdown, sparkle, firework Text: Text "Feliz Año Nuevo" in $<$color-83$>$, $<$en-font-402$>$. Text "2024" in $<$color-5$>$, $<$en-font-112$>$. Text "TODO LO MEJOR" in $<$color-1$>$, $<$en-font-58$>$. Text "Un nuevo comienzo para iniciar un cambio para mejorar." in $<$color-1$>$, $<$en-font-58$>$. } \\ \hline
\scriptsize{Row 2, Col3} & \scriptsize{Background: Instagram Posts. The image features a large archway with a blue background. The archway is adorned with numerous hanging lanterns, creating a beautiful and illuminated atmosphere. The lanterns are positioned at various heights and angles, adding depth and dimension to the scene. The combination of the blue background and the hanging lanterns creates a visually appealing and captivating image. Tags: dark blue, yellow orange white, beige, illustration, geometric, vintage, border, eid, eid mubarak, eid al fitr, ramadan, islamic, arabic, kareem, muslim, culture, festival, instagram post, annecreativedesigns, annebacani Text: Text "Deseándote un tiempo bendecido con tu familia en este Ramadán." in $<$color-41$>$, $<$en-font-105$>$. Text "Ramadán Mubarak" in $<$color-41$>$, $<$en-font-324$>$. } \\ \hline
\scriptsize{Row 2, Col4} & \scriptsize{Background: Instagram Posts. The image features a group of stuffed animals, including a teddy bear, a dragon, and a rabbit, sitting together on a table. The teddy bear is positioned in the center of the scene, while the dragon is located to the left and the rabbit is on the right side. The stuffed animals are arranged in a way that they appear to be sitting on a bench, creating a cozy and playful atmosphere. Tags: orange, illustration, cartoon, pastel, vector, announcement, greeting, event, celebration, baby day, baby, kid, child, boy, day, mother, childhood, care, happy, instagram post Text: Text "Opciones de regalo únicas" in $<$color-14$>$, $<$en-font-144$>$. Text "Imprescindible para una familia cálida" in $<$color-0$>$, $<$en-font-144$>$. Text "Muñeca de trapo" in $<$color-0$>$, $<$en-font-500$>$. } \\ \hline
\scriptsize{Row 2, Col5} & \scriptsize{Background: Cards and invitations. The image features a white background with a decorative border surrounding it. The border is adorned with various animals, including elephants, giraffes, and birds. There are two elephants, one on the left side and the other on the right side of the border. Two giraffes can be seen, one in the middle and the other on the right side of the border. Additionally, there are three birds, with one on the left side, one in the middle, and another on the right side of the border. The combination of these animals creates a visually appealing and colorful design. Tags: green, cream, illustration, colorful, forest, invitation, animal, leaves, party, nature, card, tiger, tropical Text: Text "08:00AM" in $<$color-53$>$, $<$en-font-408$>$. Text "25 DE JUNIO DE 2023" in $<$color-53$>$, $<$en-font-408$>$. Text "Estás invitado a" in $<$color-6$>$, $<$en-font-402$>$. Text "SELVA" in $<$color-18$>$, $<$en-font-408$>$. Text "FIESTA EN LA" in $<$color-53$>$, $<$en-font-408$>$. } \\ \hline

\scriptsize{Row 3, Col1} & \scriptsize{Background: The image is a digital illustration featuring a collection of items related to pets, specifically dogs. The items are arranged in a grid-like pattern, with each item occupying its own square. Starting from the top left corner and moving row by row, the items are as follows:  1. A bone-shaped cookie. 2. A toothbrush. 3. A small brown dog bone. Text: Text "\begin{CJK}{UTF8}{gbsn}领养地点: 爱心宠物救助站\end{CJK}" in $<$color-6$>$, $<$cn-font-50$>$. Text "\begin{CJK}{UTF8}{gbsn}领养代替买卖，关爱流浪生灵\end{CJK}" in $<$color-6$>$, $<$cn-font-50$>$. Text "Pet Adoption" in $<$color-65$>$, $<$en-font-44$>$. Text "\begin{CJK}{UTF8}{gbsn}愿每个小宠物都受到关怀\end{CJK}" in $<$color-6$>$, $<$cn-font-50$>$. Text "\begin{CJK}{UTF8}{gbsn}宠物领养之家\end{CJK}" in $<$color-6$>$, $<$cn-font-34$>$. Text "\begin{CJK}{UTF8}{gbsn}联系电话: 13424624624\end{CJK}" in $<$color-6$>$, $<$cn-font-34$>$. } \\ \hline
\scriptsize{Row 3, Col2} & \scriptsize{Background: The image portrays a young girl sitting on a large green leaf. The leaf is part of a plant with other green leaves. The girl is wearing a yellow dress and a straw hat. She is holding a small yellow flower in her hand. The background of the image is a light blue sky with a few clouds. The overall style of the image is a colorful, cartoon-like illustration. Text: Text "\begin{CJK}{UTF8}{gbsn}小满，夏季的次序节气，象征作物籽粒渐趋充盈，尚在成长之中，未至全盛，故称小满，而非大满。\end{CJK}" in $<$color-4$>$, $<$cn-font-50$>$. Text "2024.5.20" in $<$color-26$>$, $<$en-font-231$>$. Text "\begin{CJK}{UTF8}{gbsn}麦穗初露，夏意渐浓，小满时节正当时\end{CJK}" in $<$color-4$>$, $<$cn-font-35$>$. Text "\begin{CJK}{UTF8}{gbsn}恰逢小满\end{CJK}" in $<$color-4$>$, $<$cn-font-50$>$. Text "Grain Buds" in $<$color-4$>$, $<$en-font-231$>$.} \\ \hline
\scriptsize{Row 3, Col3} & \scriptsize{Background: The image shows a simple, cartoon-style illustration. There are three characters depicted: a person sitting on a table, and two cats sitting on the table as well. The background is a soft, pastel pink color, which gives the image a calm and serene atmosphere. Text: Text "\begin{CJK}{UTF8}{gbsn}是哪个小调皮在偷看我的动态呢\end{CJK}" in $<$color-7$>$, $<$cn-font-35$>$. } \\
\end{tabular}
}
\vspace{-3mm}
\caption{  
\footnotesize{Detailed prompts for the generated images shown in Figure 1. (1/2)}}
\label{tab:prompt_list_1}
\end{minipage}
\end{table*}

\begin{table*}[htbp]
\vspace{-3mm}
\begin{minipage}[t]{1\linewidth}  
\centering  
\tablestyle{1pt}{1.2}  
\resizebox{1.0\linewidth}{!}  
{  
\begin{tabular}{l|>{\centering\arraybackslash}m{16cm}}  
Image & Prompt \\  
\shline  

\scriptsize{Row 3, Col4} & \scriptsize{Background: Instagram Posts. This is a digitally crafted tarot card, meticulously designed to take you on a journey through the cosmos, elegantly set against the mysterious backdrop of a dark, starry sky. The tarot card is surrounded by an ornate border featuring intricate patterns of stars and crescent moons, invoking the realm of the celestial heavens. At its heart, a majestic crescent moon shines brightly, its luminous rays cutting through the darkness. Below it, a banner is adorned with a decorative frame that echoes the overall celestial theme of the card. Tags: peach, modern, minimalist, colorful, illustration Text: Text "\begin{CJK}{UTF8}{gbsn}剥去你的伪装\end{CJK}" in $<$color-89$>$, $<$cn-font-34$>$. Text "\begin{CJK}{UTF8}{gbsn}星座之谜\end{CJK}" in $<$color-89$>$, $<$cn-font-35$>$. } \\ \hline
\scriptsize{Row 3, Col5} & \scriptsize{Background: The image depicts a traditional Chinese stage with vibrant colors and intricate designs. The stage is adorned with red and gold hues, which are commonly associated with good fortune and prosperity in Chinese culture. The stage is also decorated with various Chinese motifs and symbols, such as the Chinese dragon, which is often depicted as a symbol of strength, power, and good luck.  The stage is framed by two large, ornate Chinese lanterns, which are typically used to illuminate streets and buildings during nighttime festivities and celebrations.  The overall style of the image is reminiscent of traditional Chinese paper-cut art, which is a popular form of folk art in China that involves cutting out intricate designs and patterns using scissors.  The image is likely intended to evoke a sense of cultural heritage, tradition, and artistic craftsmanship associated with Chinese culture and history. Text: Text "\begin{CJK}{UTF8}{gbsn}依照法定节假日安排，特发布以下放假安排通知：2024年春节2月10日至2月17日放假，共8天，敬告周知。\end{CJK}" in $<$color-0$>$, $<$cn-font-134$>$. Text "\begin{CJK}{UTF8}{gbsn}预祝大家节日快乐！\end{CJK}" in $<$color-0$>$, $<$cn-font-134$>$. Text "\begin{CJK}{UTF8}{gbsn}休假公告\end{CJK}" in $<$color-0$>$, $<$cn-font-134$>$. } \\ \hline
\scriptsize{Row 4, Col1} & \scriptsize{Background: The image shows a beautifully decorated cake placed on a dining table. The cake is adorned with fresh berries, cream, and chocolate. The table setting is simple, with the cake being the main focus. The image is likely used for culinary purposes, such as a recipe or a food blog. Text: Text "\begin{CJK}{UTF8}{min}人生の転機二十歳幸せ\end{CJK}" in $<$color-33$>$, $<$jp-font-159$>$. } \\ \hline
\scriptsize{Row 4, Col2} & \scriptsize{Background: The image shows a nighttime cityscape with a dark sky filled with stars. The city is illuminated with various lights, suggesting a bustling urban environment. The image is framed by a black border, and there is a watermark or logo in the bottom right corner, which appears to be a stylized letter 'C'. The overall style of the image is illustrative and colorful, with a focus on the contrast between the dark sky and the brightly lit city. Text: Text "\begin{CJK}{UTF8}{min}おおよみそか\end{CJK}" in $<$color-0$>$, $<$jp-font-159$>$. Text "\begin{CJK}{UTF8}{min}除夜を祝う\end{CJK}" in $<$color-0$>$, $<$jp-font-159$>$. } \\ \hline
\scriptsize{Row 4, Col3} & \scriptsize{Background: The image features a repeating pattern of floral motifs in a light blue color. The floral motifs are symmetrically placed around the image. The background of the image is a solid dark blue color. The overall style of the image is graphic and minimalistic. Text: Text "\begin{CJK}{UTF8}{min}2 0 2 4除夕おめでとうございます\end{CJK}" in $<$color-0$>$, $<$jp-font-170$>$. } \\ \hline
\scriptsize{Row 4, Col4} & \scriptsize{Background: The image features a Shiba Inu dog with a human-like expression. The dog is wearing a plaid jacket and a scarf. The background is a solid yellow color. There is a speech bubble with a white background and a black border. The speech bubble is empty, suggesting that the dog is speaking or about to speak. Text: Text "\begin{CJK}{UTF8}{min}独身のワンちゃん\end{CJK}" in $<$color-6$>$, $<$jp-font-147$>$. Text "\begin{CJK}{UTF8}{min}私の名前は\end{CJK}" in $<$color-6$>$, $<$jp-font-147$>$. } \\ \hline
\scriptsize{Row 4, Col5} & \scriptsize{Background: The image features two iced coffee drinks, each with a layer of cream on top. The drinks are placed on a white surface, and there is a small, white cloud-like shape above the drinks. The image has a playful and whimsical feel to it. Text: Text "\begin{CJK}{UTF8}{min}早起は困難な人々目を覚ますために助けが必要です\end{CJK}" in $<$color-1$>$, $<$jp-font-205$>$. Text "\begin{CJK}{UTF8}{min}毎日は二度目のチャンスです一日一日を大切にするために\end{CJK}" in $<$color-1$>$, $<$jp-font-205$>$. Text "\begin{CJK}{UTF8}{min}人生は良すぎず、コーヒーは悪すぎません\end{CJK}" in $<$color-1$>$, $<$jp-font-205$>$. Text "\begin{CJK}{UTF8}{min}29.9/カップ\end{CJK}" in $<$color-1$>$, $<$jp-font-205$>$. Text "\begin{CJK}{UTF8}{min}挽きたてのコーヒー豆\end{CJK}" in $<$color-1$>$, $<$jp-font-42$>$. Text "\begin{CJK}{UTF8}{min}秋季限定\end{CJK}" in $<$color-1$>$, $<$jp-font-157$>$. Text "\begin{CJK}{UTF8}{min}生活は苦しいですがカフェラテを選びましょう\end{CJK}" in $<$color-1$>$, $<$jp-font-3$>$. } \\ \hline
\scriptsize{Row 5, Col1} & \scriptsize{Background: The image is a vibrant and colorful graphic design. It features a group of people, possibly a band, depicted in a dynamic and energetic pose. The people are surrounded by a variety of colorful elements, such as musical notes, sound waves, and other abstract shapes. The overall style of the image is reminiscent of a retro or vintage aesthetic, with its bold colors and playful design elements. Text: Text "\begin{CJK}{UTF8}{mj}· 톱 10 대학 가수 대회 ·\end{CJK}" in $<$color-72$>$, $<$kr-font-79$>$. Text "\begin{CJK}{UTF8}{mj}Super Singers 캠퍼스 톤 최고, 불안해하지 마, 너무 어리게 굴지 마\end{CJK}" in $<$color-14$>$, $<$kr-font-79$>$. Text "\begin{CJK}{UTF8}{mj}12/31 가수콘테스트가 2024년 12월 31일 19:00 학생활동관 A103호에서 개최됩니다\end{CJK}" in $<$color-0$>$, $<$kr-font-79$>$. } \\ \hline
\scriptsize{Row 5, Col2} & \scriptsize{Background: The image shows a digital illustration of a laptop with a blank screen. On the right side of the laptop, there is a cartoon of a person with a speech bubble that says "Hello!" The person appears to be waving at the viewer. The background of the image is a light yellow color with a grid pattern. The overall style of the image is a combination of digital illustration and cartoon elements. Text: Text "\begin{CJK}{UTF8}{mj}창의력 부여 대회창의력이 미래에 부여됩니다창의성 —\end{CJK}" in $<$color-2$>$, $<$kr-font-32$>$. Text "2024" in $<$color-24$>$, $<$en-font-116$>$. } \\ \hline
\scriptsize{Row 5, Col3} & \scriptsize{Background: The image showcases a sumptuous spread of dishes, each a mosaic of flavors with beef, peppers, and onions taking center stage. Arranged in a communal pattern, the vibrant colors of the food stand out against a darker backdrop, creating an irresistible visual feast that whets the appetite. Text: Text "\begin{CJK}{UTF8}{mj}도움이 되는 무료\end{CJK}" in $<$color-0$>$, $<$kr-font-92$>$. Text "06.16-06.20" in $<$color-57$>$, $<$en-font-268$>$. Text "\begin{CJK}{UTF8}{mj}친구들을 부르세요\end{CJK}" in $<$color-0$>$, $<$kr-font-72$>$. } \\ \hline
\scriptsize{Row 5, Col4} & \scriptsize{Background: The image is a vibrant and colorful illustration. It features a variety of objects and elements that are arranged in a way that suggests a creative or artistic theme. The objects include what appears to be a piece of paper or a card, which is blank and centrally located. Surrounding the paper or card are various decorative elements such as what looks like a checkered pattern, a floral pattern, and some geometric shapes. The colors used in the illustration are bright and varied, with a focus on warm tones. The overall style of the image is whimsical and playful, with a sense of creativity and imagination. Text: Text "10.21~10.31" in $<$color-2$>$, $<$en-font-408$>$. Text "\begin{CJK}{UTF8}{mj}할인\end{CJK}" in $<$color-2$>$, $<$kr-font-92$>$. Text "\begin{CJK}{UTF8}{mj}55\% 할인\end{CJK}" in $<$color-23$>$, $<$kr-font-92$>$. Text "\begin{CJK}{UTF8}{mj}최고\end{CJK}" in $<$color-2$>$, $<$kr-font-92$>$. Text "\begin{CJK}{UTF8}{mj}풍부한 가을풍성한 할인\end{CJK}" in $<$color-23$>$, $<$kr-font-32$>$. Text "\begin{CJK}{UTF8}{mj}합리적이고 적절한 가격으로  풍성한 가을을 준비하세요!\end{CJK}" in $<$color-2$>$, $<$kr-font-32$>$. } \\ \hline
\scriptsize{Row 5, Col5} & \scriptsize{Background: The image is a digital illustration featuring a character that appears to be a young woman with a serene expression. She is depicted with long, flowing hair and is wearing a traditional East Asian-style dress with a floral pattern. The dress is predominantly in shades of blue and green, with a hint of pink. The character is seated on a bed of cherry blossoms, which are scattered around her. The blossoms are in full bloom, with their delicate pink petals and white stamens.  The background of the image is a pale, soft blue sky with a few wispy clouds. The overall atmosphere of the image is one of tranquility and serenity. Text: Text "\begin{CJK}{UTF8}{mj}QR 코드를 스캔하여 즉시 등록하세요\end{CJK}" in $<$color-6$>$, $<$kr-font-45$>$. Text "\begin{CJK}{UTF8}{mj}전문 메이크업 아티스트 아름다운 한복 무료 촬영\end{CJK}" in $<$color-6$>$, $<$kr-font-45$>$. Text "\begin{CJK}{UTF8}{mj}행사 기간: 5.6-5.8 행사 장소: 상사호 고전 마을\end{CJK}" in $<$color-1$>$, $<$kr-font-45$>$. Text "\begin{CJK}{UTF8}{mj}한복 동호회\end{CJK}" in $<$color-1$>$, $<$kr-font-45$>$. Text "\begin{CJK}{UTF8}{mj}한복 체험 / 국조 문화 창작전\end{CJK}" in $<$color-6$>$, $<$kr-font-45$>$. } \\ 
\end{tabular}  
}
\caption{  
\footnotesize{Detailed prompts for the generated images shown in Figure 1. (2/2)}}
\label{tab:prompt_list_2}
\vspace{3mm}
\end{minipage}
\end{table*}

\begin{table*}[htbp]
\vspace{-3mm}
\begin{minipage}[t]{1\linewidth}  
\centering  
\tablestyle{1pt}{1.2}  
\resizebox{1.0\linewidth}{!}  
{  
\begin{tabular}{l|>{\centering\arraybackslash}m{16cm}}  
Image & Prompt \\  
\shline  

\scriptsize{Row 1, Col1} & \scriptsize{Background: Cards and invitations. The image features a large gray elephant sitting in a field of flowers, holding a smaller elephant in its arms. The scene is quite serene and picturesque, with the two elephants being the main focus of the image. The field is filled with various flowers, creating a beautiful and vibrant backdrop for the elephants. Tags: Light green, orange, Illustration, watercolor, playful, Baby shower invitation, baby boy shower invitation, baby boy, welcoming baby boy, koala baby shower invitation, baby shower invitation for baby shower, baby boy invitation, background, playful baby shower card, baby shower, card, newborn, born, Baby Shirt Baby Shower Invitation Text: Text "RSVP an +123-456-7890" in $<$color-18$>$, $<$en-font-38$>$. Text "Sophia Davis" in $<$color-99$>$, $<$en-font-231$>$. Text "Babyparty" in $<$color-53$>$, $<$en-font-234$>$. Text "Bitte schließen Sie sich uns an für eine" in $<$color-18$>$, $<$en-font-231$>$. Text "Zu Ehren" in $<$color-18$>$, $<$en-font-231$>$. Text "23. November 2024 | 15:00 Uhr Fauget Hotels" in $<$color-18$>$, $<$en-font-23$>$. } \\ \hline
\scriptsize{Row 1, Col2} & \scriptsize{Background: Photos and gifts. The image features a brown teddy bear sitting on a white surface, possibly a table. The teddy bear is positioned in a relaxed manner, with its arms crossed and legs crossed. Above the teddy bear, there is a heart-shaped balloon, adding a playful touch to the scene. The overall setting appears to be a cozy and inviting environment. Tags: brown, pink, illustration, cute, bear, teddy, love, valentine, mug, drink Text: Text "Beary Much" in $<$color-6$>$, $<$en-font-231$>$. Text "Ich liebe dich" in $<$color-6$>$, $<$en-font-234$>$. } \\ \hline
\scriptsize{Row 1, Col3} & \scriptsize{Background: Pinterest pins. The image features a white card with a flower drawn on it. The flower is positioned towards the bottom right corner of the card. The card is placed on a table, and a vase with a potted plant is situated nearby. The plant is located in the middle of the table, with the vase holding it. The combination of the card and the plant creates a pleasant and artistic scene. Tags: brown, yellow, chocolate, orange, simple, modern, aesthetic, illustration, post, marketing, thank you, fashion, food, social media, quote, greeting Text: Text "Glyph-ByT5 for AIGC" in $<$color-2$>$, $<$en-font-404$>$. Text "FÜR 100,000 FOLLOWER" in $<$color-2$>$, $<$en-font-404$>$. Text "DU, SIE, IHR" in $<$color-2$>$, $<$en-font-404$>$. Text "Danke" in $<$color-2$>$, $<$en-font-404$>$. } \\ \hline
\scriptsize{Row 1, Col4} & \scriptsize{Background: Instagram Posts. The image features two brown teddy bears sitting next to each other on a grassy field. They are positioned close to one another, with one teddy bear slightly larger than the other. The bears appear to be enjoying their time together in the outdoors. Tags: beige, green, yellow, pastel, illustration, watercolor, playful, colorful, aesthetic, abstract, world wildlife day, event, animal, nature, save earth, protect, national, wildlife, quote, reminder Text: Text "Tiere wurden nicht von Gott erschaffen, damit wir sie jagen oder essen können, lasst uns sie retten" in $<$color-18$>$, $<$en-font-362$>$. Text "Wildtiere" in $<$color-18$>$, $<$en-font-362$>$. Text "Welttag" in $<$color-18$>$, $<$en-font-362$>$. Text "der" in $<$color-26$>$, $<$en-font-362$>$. } \\ \hline
\scriptsize{Row 1, Col5} & \scriptsize{Background: Videos. The image features a person's hand holding a small plant in a clump of dirt. The plant is growing out of the dirt, and it appears to be a sapling or a small tree. The hand is positioned in the center of the scene, with the plant and dirt occupying the majority of the space. The focus of the image is on the growth and nurturing of the plant, showcasing the care and attention given to it. Tags: green, illustration, modern, tree, announcement, greeting, forest, celebration, world, environment, day, earth, save, planet, conservation, environmental, plant, ecosystem, recycle, video Text: Text "Danke" in $<$color-0$>$, $<$en-font-1$>$. Text "unsere" in $<$color-0$>$, $<$en-font-1$>$. Text "Mutter Erde" in $<$color-0$>$, $<$en-font-1$>$. Text "Bitte beschütze" in $<$color-0$>$, $<$en-font-1$>$. } \\ \hline

\scriptsize{Row 2, Col1} & \scriptsize{Background: Posters. The image features a green and blue globe with a factory on top of it. The factory is surrounded by trees, giving the impression of a harmonious coexistence between the industrial structure and the natural environment. The globe is prominently displayed in the center of the image, with the factory and trees surrounding it. Tags: green, modern, earth, world, planet, ecology, background, globe, environment, day, space, map, concept, global, light, hour, energy, power, protect, illustration Text: Text "A TERRA É O QUE TODOS NÓS TEMOS EM COMUM" in $<$color-0$>$, $<$en-font-105$>$. Text "Dia da Terra" in $<$color-0$>$, $<$en-font-362$>$. } \\ \hline
\scriptsize{Row 2, Col2} & \scriptsize{Background: Instagram Posts. The image features a stack of pancakes with syrup and strawberries on top. The pancakes are arranged in a visually appealing manner, with some pancakes placed on top of each other. The syrup is drizzled generously over the pancakes, and the strawberries are scattered around, adding a touch of color and freshness to the scene. The overall presentation of the pancakes is appetizing and inviting. Tags: brown, peach, grey, modern, minimalist, simple, colorful, illustration, Instagram post, instagram, post, national pancake day, international pancake day, happy pancake day, pancake day, pancake, sweet, cake, discount, sale Text: Text "Obtenha 75\% de desconto no seu primeiro pedido" in $<$color-3$>$, $<$en-font-173$>$. Text "Peça agora" in $<$color-0$>$, $<$en-font-323$>$. Text "Dia Nacional da Panqueca" in $<$color-4$>$, $<$en-font-234$>$. } \\ \hline
\scriptsize{Row 2, Col3} & \scriptsize{Background: Facebook Post. The image features a wooden table with a variety of fruits and vegetables arranged on it. There are several apples, with one placed towards the center of the table and others scattered around. A couple of oranges are also present, with one located near the right side of the table and the other towards the left.  In addition to the fruits, there is a cup placed on the table, possibly containing a refreshing drink. The overall scene is a vibrant and healthy display of fresh produce. Tags: Yellow, Green, Bold, Simple, Modern, Minimalist, celebration, party, happy, holiday, fun, event, decoration, illustration, balloon, confetti, celebrate, background, colorful, card Text: Text "Obtenha Smoothie Grátis" in $<$color-4$>$, $<$en-font-89$>$. Text "100 mil seguidores" in $<$color-4$>$, $<$en-font-66$>$. Text "Obrigado" in $<$color-4$>$, $<$en-font-66$>$. } \\ \hline
\scriptsize{Row 2, Col4} & \scriptsize{Background: Facebook Post. The image features a small brown bunny rabbit sitting in a basket filled with various flowers. The basket is placed on a yellow background, creating a vibrant and cheerful scene. The flowers surrounding the rabbit come in different sizes and colors, adding to the overall visual appeal of the image. The rabbit appears to be the main focus of the scene, and its presence among the flowers creates a sense of harmony and balance. Tags: green, yellow, minimalist, easter day, happy easter day, easter, happy easter, decoration, happy, egg, spring, selebration, poster, illustration, greeting, season, design, colorful, cute, template Text: Text "QUE TODAS AS SUAS ORAÇÕES SEJAM ATENDIDAS" in $<$color-4$>$, $<$en-font-302$>$. Text "DOMINGO, 17 DE ABRIL" in $<$color-4$>$, $<$en-font-288$>$. Text "TENHA UM FELIZ" in $<$color-4$>$, $<$en-font-475$>$. Text "Dia da Páscoa" in $<$color-4$>$, $<$en-font-475$>$. } \\ \hline
\scriptsize{Row 2, Col5} & \scriptsize{Background: Posters. The image features a white plate filled with a variety of fruits and vegetables. There are several carrots, with one placed towards the left side of the plate, another in the middle, and the third one on the right side. A banana is also present on the plate, located towards the left side. The plate is placed on a dining table, which occupies the entire background of the image. Tags: yellow, red, green, colorful, illustration, vegetarian, vegetable, food, delicious, vegan, world, organic, healthy, healthy food, vegetarian day, lifestyle, eating, freshness, salad, nutrition Text: Text "1 DE OUTUBRO DE 2024" in $<$color-1$>$, $<$en-font-496$>$. Text- "DIA MUNDIAL DO VEGETARIANO" in $<$color-1$>$, $<$en-font-496$>$. } \\
\end{tabular}  
}
\caption{  
\footnotesize{Detailed prompts for the generated images shown in Figure 3. (1/2)}}
\label{tab:prompt_list_3}
\end{minipage}
\end{table*}

\begin{table*}[htbp]
\vspace{-3mm}
\begin{minipage}[t]{1\linewidth}  
\centering  
\tablestyle{1pt}{1.2}  
\resizebox{1.0\linewidth}{!}  
{  
\begin{tabular}{l|>{\centering\arraybackslash}m{16cm}}  
Image & Prompt \\  
\shline  
\scriptsize{Row 3, Col1} & \scriptsize{Background: Cards and invitations. The image features a bouquet of flowers, including roses, arranged in a vase. The bouquet is placed on a table, and the flowers are surrounded by green leaves. The flowers are of various sizes and are positioned in different directions, creating a visually appealing arrangement. The overall scene is a close-up view of the bouquet, showcasing its beauty and detail. Tags: gold, happy, minimalist, modern, illustration, colorful, social media, family, wellness, advocacy, love, boutique, small business, online shop, wylene, ylin creates, printable, message, personal, gratitude Text: Text "Con la tua donazione siamo più vicini ai nostri obiettivi." in $<$color-2$>$, $<$en-font-190$>$. Text "Grazie!" in $<$color-1$>$, $<$en-font-190$>$. } \\ \hline
\scriptsize{Row 3, Col2} & \scriptsize{Background: The image features a square with a pink background, adorned with a floral pattern. The pattern consists of various flowers, including a prominent rose in the center of the square. The rose is surrounded by other flowers, creating a visually appealing and colorful design. The overall appearance of the image is vibrant and eye-catching. Text: Text "OLIVIA E OWEN VI INVITANO A UNA SERATA DI DRINK E BALLO PER CELEBRARE IL LORO MATRIMONIO SABATO 14 SETTEMBRE 2024 ORE 19:00 STANHEM HOUSE, BRIGHTON" in $<$color-115$>$, $<$en-font-475$>$. } \\ \hline
\scriptsize{Row 3, Col3} & \scriptsize{Background: Facebook Post. The image features two bowls of food placed on a dining table. The first bowl is located on the left side of the table, while the second bowl is on the right side. Both bowls are filled with a variety of food items, including broccoli, carrots, and other vegetables. The broccoli can be seen in different parts of the bowls, with some pieces closer to the center and others near the edges. The carrots are also scattered throughout the bowls, adding to the colorful and appetizing presentation of the dishes. Tags: peach, brown, red, illustration, abstract, food promo, special offer, sale, discount, restaurant, chinese food, asian food, traditional food, big promo day, 70\% off, chinese traditional food, restaurant promo, chicken promo, promotion, meal discount Text: Text "ORDINA ADESSO" in $<$color-51$>$, $<$en-font-8$>$. Text "e ottieni sconti fino al" in $<$color-21$>$, $<$en-font-8$>$. Text "Godetevi il delizioso" in $<$color-21$>$, $<$en-font-8$>$. Text "70\%" in $<$color-2$>$, $<$en-font-65$>$. Text "Stufato di gamberi" in $<$color-2$>$, $<$en-font-429$>$. } \\ \hline
\scriptsize{Row 3, Col4} & \scriptsize{Background: Pinterest pins. The image features a dining table filled with a variety of food items. There is a large roasted chicken placed in the center of the table, surrounded by several bowls containing different dishes. A pumpkin can also be seen on the table, adding to the festive atmosphere.  In addition to the food, there are utensils such as forks and knives scattered across the table, ready for use. A person is visible in the background, likely preparing to enjoy the meal. The table is set for a delightful and hearty feast. Tags: beige, gold, red, playful, abstract, organic, minimalist, boho, promo, recipes, thanksgiving, thanksgiving recipes, fall, blogger, blog post, small business, food blogger, greenery, fall leaves, illustration Text: Text "LEGGI ORA" in $<$color-7$>$, $<$en-font-7$>$. Text "Le mie ricette preferite per il Ringraziamento" in $<$color-25$>$, $<$en-font-7$>$. } \\ \hline
\scriptsize{Row 3, Col5} & \scriptsize{Background: Instagram Posts. The image features a lush green forest with a blue sky in the background. The sky is filled with clouds, creating a picturesque scene. The forest is filled with trees, some of which have leaves that are larger than the others. The leaves are scattered throughout the scene, adding depth and texture to the image. Tags: Green, Blue, Creative, Illustration, text, notes, motivation, positive, inspiration, quotes, greetings, messages, nature, environment, world, globe, earth, ozone, layer, life Text: Text "Salvare la natura inizia con un piccolo passo dall'interno. Preservare lo strato di ozono, incoraggiando tutti a riciclare." in $<$color-2$>$, $<$en-font-168$>$. } \\ \hline

\scriptsize{Row 4, Col1} & \scriptsize{Background: Cards and invitations. The image features a man and a woman standing close to each other, hugging each other in a forest setting. They are surrounded by various objects, such as a suitcase, a book, a heart, and a teddy bear. The scene appears to be a collage of different elements, possibly representing a romantic moment or a special memory. Tags: love, pink, handcrafted, couple, happy, illustrative, general audience, red, ripped, paper, postcard, valentines, valentine's day greeting, valentine's postcard, playful, whimsical, valentines day, map, stamps, handcrafted \& heartfelt, artistic, photo, cheerful, valentines day postcard, valentine greeting, personal, personal / generic greetings, positive, creative, heart, script, valentine's day, lively, livelyfree, valentines day greeting, heartfelt, valentine's day postcard, valentines postcard, valentine's greeting, illustration, general greeting, fun Text: Text "\foreignlanguage{russian}{Ты значишь для меня мир!}" in $<$color-21$>$, $<$en-font-55$>$. } \\ \hline
\scriptsize{Row 4, Col2} & \scriptsize{Background: Cards and invitations. The image features a bouquet of flowers, including roses, arranged in a vase. The bouquet is placed on a table, and the flowers are surrounded by green leaves. The flowers are of various sizes and are positioned in different directions, creating a visually appealing arrangement. The overall scene is a close-up view of the bouquet, showcasing its beauty and detail. Tags: gold, happy, minimalist, modern, illustration, colorful, social media, family, wellness, advocacy, love, boutique, small business, online shop, wylene, ylin creates, printable, message, personal, gratitude Text: Text "\foreignlanguage{russian}{С вашим пожертвованием мы становимся ближе к нашим целям.}" in $<$color-2$>$, $<$en-font-55$>$. Text "\foreignlanguage{russian}{Спасибо!}" in $<$color-1$>$, $<$en-font-55$>$. } \\ \hline
\scriptsize{Row 4, Col3} & \scriptsize{Background: The image features a castle with a dragon perched on top of it. The dragon is positioned in the middle of the scene, with its body extending from the left side to the right side of the castle. The castle itself is situated on a grassy hill, giving the impression of a fantasy setting. The scene is set against a backdrop of a blue sky, which adds to the overall atmosphere of the image. Text: Text "\foreignlanguage{russian}{Однажды ночью, когда принцесса Лили смотрела в окно, она увидела Драко, сидящего снаружи. Сначала она испугалась, но затем заметила добрые глаза и нежную улыбку дракона.}" in $<$color-50$>$, $<$en-font-488$>$. } \\ \hline
\scriptsize{Row 4, Col4} & \scriptsize{Background: Cards and invitations. The image features a group of robots standing together, with some of them holding hands. There are a total of six robots in the scene, with varying sizes and positions. Some of the robots are closer to the foreground, while others are further back, creating a sense of depth in the image. The robots are arranged in a way that they appear to be interacting with each other, possibly as a part of a celebration or gathering. Tags: sage, sage green, green, orange, red, fun, playful, illustration, illustrated, kid, kids, birthday, celebration, party, invitation, happy birthday, robot, futuristic, gear, fireworks Text: Text "\foreignlanguage{russian}{СЕБАСТЬЯН 7-ОЙ ДЕНЬ РОЖДЕНИЯ}" in $<$color-23$>$, $<$en-font-380$>$. Text "\foreignlanguage{russian}{ВЫ ПРИГЛАШЕНЫ НА}" in $<$color-2$>$, $<$en-font-320$>$. } \\ \hline
\scriptsize{Row 4, Col5} & \scriptsize{Background: Instagram Posts. The image features a clown sitting in a lotus position, surrounded by several colorful eggs. The clown is positioned in the center of the scene, with the eggs scattered around him. Some of the eggs are located near the clown's feet, while others are placed further away. The clown appears to be balancing the eggs on his body, showcasing his unique talent. Tags: purple, colorful, creative, modern, April, happy, celebration, holiday, day, vector, funny, spring, joker, party, illustration, sale, comedy, joy, festival, crazy Text: Text "\foreignlanguage{russian}{ДЕНЬ ДУРАКА (1 АПРЕЛЯ)}" in $<$color-0$>$, $<$en-font-106$>$. Text "\foreignlanguage{russian}{ДЕНЬ РЫБАКА В ПОИСКАХ СЧАСТЬЯ}" in $<$color-0$>$, $<$en-font-15$>$. Text "\foreignlanguage{russian}{СЧАСТЛИВЫЙ}" in $<$color-0$>$, $<$en-font-352$>$. } \\

\end{tabular}  
}
\caption{  
\footnotesize{Detailed prompts for the generated images shown in Figure 3. (2/2)}}
\label{tab:prompt_list_4}
\vspace{3mm}
\end{minipage}
\end{table*}  

{
    \small
    \bibliographystyle{ieeenat_fullname}
    \bibliography{main}
}

\end{document}